\documentclass[11pt]{article}
\usepackage[T1]{fontenc}
\usepackage{lmodern,amsmath,amsthm,amsfonts,amssymb,mathrsfs,bm}
\usepackage[a4paper,margin=1.15in]{geometry}
\usepackage{hyperref}
\hypersetup{hidelinks}

\newtheorem{theorem}{Theorem}[section]
\newtheorem{lemma}[theorem]{Lemma}
\newtheorem{corollary}[theorem]{Corollary}
\newtheorem{proposition}[theorem]{Proposition}
\newtheorem{assumption}[theorem]{Assumption}
\newtheorem{definition}{Definition}[section]
\newtheorem{remark}{Remark}[section]
\newcommand{\R}{\mathbb R}
\newcommand{\N}{\mathbb N}
\newcommand{\E}{\mathbb E}
\renewcommand{\P}{\mathbb P}
\newcommand{\M}{\mathcal M}
\newcommand{\X}{\mathcal X}
\newcommand{\Sset}{\mathcal S}
\newcommand{\Gcal}{\mathcal G}
\newcommand{\Kcal}{\mathcal K}
\newcommand{\Lcal}{\mathcal L}
\newcommand{\eps}{\varepsilon}
\newcommand{\one}{\mathbf 1}
\newcommand{\dd}{\,\mathrm d}
\DeclareMathOperator{\var}{Var}
\DeclareMathOperator{\diag}{diag}

\DeclareMathOperator{\Tr}{Tr}

\DeclareMathOperator{\diam}{diam}

\DeclareMathOperator{\op}{op}
\DeclareMathOperator{\HS}{HS}
\DeclareMathOperator{\vol}{vol}

\makeatletter
\renewcommand*{\@cite@ofmt}{\hbox}
\makeatother

\begin{document}
\title{Sampling with graph wavelet kernels: optimality and adaptivity to regularity and geometry}
\author{
  \small\begin{tabular}{c}
		{Pranav Gupta
        \thanks{Corresponding author}}
		 \\  Division of Applied Mathematics\\
		 Brown University\\
         pranav\_gupta@brown.edu
	\end{tabular}
    \and
    \small\begin{tabular}{c}
		{Hoang-Son Tran  \thanks{Corresponding author}}\\  
        Department of Mathematics\\
		 National University of Singapore\\ transonsp97@gmail.com
	\end{tabular}
    \and
    \small\begin{tabular}{c}
		{Subhro Ghosh}
		 \\  Department of Mathematics\\
		 National University of Singapore\\
         subhrowork@gmail.com
	\end{tabular}
}
\date{}
\maketitle

\begin{abstract}
We construct wavelet-based determinantal samplers on manifolds and geometric graphs whose approximation rates adapt to the regularity of the observable. On a Riemannian manifold of dimension $m$, our wavelet DPPs satisfy variance reduction of order $n^{1-2s/m}$ for linear statistics of both $s$-H\"older and $H^s$ observables, where $0<s\le1$. Using these samplers, we further construct unbiased quadrature rules with root mean square error $O(n^{-1/2-s/m})$ using at most $3n/2$ function evaluations, matching the classical optimal randomized integration exponent. For point clouds sampled from an $m$-dimensional manifold, we develop corresponding wavelet determinantal samplers on two standard classes of geometric random graphs: $\varepsilon$-graphs and symmetric $k$-nearest-neighbour graphs. These samplers satisfy the same variance reduction for linear statistics of $s$-H\"older observables and yield conditionally unbiased estimators of empirical averages with root mean square error $O(n^{-1/2-s/m})$, using at most $2n$ function evaluations. In both settings, the rates are governed by the intrinsic dimension $m$ and the regularity $s$ of the observable.
\end{abstract}

\tableofcontents

\section{Introduction}\label{sec:intro}
Many problems in numerical integration and machine learning require approximating an integral or a large empirical average using only a small number of function evaluations. Formally, let $\mu$ be a probability measure on a space $\mathcal{X}$, and let $\mathcal{F}$ be a family of integrable functions $f:\mathcal{X}\to\mathbb{R}$. For $f\in\mathcal{F}$, we seek an estimator $\widehat{\mathcal{I}}_n(f)$ of
$\mathcal{I}(f):=\int_{\mathcal{X}} f\,\mathrm{d}\mu$
using only $n$ evaluations of $f$. Examples include numerical quadrature, gradient estimation using minibatches in stochastic gradient descent (SGD), and the construction of small weighted subsets that uniformly approximate a family of objective functions, known as coresets. For square-integrable observables, independent sampling provides a simple and widely applicable approach, with root mean square error (RMSE) of order $n^{-1/2}$. This general rate, however, does not exploit the regularity of the observable or the geometry of the underlying space. A natural question is whether these additional structures can improve approximation accuracy for the same evaluation budget.

Determinantal point processes (DPPs) \cite{Mac75, HKPV06, Lyo03, Sos00} provide a kernel-based approach to this question~\cite{KuTa12}. Their negative dependence can reduce redundancy among selected points and yield more accurate integration estimates~\cite{bardenet2020monte, dpp-sgd, bardenet2024small, tran2025negative}, while their algebraic structure allows fluctuations to be studied through the variance of linear statistics. Kernel design plays a central role: quantitative improvements depend on how the sampling distribution interacts with the observables of interest. Recent work has developed two complementary approaches for DPP kernel designs: wavelet DPPs in Euclidean spaces exploit observable regularity, including limited smoothness~\cite{tran2026state}; spectral DPPs built from diffusion operators provide intrinsic constructions on manifolds, point clouds, and networks, with guarantees that reflect the underlying geometry~\cite{tran2026determinantals}.

In this work, we bring these two approaches together through localized spectral constructions on manifolds and geometric graphs. Spectral wavelets and heat kernels provide localized features, from which we construct finite-rank orthogonal projections defining the DPP kernels. The central challenge is to retain sufficient localization when passing from these features to an orthogonal projection. We use this localization to establish variance reduction and approximation rates that adapt to observable regularity. The resulting constructions provide a common approach to sampling in continuous and discrete settings, with rates governed by intrinsic dimension and regularity.


\section{Spectral wavelet DPPs on manifolds}\label{sec:manifolds}
Let $(\mathcal{M},g)$ be a smooth, connected, closed Riemannian
manifold of dimension $m$. Set
$\mu:=\vol_g/\vol_g(\mathcal{M}),$
and write $d$ for the geodesic distance.
For a function $f:\mathcal{M}\to\mathbb{R}$ and $0<s\le1$, define
\begin{equation}\label{eq:s-holder}
[f]_{C^{0,s}}
:=
\sup_{x\ne y}\frac{|f(x)-f(y)|}{d(x,y)^s}.
\end{equation}
We also define
\begin{equation}\label{eq:s-sobolev}
[f]_{H^s}
:=
\begin{cases}
\displaystyle
\left(
\iint_{\{x\ne y\}}
\frac{|f(x)-f(y)|^2}{d(x,y)^{m+2s}}
\,\mathrm{d}\mu(x)\,\mathrm{d}\mu(y)
\right)^{1/2},
&0<s<1,\\[1.2em]
\displaystyle
\left(
\int_{\mathcal{M}}|\nabla f|_g^2\,\mathrm{d}\mu
\right)^{1/2},
&s=1.
\end{cases}
\end{equation}
We denote by $C^{0,s}(\mathcal{M})$ the class of functions with
finite seminorm \eqref{eq:s-holder}, and by $H^s(\mathcal{M})$
the space of functions in $L^2(\mu)$ with finite seminorm
\eqref{eq:s-sobolev}.

Let $\Delta_{\mathcal{M}}$ be the Laplace--Beltrami operator and
set $L:=-\Delta_{\mathcal{M}}$. Fix a real orthonormal eigenbasis
$\{\phi_k\}_{k\ge1}$ of $L^2(\mu)$ such that $\phi_1=1$ and
\[
L\phi_k=\lambda_k\phi_k,
\qquad
0=\lambda_1\le\lambda_2\le\cdots.
\]
A \emph{spectral filter} is a bounded continuous function
$u:[0,\infty)\to\mathbb{R}$.
Throughout, we consider filters $u\in C^\infty([0,\infty))$
satisfying
\begin{equation}\label{eq:spec-filter}
\sup_{t\ge0}(1+t)^p|u^{(q)}(t)|<\infty
\qquad
\text{for all integers }p,q\ge0.
\end{equation}
For such a filter, the operator $u(h^2L)$ has a smooth integral
kernel with respect to $\mu$, given by
\begin{equation*}
W_{u,h}(x,y)
:=
\sum_{k=1}^{\infty}
u(h^2\lambda_k)\phi_k(x)\phi_k(y),
\qquad h>0.
\end{equation*}
For each fixed $h>0$, the series
converges uniformly.

\begin{remark}[Relation to spectral wavelets]
For a nonzero filter $u$ satisfying \eqref{eq:spec-filter} and
$u(0)=0$, the kernels $W_{u,h}$ are spectral wavelets on
$\mathcal{M}$ in the sense of \cite{geller-mayeli}.
\end{remark}

\subsection{Spectral wavelet DPPs and variance reduction}

Fix a localization scale $h>0$, a finite set of centers
$\mathcal{Z}=\{z_1,\ldots,z_I\}\subset\mathcal{M}$, and a finite
family of spectral filters $\mathcal{U}=\{u_1,\ldots,u_J\}$
satisfying \eqref{eq:spec-filter}. Define
\[
\psi_{i,j}(x)
:=
h^{m/2}W_{u_j,h}(x,z_i),
\qquad
1\le i\le I,\quad 1\le j\le J,
\]
and
\[
H_h
:=
\operatorname{span}
\{\psi_{i,j}:1\le i\le I,\ 1\le j\le J\}
\subset L^2(\mu).
\]
Let $\Pi_h$ be the orthogonal projection from $L^2(\mu)$ onto
$H_h$, and denote its integral kernel with respect to $\mu$
by $K_h$, so that
\[
\Pi_h f(x)
=
\int_{\mathcal{M}}K_h(x,y)f(y)\,\mathrm{d}\mu(y).
\]
Since $\Pi_h$ is a finite-rank orthogonal projection,
$(K_h,\mu)$ defines a DPP with exactly
$n_h:=\dim H_h$ points almost surely.

\begin{definition}
The DPP $\mathcal{S}_h$ with kernel $K_h$ and reference measure
$\mu$ is called the \emph{spectral wavelet DPP} associated with
$(\mathcal{Z},\mathcal{U})$ at localization scale $h$.
\end{definition}

Our main result bounds the variance of the linear statistic
\[
\mathcal{S}_h(f)
:=
\sum_{x\in\mathcal{S}_h}f(x).
\]
The choice of centers is specified through the following
separation condition.

\begin{definition}[Separated centers]
Let $r>0$. A set
$\mathcal{Z}=\{z_1,\ldots,z_I\}\subset\mathcal{M}$
is \emph{$r$-separated} if $d(z_i,z_j)>r$ whenever $i\ne j$.
It is \emph{maximally $r$-separated} if it is $r$-separated
and no $r$-separated set properly contains it.
\end{definition}

Now we can state our main theorem:

\begin{theorem}[Variance reduction of spectral wavelet DPPs]
\label{thm:var-reduce-manifold}
Fix a finite family $\mathcal{U}=\{u_1,\ldots,u_J\}$ of real
spectral filters satisfying \eqref{eq:spec-filter}, with at
least one nonzero $u_j$. There exists
$a_0=a_0(\mathcal{M},g,\mathcal{U})\ge1$ such that, for every
fixed $a\ge a_0$, there is
$h_0=h_0(\mathcal{M},g,\mathcal{U},a)>0$ with the following
property.

For $0<h\le h_0$, let
$\mathcal{Z}=\{z_1,\ldots,z_I\}$ be any maximally
$ah$-separated set, and let $\mathcal{S}_h$ be the spectral
wavelet DPP associated with $(\mathcal{Z},\mathcal{U})$
at scale $h$. Then
$n_h\asymp h^{-m}$. Moreover, 
for every $0<s\le1$, 
\begin{equation*}
\var\bigl[\mathcal{S}_h(f)\bigr]
\lesssim
\begin{cases}
n_h^{1-2s/m}[f]_{C^{0,s}}^2,
&f\in C^{0,s}(\mathcal{M}),\\
n_h^{1-2s/m}[f]_{H^s}^2,
&f\in H^s(\mathcal{M}).
\end{cases}
\end{equation*}
The implicit constants depend only on
$(\mathcal{M},g)$, $\mathcal{U}$, $a$, and $s$.
\end{theorem}

\subsection{Application to quadrature}

Let $f$ belong to either $C^{0,s}(\mathcal{M})$ or
$H^s(\mathcal{M})$, where $0<s\le1$. We aim to approximate
\[
\mathcal{I}(f)
:=
\int_{\mathcal{M}}f(x)\,\mathrm{d}\mu(x).
\]
Our quadrature rules use the spectral wavelet DPP associated
with the filters $\mathcal{U}=\{u_1,u_2\}$, where
\begin{equation}\label{eq:heat-filters}
u_1(t)=e^{-t}-e^{-2t},
\qquad
u_2(t)=e^{-2t}-e^{-4t}=u_1(2t).
\end{equation}
For this fixed pair, we will show that the constants
$a_0=a_0(\mathcal{M},g)$ and
$h_0=h_0(\mathcal{M},g,a)$ in
Theorem~\ref{thm:var-reduce-manifold} can be chosen so that
\[
c n_h\le K_h(x,x)\le C n_h,
\qquad x\in\mathcal{M},
\]
for every maximally $ah$-separated set whenever
$a\ge a_0$ and $0<h\le h_0$.
Here $c,C>0$ depend only on $(\mathcal{M},g)$ and $a$.
This estimate is established in the proof of
Theorem~\ref{thm:quadrature-manifold} below and ensures that
the quadrature weights are well defined.

Fix $a$ and $h$ in this regime, and let
$\mathcal{Z}=\{z_1,\ldots,z_I\}$ be a maximally
$ah$-separated set. Let $V_1,\ldots,V_I$ be its measurable
geodesic Voronoi cells, with ties assigned to the smallest
index. Define
$\iota:\mathcal{M}\to\{1,\ldots,I\}$ by
$\iota(x)=i$ whenever $x\in V_i$, and let $\mathcal{S}_h$
be the spectral wavelet DPP associated with
$(\mathcal{Z},\mathcal{U})$ at scale $h$.

For $f\in C^{0,s}(\mathcal{M})$, define
\begin{equation}\label{eq:est-holder}
\widehat{\mathcal{I}}_h(f)
:=
\sum_{i=1}^I\mu(V_i)f(z_i)
+
\sum_{x\in\mathcal{S}_h}
\frac{f(x)-f(z_{\iota(x)})}{K_h(x,x)}.
\end{equation}
For $f\in H^s(\mathcal{M})$, evaluations of $f$ at finitely many points $z_1,\ldots,z_I$ do not make sense. We modify our estimator as follows.
Let $U_1,\ldots,U_I$ be mutually independent and also independent of $\mathcal{S}_h$, with
$U_i\sim\mu(V_i)^{-1}\mu|_{V_i}$.
Using any measurable representative of $f$, define
\begin{equation}\label{eq:est-sobolev}
\widehat{\mathcal{I}}_h(f)
:=
\sum_{i=1}^I\mu(V_i)f(U_i)
+
\sum_{x\in\mathcal{S}_h}
\frac{f(x)-f(U_{\iota(x)})}{K_h(x,x)}.
\end{equation}
These random evaluations are unchanged almost surely if
$f$ is modified on a $\mu$-null set.

\begin{theorem}[Spectral wavelet DPPs for quadrature]
\label{thm:quadrature-manifold}
Let $\mathcal{U}=\{u_1,u_2\}$ be the filters in
\eqref{eq:heat-filters}. Fix
$a\ge a_0(\mathcal{M},g)$ and
$0<h\le h_0(\mathcal{M},g,a)$ as above.
Let $\mathcal{Z}$ be any maximally $ah$-separated set,
and let $\mathcal{S}_h$ be the spectral wavelet DPP
associated with $(\mathcal{Z},\mathcal{U})$ at scale $h$.
Then, for every $0<s\le1$ and every
$f\in C^{0,s}(\mathcal{M})$ or $f\in H^s(\mathcal{M})$,
the corresponding estimator in \eqref{eq:est-holder}
or \eqref{eq:est-sobolev} is unbiased:
\[
\mathbb{E}\bigl[\widehat{\mathcal{I}}_h(f)\bigr]
=
\mathcal{I}(f).
\]
Moreover,
\begin{equation*}
\mathbb{E}
\left|\widehat{\mathcal{I}}_h(f)-\mathcal{I}(f)\right|^2
\lesssim
\begin{cases}
n_h^{-1-2s/m}[f]_{C^{0,s}}^2,
&f\in C^{0,s}(\mathcal{M}),\\
n_h^{-1-2s/m}[f]_{H^s}^2,
&f\in H^s(\mathcal{M}).
\end{cases}
\end{equation*}
The implicit constants depend only on
$(\mathcal{M},g)$, $a$, and $s$. In particular, they are uniform in $h$,
the center set $\mathcal{Z}$, and $f$. 
\end{theorem}

We briefly comment on the number of function evaluations required
by these estimators. In the proof of Theorem \ref{thm:quadrature-manifold}, we will show that $n_h=2I$.
Therefore, each estimator requires at most
$n_h+I=3n_h/2$
evaluations of $f$: one at each DPP point and one at each
center $z_i$ or auxiliary point $U_i$.
The proof of Theorem~\ref{thm:quadrature-manifold} also gives
an explicit expression for $K_h(x,x)$, which specifies the
weights in \eqref{eq:est-holder} and \eqref{eq:est-sobolev}.

\begin{remark}[Optimality of the quadrature rate]
It is interesting that our estimator attains the optimal rate over the class of Sobolev functions $H^s$ with $0<s\le 1$. Indeed,
in the classical setting of randomized quadrature on an
$m$-dimensional Euclidean cube, the optimal root mean square
error (RMSE) over the unit ball of $H^s$ is $n^{-1/2-s/m}$, where $n$
is the number of function evaluations. Our estimator achieves the same rate on
closed Riemannian manifolds: Theorem~\ref{thm:quadrature-manifold}
gives an RMSE of order $n_h^{-1/2-s/m}$
using at most $3n_h/2$ function evaluations.
\end{remark}

\section{Spectral wavelet DPPs on geometric graphs}\label{sec:graphs}
Let $(\M,g)$ be a fixed smooth, compact, connected Riemannian manifold
without boundary, of dimension $m\geq1$, isometrically embedded in $\R^d$.
We identify $\M$ with its image and take $N\geq2$. Let $d_{\M}$ be its geodesic distance and
let $\mu=\vol_g/\vol_g(\M)$ be the normalized volume-form. For $N \geq 2$, we consider the dataset $\mathcal{X} :=\{x_1,\ldots,x_N\} \subset \M$, where 
\begin{equation}\label{eq:model}
 x_1,\ldots,x_N\stackrel{\mathrm{i.i.d.}}{\sim}\nu=\rho\mu,
 \text{ where } 0<\rho_{\min}\leq\rho\leq\rho_{\max}<\infty
 \ \ \mu\text{-a.s}.
\end{equation}
In particular, the points are distinct almost surely. We equip $\X$ with the empirical measure
$\mu_N=N^{-1}\sum_i\delta_{x_i}$ and the inner product
\begin{equation*}
 \langle u,v\rangle_{L^2(\mu_N)}
 :=\frac1N\sum_{i=1}^N u(x_i)v(x_i).
\end{equation*}

\subsection{Graph constructions}
Taking $\mathcal{X}$ as the vertex set, we consider the following two graph constructions, where the set of edges is fully characterized by the adjacency matrix $A$:

\medskip
\noindent
\textbf{$\eps$-graphs:}
For $\eps>0$, define
\begin{equation}\label{eq:epsilon-adjacency}
 A_{i,j} = A_{ij}^{\eps}:=\one_{\{|x_i-x_j|\leq\eps\}},
 \quad 1\leq i,j\leq N.
\end{equation}
\\
\textbf{Symmetric $k$-NN graphs:}
Let $1\leq k<N$. For each $i$, let $\mathcal N_k(i)$ be the set of the
$k$ nearest vertices to $x_i$ (excluding $x_i$), with ties resolved by label order.
Let $R_i$ be the ambient Euclidean distance to the $k$th nearest other vertex. Define
\begin{equation}\label{eq:knn-adjacency}
 A_{ij} = A_{ij}^{k}:=
 \begin{cases}
  1,&i=j,\text{ or }j\in\mathcal N_k(i),\text{ or }i\in\mathcal N_k(j),\\
  0,&\text{otherwise}.
 \end{cases}
\end{equation}
\\
For either construction, we set
\begin{equation}\label{eq:graph-scale}
 r:=\begin{cases}\eps,&A=A^{\eps},\\ (k/N)^{1/m},&A=A^k,\end{cases}
 \quad K:=Nr^m,
 \quad d_i:=\sum_j A_{ij},\quad D:=\diag(d_1,\ldots,d_N).
\end{equation}

Finally, we define the rescaled symmetric normalized graph Laplacian
and its associated heat semigroup by
\begin{equation}\label{eq:laplacian}
    \Lcal_N:=r^{-2}(I-D^{-1/2}AD^{-1/2}),\qquad
    T_t:=e^{-t\Lcal_N},\quad t\geq 0.
\end{equation}
The matrix $\Lcal_N$ is positive semidefinite and self-adjoint
with respect to the inner product $\langle \cdot,\cdot\rangle_{L^2(\mu_N)}$.

We equip the graph with a shortest-path distance. For $i\ne j$, we assign each edge
that joins $x_i,x_j$ the length $|x_i-x_j|$,
and we denote the resulting shortest-path distance by $d_N(i,j)$, with the convention that
$d_N(i,j)=\infty$ if $i$ and $j$ belong to different connected
components. 
We write $B_N(i,R):=\{j:d_N(i,j)\leq R\}$
for the corresponding closed balls. Note that by the triangle inequality
in $\R^d$, we have
\begin{equation}\label{eq:chord-graph}
    |x_i-x_j|\leq d_N(i,j).
\end{equation}

\begin{remark}[Convention on constants]\label{rem:constants}
Throughout this section, $(\M,g)$, its embedding, $\rho_{\min}$, and
$\rho_{\max}$ are fixed. We call a constant structural if it depends only
on these data. The notation $a\lesssim b$ means $a\leq Cb$
for a structural constant $C>0$, and $a\asymp b$ means both
$a\lesssim b$ and $b\lesssim a$. Additional dependence on a fixed
parameter is indicated by a subscript. Constants may change from one
display to the next.
\end{remark}

\subsection{Determinantal processes from heat columns}\label{sec:construction}
We now construct a determinantal point process on either graph
using suitably selected columns of the heat semigroup.
Fix $a\geq 1$ and $h>0$, and let
$Z=\{z_1,\ldots,z_n\}$ be a maximal strictly
$ah$-separated set of vertices with respect to $d_N$.
Such a set can be constructed greedily by examining the
vertices in increasing label order and retaining each vertex
whose distance from all previously selected vertices exceeds
$ah$. 
By the definition of a maximal $ah$-separated set, we must have
\begin{equation}\label{eq:centers}
 d_N(z_\ell,z_j)>ah\quad \forall \ell\ne j,\text{ and }
 \min_{1\leq\ell\leq n}d_N(i,z_\ell)\leq ah \quad \forall 1\leq i\leq N.
\end{equation}

Let $e_i$ denote the $i$th coordinate vector in $\R^N$.
We consider the heat-kernel columns associated with the
selected vertices $Z$ and define
\begin{equation}\label{eq:projection}
    E_Z:=(e_{z_1},\ldots,e_{z_n}),\quad
    B_h:=T_{h^2}E_Z,\quad
    \Pi_h:=B_h(B_h^*B_h)^{-1}B_h^*.
\end{equation}
Since $T_{h^2}$ is invertible, the columns of $B_h$ are
linearly independent, and hence $\Pi_h$ is the rank-$n$
orthogonal projection onto their span. Moreover, by the
semigroup property,
\begin{equation}\label{eq:gram-identity}
    B_h^*B_h=E_Z^*T_{2h^2}E_Z.
\end{equation}
The integral kernel of $\Pi_h$ with respect to the empirical
measure $\mu_N$ is given by
\begin{equation}\label{eq:integral-kernel}
    \Kcal_{h,N}(x_i,x_j):=N(\Pi_h)_{ij}.
\end{equation}
\\
Conditional on $\X$, let $\Sset_{h,N}$ be the projection DPP with projection matrix
$\Pi_h$. Thus, $\Sset_{h,N}$ is a random subset of $\{1,\ldots,N\}$
of cardinality $n$, with distribution
\begin{equation}\label{eq:dpp-law}
    \P(\Sset_{h,N}=J\mid\X)=\det((\Pi_h)_{J,J}),
    \quad J\subset\{1,\ldots,N\},\quad |J|=n.
\end{equation}
For each vertex $i$, we denote its inclusion probability by
\[
    \pi_i:=\P(i\in\Sset_{h,N}\mid\X)=(\Pi_h)_{ii}.
\]
Given a function $f:\X\to\R$, we consider the associated
linear statistic
\begin{equation}\label{eq:statistic}
    \Sset_{h,N}(f):=\sum_{i\in\Sset_{h,N}}f(x_i).
\end{equation}
Note that $\Kcal_{h,N}(x_i,x_i)=N\pi_i$.
\\
Next, for $0<s\leq1$ and $f:\X\to\R$, we define
\begin{equation}\label{eq:holder}
 [f]_{s,\X}:=\max_{i\ne j}
       \frac{|f(x_i)-f(x_j)|}{|x_i-x_j|^s}.
\end{equation}
The estimates below hold simultaneously for all functions satisfying
\eqref{eq:holder}. In particular, they apply to restrictions of
$f:\M\to\R$ with
\begin{equation}\label{eq:holder-manifold}
 [f]_{s,\M}:=\sup_{x\ne y\in\M}
       \frac{|f(x)-f(y)|}{|x-y|^s}<\infty.
\end{equation}

\subsection{Volume and heat-kernel estimates}\label{sec:heat-results}
We now establish the geometric and heat-kernel estimates needed to analyze the projection $\Pi_h$ constructed above. We show that, with high probability, both graph constructions have uniformly controlled degrees, graph distances comparable to the geodesic distance on $\M$, and suitable volume bounds
for graph-distance balls. On the same event, we obtain off-diagonal decay estimates for the heat semigroup, together with diagonal and near-diagonal lower bounds. The following
two theorems make these estimates precise for $\eps$-graphs
and symmetric $k$-nearest-neighbour graphs, respectively.

\begin{theorem}\label{thm:heat-epsilon}
Assume \eqref{eq:model}, and let $A=A^{\eps}$ in \eqref{eq:laplacian}.
There exist positive structural constants $r_0,R_0,C_\delta,c,C$ such
that the following holds: 

Suppose that
\begin{equation}\label{eq:epsilon-regime}
 0<\eps\leq r_0,\text{ and } K=N\eps^m\geq C\log N.
\end{equation}
Then, there exists an event $\Gcal_N^{\eps}$ that occurs with probability 
\begin{equation}\label{eq:epsilon-probability}
 \P(\Gcal_N^{\eps})\geq1-CN e^{-cK}
\end{equation}
on which the graph is connected and
\begin{align}
 &c\,d_{\M}(x_i,x_j)\leq d_N(i,j)\leq C\,d_{\M}(x_i,x_j) \text{  and  } d_i \asymp K,
 \label{eq:graph-geometry}\\
 &\mu_N(B_N(i,R))\leq C(R+\delta_N)^m\quad \text{for } \delta_N:=C_\delta\eps \text{ and }\forall R>0,
 \label{eq:graph-volume-upper}\\
 &\mu_N(B_N(i,R))\geq cR^m\quad \text{for }\delta_N\leq R\leq R_0.
 \label{eq:graph-volume-lower}
\end{align}
Moreover, for every fixed $A_0>0$ and $a\geq1$, there exist positive
constants $C_*,h_0,C_{A_0},c_a$, depending only on $A_0$ and $a$ and 
the structural data, such that on the event $\Gcal_N^{\eps}$,
\begin{align}
 &0\leq(T_\tau)_{ij}
 \leq\frac{C_{A_0}}{Nh^m}
       \left(1+\frac{d_N(i,j)}h\right)^{-A_0},
       \quad \forall\tau\in\{h^2,2h^2\},\label{eq:heat-upper}\\
 &(T_{2h^2})_{ii}\geq\frac{c}{Nh^m},\label{eq:heat-diagonal}\\
 &(T_{h^2})_{ij}\geq\frac{c_a}{Nh^m}, \quad
       \forall i,j \text{ with } d_N(i,j)\leq ah,\label{eq:heat-lower}
\end{align}
hold simultaneously for all $h$ satisfying
\begin{equation}\label{eq:heat-window}
 C_*\eps^2\log(eK)\leq h^2\leq h_0^2.
\end{equation}
Furthermore, the constants can be chosen so that $h\geq\delta_N$, $ah\geq2\delta_N$,
and $ah\leq R_0$. The constant in \eqref{eq:heat-upper} does not depend
on $a$; the lower bound of \eqref{eq:heat-window} may depend on $a$.
\end{theorem}

\begin{theorem}\label{thm:heat-knn}
Assume \eqref{eq:model}, and let $A=A^k$ in \eqref{eq:laplacian}.
Then, there exist positive structural constants $r_0,c,C$ such that, if
\begin{equation}\label{eq:knn-regime}
 C\log N\leq k<N,\quad r=(k/N)^{1/m}\leq r_0,
\end{equation}
then there is an event $\Gcal_N^k$ with
\begin{equation}\label{eq:knn-probability}
 \P(\Gcal_N^k)\geq1-CN e^{-ck}
\end{equation}
on which \eqref{eq:graph-geometry}--\eqref{eq:heat-lower} hold with
$K=k$, $\delta_N=C_\delta r$, and
\begin{equation}\label{eq:knn-heat-window}
 C_*r^2\log(ek)\leq h^2\leq h_0^2.
\end{equation}
The dependence of the constants is the same as in
Theorem~\ref{thm:heat-epsilon}.
\end{theorem}

Furthermore, we shall give explicit choices of parameters in Corollary~\ref{cor:parameters} so that the intervals \eqref{eq:heat-window} and \eqref{eq:knn-heat-window}  are non-empty.

\subsection{Variance reduction on geometric graphs}\label{sec:variance-results}

\begin{theorem}\label{thm:projection}
Fix $B>m+2$ and $A_0>B+m$. There exists a sufficiently
large $a\geq 1$, depending only on $A_0$, $B$, and the
structural data, such that the following holds:

For either graph construction, let $Z$ and $\Pi_h$ be
defined by \eqref{eq:centers}--\eqref{eq:projection}
with this choice of $a$. On the corresponding event
$\Gcal_N^\eps$ or $\Gcal_N^k$, for every $h$ satisfying
\eqref{eq:heat-window} or \eqref{eq:knn-heat-window},
respectively, we have
\begin{align}
    n&\asymp h^{-m},
    \label{eq:rank-bound}\\
    |(\Pi_h)_{ij}|
    &\leq \frac{C}{Nh^m}
    \left(1+\frac{d_N(i,j)}{h}\right)^{-B},
    \label{eq:projection-decay}\\
    c\frac{n}{N}
    &\leq \pi_i\leq C\frac{n}{N},
    \quad 1\leq i\leq N.
    \label{eq:intensity-bound}
\end{align}
Moreover, for every $0<s\leq 1$ and every
$f:\X\to\R$, the associated linear statistic satisfies
\begin{equation}\label{eq:raw-variance}
    \frac{1}{n^2}
    \var[\Sset_{h,N}(f)\mid\X]
    \leq C_s[f]_{s,\X}^2n^{-1-2s/m}.
\end{equation}
The constants depend on $A_0, B, s$ and
the structural data, but are independent of $N$, $h$, and $\mathcal{X}$.
\end{theorem}

\begin{remark}\label{rem:rank}
In the construction \eqref{eq:projection}, the heat scale $h$
is prescribed, whereas the rank $n=|Z|$ is determined by the
choice of the maximal $ah$-separated set and may depend on
the realization of $\X$. Nevertheless, \eqref{eq:rank-bound}
provides uniform control of the rank on the stated events.
In particular, we do not require that every prescribed rank
can be attained by varying $h$.
\end{remark}

\begin{corollary}\label{cor:parameters}
Fix $q>0$. There exist constants $K_0,C_*>0$ and
$N_0\in\N$, depending only on $q$ and the fixed data
of Theorem~\ref{thm:projection}, such that the following
choices are admissible for every $N\geq N_0$:
\begin{align}
    &\eps_N:=\left(\frac{K_0\log N}{N}\right)^{1/m},
    \quad
    h_N:=\eps_N\sqrt{C_*\log(eK_0\log N)},
    \label{eq:explicit-epsilon}
    \\
    &k_N:=\lceil K_0\log N\rceil,
    \quad
    h_N:=\left(\frac{k_N}{N}\right)^{1/m}
    \sqrt{C_*\log(ek_N)}.
    \label{eq:explicit-knn}
\end{align}
For the $\eps$-graph and the symmetric $k$-nearest-neighbour
graph, respectively, the corresponding choices above ensure
that, with probability at least $1-N^{-q}$, the conclusions
\eqref{eq:rank-bound}--\eqref{eq:raw-variance} hold, with
\begin{equation}\label{eq:rank-explicit}
    n\asymp_q
    \frac{N}{\log N(\log\log N)^{m/2}}.
\end{equation}
The estimates \eqref{eq:rank-bound}--\eqref{eq:raw-variance} remain valid for any $h_N\leq h\leq h_0$, with the corresponding rank satisfying $n\asymp h^{-m}$.
\end{corollary}

\subsection{Unbiased estimators of the empirical average}\label{sec:estimator}
The variance estimate \eqref{eq:raw-variance} concerns the linear statistic $\Sset_{h,N}(f)$, whose normalized version
need not be an unbiased estimator of the empirical average, since the inclusion probabilities $\pi_i$ are not necessarily constant. We now construct a conditionally unbiased estimator that achieves the same variance decay rate while requiring at most $2n$ function evaluations. 

Our estimator is based on a control variate obtained from a
piecewise constant approximation of $f$. To construct this
approximation, we assign each vertex to its nearest center $z_{\ell}$
in $Z$ with respect to $d_N$, resolving ties by the order
of the centers. This defines a partition $C_1,\ldots,C_n$
of the vertex set. We then define
\begin{equation}\label{eq:approximation}
\begin{aligned}
    A_hf(x_i)&:=f(x_{z_\ell}),\quad i\in C_\ell,\\
    I_N(A_hf)&:=\sum_{\ell=1}^n
    \frac{|C_\ell|}{N}f(x_{z_\ell}).
\end{aligned}
\end{equation}
Using this approximation as a control variate, we define
the estimator
\begin{equation}\label{eq:estimator}
    \widehat I_{h,N}(f):=I_N(A_hf)
    +\frac{1}{N}\sum_{i\in\Sset_{h,N}}
    \frac{f(x_i)-A_hf(x_i)}{\pi_i}.
\end{equation}
The estimator requires at most $2n$ evaluations of $f$,
since values at vertices belonging to both the set of
centers and the determinantal sample can be reused.

\begin{theorem}\label{thm:estimator}
Under the hypotheses of Theorem~\ref{thm:projection},
on the corresponding high-probability event, the estimator
$\widehat I_{h,N}(f)$ is conditionally unbiased for
$I_N(f)$. More precisely, for every $0<s\leq 1$
and every $f:\X\to\R$, we have
\begin{align}
    \E[\widehat I_{h,N}(f)\mid\X]
    &=I_N(f),\label{eq:unbiased}\\
    \E\big[|\widehat I_{h,N}(f)-I_N(f)|^2\mid\X\big]
    &\leq C_s[f]_{s,\X}^2n^{-1-2s/m}.
    \label{eq:mse}
\end{align}
The estimator requires at most $2n$ evaluations of $f$.
In particular, for either choice of graph parameters in
Corollary~\ref{cor:parameters}, these conclusions hold
with probability at least $1-N^{-q}$.
\end{theorem}

\begin{proposition}\label{prop:population}
Let $f:\M\to\R$ be bounded, measurable, and fixed independently of the
cloud. Suppose $\widehat I_N$ is square integrable and satisfies
$\E[\widehat I_N\mid\X]=I_N(f)$ almost surely. Then
\begin{equation}\label{eq:population}
 \E\left[\left|\widehat I_N-\int_{\M}f\dd\nu\right|^2\right]
 =\E\big[\var(\widehat I_N\mid\X)\big]
       +\frac{\var_{\nu}(f)}{N}.
\end{equation}
\end{proposition}
Thus \eqref{eq:mse} estimates the additional error from subsampling the
observed cloud. The approximation of $\int_{\M}f\dd\nu$ also includes
the second term in \eqref{eq:population}. The event in
Theorem~\ref{thm:estimator} concerns the first term conditionally; an
unconditional bound additionally requires control on its complement.

\subsection{Central limit theorems}\label{clt:sec:results}

We next consider the conditional distributions of $\Sset_{h,N}(f)$ and
$\widehat I_{h,N}(f)$ for a fixed function on $\M$. For a nonconstant
$f\in C^1(\M)$, we prove matching variance lower bounds and central
limit theorems for both statistics. The limits hold conditionally on
almost every realization of the point cloud.

Write
\begin{equation}\label{clt:eq:variances}
 V_{h,N}(f):=\var[\Sset_{h,N}(f)\mid\X],\qquad
 W_{h,N}(f):=\var[\widehat I_{h,N}(f)\mid\X],
\end{equation}
and let $\Phi$ denote the distribution function of $\mathcal N(0,1)$.

\begin{theorem}\label{clt:thm:main}
Fix $q>1$, and choose either graph construction with the graph
parameters in Corollary~\ref{cor:parameters}. There is a sufficiently
large fixed choice of $a$ in Theorem~\ref{thm:projection}, depending
only on $A_0$, $B$, and the structural data, such that the following
holds. Choose $C_*,h_0$ for this value of $a$, and let $h=h(N)$ be a
deterministic sequence satisfying \eqref{eq:heat-window} or
\eqref{eq:knn-heat-window}, respectively, with $h(N)\to0$.

Let $f\in C^1(\M)$ be nonconstant and fixed independently of the
cloud. For almost every realization of the infinite i.i.d. sequence
in \eqref{eq:model}, and all sufficiently large $N$,
\begin{equation}\label{clt:eq:variance-orders}
 \frac{V_{h,N}(f)}{n^2}
 \asymp_f W_{h,N}(f)
 \asymp_f \frac{h^2}{n}
 \asymp_f n^{-1-2/m}.
\end{equation}
Moreover, for every $t\in\R$, almost surely,
\begin{align}
 \P\left(
   \frac{\Sset_{h,N}(f)-\sum_{i=1}^N\pi_i f(x_i)}
        {\sqrt{V_{h,N}(f)}}\leq t\,\middle|\,\X
 \right)&\longrightarrow\Phi(t),\label{clt:eq:usual}\\
 \P\left(
   \frac{\widehat I_{h,N}(f)-I_N(f)}
        {\sqrt{W_{h,N}(f)}}\leq t\,\middle|\,\X
 \right)&\longrightarrow\Phi(t).\label{clt:eq:residual}
\end{align}
The constants in \eqref{clt:eq:variance-orders} may depend on $f$
and the fixed parameters, but not on $N$, $h$, or the realized cloud.
\end{theorem}

The conclusions hold in every dimension $m\geq1$. In particular,
they apply to the heat scales in Corollary~\ref{cor:parameters}.
Equation~\eqref{clt:eq:usual} is also the central limit theorem for
$n^{-1}\Sset_{h,N}(f)$, centered at
$n^{-1}\sum_i\pi_i f(x_i)$. This conditional mean need not equal
$I_N(f)$, whereas \eqref{clt:eq:residual} is centered at $I_N(f)$
by \eqref{eq:unbiased}.

\section{Haar DPPs on geometric graphs}\label{sec:block-dpp}
We next consider DPPs constructed from block projections
onto the normalized indicators of a partition of the vertex
set. These indicators are precisely the corresponding graph
Haar scaling functions.

Let $\X=\{x_1,\ldots,x_N\}\subset\M$. We first consider
a partition $C_1,\ldots,C_n$ of the vertex set.
In Subsection~\ref{sec:partition}, we give an explicit
construction of such a partition using $d_N$, valid for
any dataset. Under \eqref{eq:model}, we then establish
bounds on the cardinalities and diameters of the
resulting cells.

\subsection{The projection associated with a partition}

Let $C_1,\ldots,C_n$ be a partition of $\{1,\ldots,N\}$ into nonempty
sets, chosen deterministically given $\X$.
For $1\leq\ell\leq n$, write
\begin{equation}\label{blk:eq:cells}
 M_\ell:=|C_\ell|,\quad w_\ell:=\frac{M_\ell}{N},\quad
 \Delta_\ell:=\max_{i,j\in C_\ell}d_{\M}(x_i,x_j).
\end{equation}
The vectors $\phi_\ell:=M_\ell^{-1/2}\one_{C_\ell}$ are orthonormal
in $\R^N$ with its Euclidean inner produc, and the orthogonal projection
onto their span is given by
\begin{equation}\label{blk:eq:projection}
 P_{\mathrm{sc}}:=\sum_{\ell=1}^n\phi_\ell\phi_\ell^*,\qquad
 (P_{\mathrm{sc}})_{ij}=
 \begin{cases}
 M_\ell^{-1},& i,j\in C_\ell,\\
 0,& i,j\text{ in distinct sets}.
 \end{cases}
\end{equation}
Thus $P_{\mathrm{sc}}$ has rank $n$, and its integral kernel with
respect to $\mu_N$ takes the value $N(P_{\mathrm{sc}})_{ij}$ at
$(x_i,x_j)$.

Conditional on $\X$, let $\Sset_{\mathrm{bl}}$ be the DPP with
projection $P_{\mathrm{sc}}$. For $|J|=n$, its conditional probability
of selecting $J$ is $\det((P_{\mathrm{sc}})_{J,J})$.

\begin{proposition}\label{blk:prop:law}
The set $\Sset_{\mathrm{bl}}$ contains exactly one vertex $U_\ell$
from each $C_\ell$. Conditional on $\X$, the variables
$U_1,\ldots,U_n$ are independent and $U_\ell$ is uniform on $C_\ell$.
In particular,
\begin{equation}\label{blk:eq:law}
 \begin{aligned}
 \P(U_1=u_1,\ldots,U_n=u_n\mid\X)
    &=\prod_{\ell=1}^n M_\ell^{-1} &&(u_\ell\in C_\ell),\\
 \pi_i:=\P(i\in\Sset_{\mathrm{bl}}\mid\X)
    &=M_\ell^{-1} &&(i\in C_\ell).
 \end{aligned}
\end{equation}
If every $M_\ell$ is even, split $C_\ell=C_\ell^+\sqcup C_\ell^-$
into two sets of size $M_\ell/2$. The Haar details
$\psi_\ell:=M_\ell^{-1/2}(\one_{C_\ell^+}-\one_{C_\ell^-})$
give a second projection $P_{\mathrm H}:=\sum_{\ell=1}^n
\psi_\ell\psi_\ell^*$. Its DPP also satisfies \eqref{blk:eq:law}.
\end{proposition}

\subsection{Constructing the partition on geometric graphs}\label{sec:partition}

We now construct a partition of the vertex set for both
the $\varepsilon$-graph \eqref{eq:epsilon-adjacency} and the
symmetric $k$-nearest-neighbour graph
\eqref{eq:knn-adjacency}, with edge lengths $|x_i-x_j|$
and shortest-path distance $d_N$.

We first select a set of centers. Fix $h>0$, and examine
the vertices in label order, retaining each vertex whose
distance from every previously retained vertex exceeds $h$.
Denote the retained vertices, in order, by
$z_1,\ldots,z_n$. Then
\begin{equation}\label{blk:eq:centers}
 d_N(z_\ell,z_j)>h\quad(\ell\ne j),\qquad
 \min_{1\leq\ell\leq n}d_N(i,z_\ell)\leq h\quad(1\leq i\leq N).
\end{equation}
We may construct the partition in two ways: the
\emph{first-cover partition} and the \emph{Voronoi partition}.
For the first-cover partition, assign $i$ to $C_\ell$
for the least $\ell$ such that $i\in B_N(z_\ell,h)$.
For the Voronoi partition, assign $i$ to its nearest
$z_\ell$ in $d_N$, choosing the least $\ell$ in case
of a tie. In either case,
$z_\ell\in C_\ell\subseteq B_N(z_\ell,h)$.
In particular, $(P_{\mathrm{sc}})_{ij}=0$ whenever
$d_N(i,j)>2h$.

Let $\ell_0$ and $\kappa_0$ be as in \eqref{eq:local-chord}, so that
$d_{\M}(x,y)\leq\kappa_0|x-y|$ whenever $|x-y|\leq\ell_0$.
Use $C_\delta$ and $R_0$ from Theorem~\ref{thm:heat-epsilon}, where
$R_0$ is the upper radius in \eqref{eq:graph-volume-lower}.

\begin{theorem}\label{blk:thm:geometry}
Assume \eqref{eq:model}. Fix $q>0$. There are constants $K_0,K_1>0$
and $N_0$, depending only on $q$ and the structural data, and a
structural constant $c_*\in(0,1)$, with the following property. Put
\begin{equation}\label{blk:eq:resolution}
 t_N:=\left(\frac{K_0\log N}{N}\right)^{1/m}.
\end{equation}
For $N\geq N_0$, there is an event $\Gamma_N$ of probability at least
$1-N^{-q}$ on which the following conclusions hold simultaneously for
\begin{equation}\label{blk:eq:graph-ranges}
 t_N\leq\eps\leq\ell_0 \quad\text{in \eqref{eq:epsilon-adjacency}},
 \qquad
 K_1\log N\leq k\leq c_*N \quad\text{in \eqref{eq:knn-adjacency}}.
\end{equation}
For every $2C_\delta t_N\leq h\leq R_0$, both partitions satisfy
\begin{equation}\label{blk:eq:geometry}
 n\asymp h^{-m},\quad
 \Delta_\ell\leq 2\kappa_0 h,\quad
 w_\ell\leq \frac{C}{n}\quad(1\leq\ell\leq n).
\end{equation}
For the Voronoi partition we also have $w_\ell\geq c/n$ for every
$\ell$, and hence $\pi_i\asymp n/N$ for every $i$.
\end{theorem}

\subsection{Variance of the linear statistic}

For $f\in C^{0,s}(\M)$, define
\[
 \Sset_{\mathrm{bl}}(f):=\sum_{i\in\Sset_{\mathrm{bl}}}f(x_i)
 =\sum_{\ell=1}^n f(x_{U_\ell}),\qquad
 \bar f_\ell:=\frac{1}{M_\ell}\sum_{i\in C_\ell}f(x_i).
\]

\begin{theorem}\label{blk:thm:statistic}
For every partition in \eqref{blk:eq:cells},
$\E[\Sset_{\mathrm{bl}}(f)\mid\X]=\sum_{\ell=1}^n\bar f_\ell$ and
\begin{align}
 \var[\Sset_{\mathrm{bl}}(f)\mid\X]
 &=\frac12\sum_{\ell=1}^n\frac{1}{M_\ell^2}
       \sum_{i,j\in C_\ell}|f(x_i)-f(x_j)|^2
 \label{blk:eq:raw-exact}\\
 &\leq \frac12[f]_{C^{0,s}}^2\sum_{\ell=1}^n\Delta_\ell^{2s},
 \label{blk:eq:raw-diameter}
\end{align}
where $Delta_\ell$ is defined in \eqref{blk:eq:cells}. In particular, on $\Gamma_N$, for either partition in
Theorem~\ref{blk:thm:geometry},
\begin{equation}\label{blk:eq:raw-rate}
 \frac{1}{n^2}\var[\Sset_{\mathrm{bl}}(f)\mid\X]
 \leq C_s[f]_{C^{0,s}}^2 n^{-1-2s/m},
 \qquad 0<s\leq1,\quad f\in C^{0,s}(\M).
\end{equation}
\end{theorem}

\subsection{Unbiased estimation of the empirical average}

By \eqref{blk:eq:law}, the conditional mean of
$n^{-1}\Sset_{\mathrm{bl}}(f)$ is
$n^{-1}\sum_{\ell=1}^n\bar f_\ell$. On the other hand,
the empirical average is
\[
I_N(f)=\frac1N\sum_{i=1}^N f(x_i)
=\sum_{\ell=1}^n w_\ell\bar f_\ell.
\]
The weights in these two expressions coincide only when
the cells $C_\ell$ have equal cardinalities. To obtain
a conditionally unbiased estimator without this restriction,
we therefore define
\begin{equation}\label{blk:eq:estimator}
\widehat I_{\mathrm{bl}}(f)
:=\sum_{\ell=1}^n w_\ell f(x_{U_\ell})
=\frac1N\sum_{i\in\Sset_{\mathrm{bl}}}
\frac{f(x_i)}{\pi_i}.
\end{equation}
The estimator requires exactly $n$ evaluations of $f$.

\begin{theorem}\label{blk:thm:quadrature}
For every partition in \eqref{blk:eq:cells},
$\E[\widehat I_{\mathrm{bl}}(f)\mid\X]=I_N(f)$ and
\begin{align}
 \E\big[|\widehat I_{\mathrm{bl}}(f)-I_N(f)|^2\mid\X\big]
 &=\frac{1}{2N^2}\sum_{\ell=1}^n
        \sum_{i,j\in C_\ell}|f(x_i)-f(x_j)|^2
 \label{blk:eq:mse-exact}\\
 &\leq\frac12[f]_{C^{0,s}}^2
         \sum_{\ell=1}^n w_\ell^2\Delta_\ell^{2s}.
 \label{blk:eq:mse-diameter}
\end{align}
In particular, on $\Gamma_N$, for either partition in
Theorem~\ref{blk:thm:geometry},
\begin{equation}\label{blk:eq:mse-rate}
 \E\big[|\widehat I_{\mathrm{bl}}(f)-I_N(f)|^2\mid\X\big]
 \leq C_s[f]_{C^{0,s}}^2 n^{-1-2s/m},
 \qquad  f\in C^{0,s}(\M),\quad0<s\leq1.
\end{equation}
\end{theorem}

The bounds \eqref{blk:eq:raw-diameter} and
\eqref{blk:eq:mse-diameter} hold for any partition
$C_1,\ldots,C_n$ of a fixed point cloud. In particular,
if $\max_\ell\Delta_\ell\leq b n^{-1/m}$, then
\eqref{blk:eq:raw-rate} holds. If, in addition,
$\max_\ell w_\ell\leq B/n$, then
\eqref{blk:eq:mse-rate} holds with
$C_s=Bb^{2s}/2$.


We next use \eqref{eq:population} to bound the error of
$\widehat I_{\mathrm{bl}}(f)$ as an estimator of
$\int_{\M}f\,\dd\nu$.

\begin{corollary}\label{blk:cor:population}
Under the hypotheses of Theorem~\ref{blk:thm:geometry},
choose deterministic graph parameters satisfying
\eqref{blk:eq:graph-ranges} and
$h\in[2C_\delta t_N,R_0]$.
For every $0<s\leq1$ and every
$f\in C^{0,s}(\M)$ fixed independently of $\X$,
the estimator \eqref{blk:eq:estimator}, constructed
using either partition, satisfies
\[
0\leq
\E\left|\widehat I_{\mathrm{bl}}(f)
-\int_{\M}f\,\dd\nu\right|^2
-\frac{\var_\nu(f)}{N}
\leq C_s[f]_{C^{0,s}}^2
\left(h^{m+2s}+N^{-q}\right).
\]
\end{corollary}

\section{Computational costs}\label{sec:cost}
Let $E$ denote the number of nonzero
entries of $A$, including its diagonal. The constants in
the operation counts below may depend on the fixed
dimension $d$.

\subsection{Heat Kernel DPPs}
We distinguish between the cost of computing the projection
$\Pi_h$ and the costs incurred for each draw from
\eqref{eq:dpp-law}.
\paragraph{Preprocessing.}
We first consider the construction of $A$ and $Z$.
For the $\varepsilon$-graph
\eqref{eq:epsilon-adjacency}, we first need to compute the pairwise distances $|x_i-x_j|$, which takes $O(dN^2)$ operations, and the graph is then obtained by
comparing these distances with $\eps$. For the
symmetric $k$-nearest-neighbour graph
\eqref{eq:knn-adjacency}, selecting the $k$ nearest
vertices to each $x_i$ costs $O(N^2)$ operations
in total, while forming $A^k$ from these selections
requires an additional $O(Nk)$ operations. 

On the events of Theorems~\ref{thm:heat-epsilon}
and~\ref{thm:heat-knn}, we have $E=O(NK)$, where $K$ is defined in \eqref{eq:graph-scale}.
To construct $Z=\{z_1,\ldots,z_n\}$ in
\eqref{eq:centers}, we examine the vertices in label
order. Whenever a vertex $z_\ell$ is selected, we use
Dijkstra's algorithm to find all vertices within
distance $ah$ of $z_\ell$. These vertices are excluded
from subsequent selections. Using a binary heap, each
search costs at most $O((N+E)\log N)$ operations.
Since there are $n$ centers, the total cost is
$O(n(N+E)\log N)$.

We next consider the computation of $\Pi_h$.
Recall from \eqref{eq:projection} that
$\Pi_h$ is the orthogonal projection onto the
range of $B_h$. It is therefore sufficient to
compute an $N\times n$ matrix $Q_h$ whose columns
form an orthonormal basis of this range.
By taking a reduced QR factorization of $B_h$, we get 
\begin{equation}\label{eq:qr}
 B_h=Q_hR_h,\qquad
 \Pi_h=Q_hQ_h^*,\qquad
 \pi_i=\sum_{\ell=1}^n(Q_h)_{i\ell}^2.
\end{equation}
The factorization requires $O(Nn^2)$ operations,
and storing $Q_h$ requires $O(Nn)$ memory.
In particular, we do not need to form the
$N\times N$ matrix $\Pi_h$.

Next, we consider the cost of computing an approximation to $B_h=T_{h^2}E_Z$.
We use the expansion
\begin{equation}\label{eq:matrix-series}
 T^{(q)}:=e^{-u}\sum_{j=0}^q\frac{u^j}{j!}S^j, \quad 
 \text{ where }
 S:=D^{-1/2}AD^{-1/2},\ 
 u:=h^2/r^2.
\end{equation}
Since $\|S\|_{\op}\leq1$, for any $0<\eta<1$
the choice
\begin{equation}\label{eq:polynomial-degree}
 q:=\left\lceil2eu+2\log(1/\eta)\right\rceil
\end{equation}
satisfies
\[
\|T^{(q)}-T_{h^2}\|_{\op}\leq\eta.
\]
The proof is given in Appendix~\ref{app:numerics}.

Moreover, computing $T^{(q)}E_Z$ using the series expansion
requires $O(qEn)$ operations.
Combining this with the cost of the QR factorization
in \eqref{eq:qr}, we obtain a total cost of
\begin{equation}\label{eq:projection-cost}
 O\big((h^2/r^2+\log(1/\eta))En+Nn^2\big)
\end{equation}
for computing an approximation to $\Pi_h$, with memory
requirement $O(E+Nn)$. At the lower endpoint of
\eqref{eq:heat-window}, $h^2/r^2$ is of order $\log(eK)$.

Appendix~\ref{app:numerics} also provides
sufficient choices of $\eta$ for which the
projection computed from $T^{(q)}E_Z$
satisfies \eqref{eq:intensity-bound},
\eqref{eq:raw-variance}, and \eqref{eq:mse}.
\\\\
For the estimator $\widehat I_{h,N}(f)$ in
\eqref{eq:estimator}, the partition $C_1,\ldots,C_n$
in \eqref{eq:approximation} can be constructed by
a multi-source Dijkstra search in
$O((N+E)\log N)$ operations. Constructing $Z$
and $C_1,\ldots,C_n$ requires only $A$, the edge
lengths $|x_i-x_j|$ for $A_{ij}=1$, and the vertex
indices, without knowledge of $\M$ or $\rho$.
For a fixed function $f$, we additionally
evaluate $f(x_{z_1}),\ldots,f(x_{z_n})$.
These $n$ evaluations determine $A_hf$
and $I_N(A_hf)$ in \eqref{eq:approximation}
and can be reused for subsequent draws.

\paragraph{Costs per draw.}
Given $Q_h$, the standard sequential
projection-DPP sampler requires $O(Nn^2)$
operations per draw. Alternatively, the exact
sampler of Tremblay, Amblard, and Barthelm\'e
\cite{tremblay2017graphsamplingdeterminantalprocesses}
has expected cost $O(Nn+n^3\log(en))$ for
the first draw and $O(n^3\log(en))$ for
each subsequent draw from the same projection.

\subsection{Haar DPPs}
We next consider the computational costs of the Haar
DPPs introduced in Subsection~\ref{sec:block-dpp}. Unlike
\eqref{eq:projection}, the projection $P_{\mathrm{sc}}$
in \eqref{blk:eq:projection} can be constructed directly from
the partition $C_1,\ldots,C_n$.

We first consider the construction of $Z$ and
$C_1,\ldots,C_n$. Once $A$ has been constructed,
we examine the vertices in label order, as in
\eqref{blk:eq:centers}. Whenever we encounter a vertex
that has not yet been covered, we add it to $Z$ and
use Dijkstra's algorithm to determine $B_N(z_\ell,h)$,
stopping the search at distance $h$. Every vertex in
this ball is marked as covered, and those not previously
covered are assigned to $C_\ell$ for the first-cover
partition. 

We now estimate the cost of these searches. For any vertex
$i$, the balls $B_N(z_\ell,h/2)$ with
$i\in B_N(z_\ell,h)$ are pairwise disjoint by
\eqref{blk:eq:centers}, and all lie in $B_N(i,3h/2)$.
By \eqref{blk:eq:volume}, we have
\[
\#\{\ell:i\in B_N(z_\ell,h)\}\,c(h/2)^m
\leq C(3h/2+\delta)^m.
\]
Since $\delta\leq h/2$, each vertex $i$ is visited
in at most a constant number of searches. Thus the
total number of edge examinations is bounded by
\[
C\sum_{i=1}^N\#\{j:A_{ij}=1\}=O(E).
\]
Using a binary heap, the construction of $Z$ and the
first-cover partition therefore requires
$O((N+E)\log N)$ operations and $O(N+E)$ storage.

For the Voronoi partition, we first construct $Z$ as
above and then assign each vertex to its nearest
$z_\ell$ in $d_N$, resolving ties by the index $\ell$.
This can be done by a multi-source Dijkstra search,
with the same computational and storage bounds.

Once $C_1,\ldots,C_n$ are stored, \eqref{blk:eq:law}
allows us to select one vertex independently and
uniformly from each $C_\ell$. Each draw therefore
requires $O(n)$ operations and exactly $n$ evaluations
of $f$.


\section*{AI statement}
In this work, we have used standard generative AI tools to support our work. The authors supplied the research directions and detailed proof ideas, and standard generative AI tools were used for exploratory simulations, literature search, calculations,  the development and simplification of some parts of the proofs and typesetting. The authors have verified all steps, and take responsibility for the final content of this work.

\bibliographystyle{plain}
\bibliography{bibliography,stats}

@article{bardenet2020monte,
  title={Monte Carlo with determinantal point processes},
  author={Bardenet, R{\'e}mi and Hardy, Adrien},
  journal={The Annals of Applied Probability},
  volume={30},
  number={1},
  pages={368--417},
  year={2020}
}

@article{bardenet2024small,
  title={Small coresets via negative dependence: DPPs, linear statistics, and concentration},
  author={Bardenet, R{\'e}mi and Ghosh, Subhroshekhar and Simon-Onfroy, Hugo and Tran, Hoang Son},
  journal={Advances in Neural Information Processing Systems},
  volume={37},
  pages={84329--84349},
  year={2024}
}

@article{tran2026state,
  title={State-of-art minibatches via novel DPP kernels: discretization, wavelets, and rough objectives},
  author={Tran, Hoang-Son and Gupta, Pranav and Bardenet, R{\'e}mi and Ghosh, Subhroshekhar},
  journal={arXiv preprint arXiv:2605.13127},
  year={2026}
}

@article{dpp-sgd,
  title={Determinantal point processes based on orthogonal polynomials for sampling minibatches in SGD},
  author={Bardenet, R. and Ghosh, S. and Lin, M.},
  journal={Advances in Neural Information Processing Systems},
  volume={34},
  pages={16226--16237},
  year={2021}
}

@misc{tremblay2017graphsamplingdeterminantalprocesses,
      title={Graph sampling with determinantal processes}, 
      author={Nicolas Tremblay and Pierre-Olivier Amblard and Simon Barthelmé},
      year={2017},
      eprint={1703.01594},
      archivePrefix={arXiv},
      primaryClass={cs.LG},
      url={https://arxiv.org/abs/1703.01594}, 
}

@article{tran2026determinantals,
      title={Determinantal sampling on general spaces: fast approximation via diffusion geometry}, 
      author={Hoang-Son Tran and Pranav Gupta and Subhroshekhar Ghosh},
      year={2026},
      journal={arXiv preprint arXiv:2607.06644}, 
}

@article{MuruganSaloffCoste2023,
  author  = {Murugan, Mathav and Saloff-Coste, Laurent},
  title   = {Harnack Inequalities and Gaussian Estimates for Random Walks on Metric Measure Spaces},
  journal = {Electronic Journal of Probability},
  volume  = {28},
  number  = {64},
  pages   = {1--81},
  year    = {2023},
  doi     = {10.1214/23-EJP954},
  url     = {https://doi.org/10.1214/23-EJP954}
}

@article{geller-mayeli,
  title={Continuous wavelets on compact manifolds},
  author={Geller, Daryl and Mayeli, Azita},
  journal={Mathematische Zeitschrift},
  volume={262},
  number={4},
  pages={895--927},
  year={2009},
  publisher={Springer}
}

@article{Sos00,
	author = {Soshnikov, A.},
	journal = {Russian Mathematical Surveys},
	pages = {923--975},
	title = {Determinantal random point fields},
	volume = {55},
	year = {2000}}

@book{HKPV09,
	author = {Hough, J. B. and Krishnapur, M. and Peres, Y. and Vir\'ag, B.},
	publisher = {American Mathematical Society},
	title = {Zeros of {G}aussian analytic functions and determinantal point processes},
	volume = {51},
	year = {2009}}

@article {Mac75,
    AUTHOR = {Macchi, Odile},
     TITLE = {The coincidence approach to stochastic point processes},
   JOURNAL = {Advances in Appl. Probability},
  FJOURNAL = {Advances in Applied Probability},
    VOLUME = {7},
      YEAR = {1975},
     PAGES = {83--122},
      ISSN = {0001-8678,1475-6064},
   MRCLASS = {60G55},
  MRNUMBER = {380979},
MRREVIEWER = {Jan\ Grandell},
       DOI = {10.2307/1425855},
       URL = {https://doi-org.libproxy1.nus.edu.sg/10.2307/1425855},
}

@article {Lyo03,
    AUTHOR = {Lyons, Russell},
     TITLE = {Determinantal probability measures},
   JOURNAL = {Publ. Math. Inst. Hautes \'Etudes Sci.},
  FJOURNAL = {Publications Math\'ematiques. Institut de Hautes \'Etudes
              Scientifiques},
    NUMBER = {98},
      YEAR = {2003},
     PAGES = {167--212},
      ISSN = {0073-8301,1618-1913},
   MRCLASS = {60D05 (05C05 60C05 60E15)},
  MRNUMBER = {2031202},
MRREVIEWER = {Lutz\ Peter\ Klotz},
       DOI = {10.1007/s10240-003-0016-0},
       URL = {https://doi-org.libproxy1.nus.edu.sg/10.1007/s10240-003-0016-0},
}

@article {HKPV06,
    AUTHOR = {Hough, J. Ben and Krishnapur, Manjunath and Peres, Yuval and
              Vir\'ag, B\'alint},
     TITLE = {Determinantal processes and independence},
   JOURNAL = {Probab. Surv.},
  FJOURNAL = {Probability Surveys},
    VOLUME = {3},
      YEAR = {2006},
     PAGES = {206--229},
      ISSN = {1549-5787},
   MRCLASS = {60G55 (60F05 82B31)},
  MRNUMBER = {2216966},
MRREVIEWER = {Lutz\ Peter\ Klotz},
       DOI = {10.1214/154957806000000078},
       URL = {https://doi-org.libproxy1.nus.edu.sg/10.1214/154957806000000078},
}

@article{tran2025negative,
  title={Negative Dependence as a toolbox for machine learning: review and new developments},
  author={Tran, H.S. and Petrovic, V. and Bardenet, R. and Ghosh, S.},
  journal={arXiv preprint arXiv:2502.07285},
  year={2025}
}

@article{KuTa12,
	author = {Kulesza, A. and Taskar, B.},
	journal = {Foundations and Trends in Machine Learning},
	title = {Determinantal point processes for machine learning},
	year = {2012}}

\appendix

\section{Proofs from Section \ref{sec:manifolds}} \label{app:manifolds}
Throughout this appendix, we work with the setting in
Section~\ref{sec:manifolds}. In particular, $(\mathcal M,g)$
is a fixed smooth, connected, closed Riemannian manifold of
dimension $m\ge1$. We write
\[
 V=\operatorname{vol}_g(\mathcal M),\qquad
 \mu=V^{-1}\operatorname{vol}_g,\qquad
 L=-\Delta_{\mathcal M},
\]
and let $d$ denote the geodesic distance. Fix a real
orthonormal eigenbasis $\{\phi_k\}_{k\ge1}$ of $L^2(\mu)$,
with $\phi_1=1$ and
\[
 L\phi_k=\lambda_k\phi_k,\qquad
 0=\lambda_1\le\lambda_2\le\cdots.
\]

Let $\mathcal U=\{u_1,\ldots,u_J\}$ be a fixed finite family
of real spectral filters satisfying \eqref{eq:spec-filter},
with at least one nonzero member. For such a filter $u$, write
\[
 W_{u,h}(x,y)=\sum_{k\ge1}u(h^2\lambda_k)\phi_k(x)\phi_k(y).
\]
Let $a\ge1$ and let $h>0$ be the localization scale. Let
$\mathcal Z=\{z_1,\ldots,z_I\}$ be a maximally
$ah$-separated set, and define
\[
 \psi_{i,j}(x)=h^{m/2}W_{u_j,h}(x,z_i),\qquad
 1\le i\le I,\quad 1\le j\le J.
\]
Set
\[
 H_h:=\operatorname{span}\{\psi_{i,j}:1\le i\le I,\ 1\le j\le J\},
 \qquad n_h:=\dim H_h.
\]
Let $\Pi_h$ be the orthogonal projection from $L^2(\mu)$ onto
$H_h$, and let $K_h$ be its integral kernel. All kernels are
taken with respect to $\mu$, and $\|\cdot\|_2$ denotes the
norm in $L^2(\mu)$. 

\subsection{Localization property of the kernel}
\label{app:localization}
\label{app:constructed-localization-section}

The aim of this subsection is to show Lemma \ref{app:constructed-kernel-localization} below, which states that the kernel $K_h$ above satisfies
\[
 |K_h(x,y)|\le C h^{-m}
 \left(1+\frac{d(x,y)}h\right)^{-(m+3)}
\]
when $a$ is sufficiently large and $h$ is sufficiently small. To this end, we need the following technical ingredients.

\paragraph{Spectral multiplier estimates.}
\begin{lemma}
\label{app:multiplier}
Let $u$ be a fixed real function in $C^\infty([0,\infty))$ such that
\[
 \sup_{t\geq0}(1+t)^N|u^{(q)}(t)|<\infty
 \quad\text{for every pair of integers }N,q\geq0.
\]
For every $h>0$, the kernel $W_{u,h}$ of $u(h^2L)$ is smooth and real.
For every integer $p\geq0$, there is a constant $C_p$, depending
only on $(\mathcal M,g)$, $u$, and $p$, such that
\begin{equation}
 |W_{u,h}(x,z)|\leq C_p h^{-m}(1+d(x,z)/h)^{-p},
 \qquad x,z\in\mathcal{M},\quad 0<h\leq1.
 \label{app:multiplier-decay}
\end{equation}
\end{lemma}

\begin{proof}
The spectral multiplier theorem of Geller--Mayeli \cite[Lemma 4.1 and Remark 4.2]{geller-mayeli} gives, for every integer $p\geq0$,
\[
 h^m\bigl(d(x,z)/h\bigr)^p|W_{u,h}(x,z)|\leq C_p,
 \qquad 0<h\leq1.
\]
Combining the bound for $p=0$ with the bound for the chosen $p$, separately on $d(x,z)\leq h$ and $d(x,z)>h$, proves \eqref{app:multiplier-decay}. 
\end{proof}

\begin{lemma}
\label{app:diagonal-asymptotics}
Let $u,v$ be fixed real filters satisfying \eqref{eq:spec-filter}.
Then, uniformly in $z\in\mathcal M$, as $h\downarrow0$,
\begin{equation}
 h^mW_{uv,h}(z,z)
 =\frac{V}{(2\pi)^m}\int_{\mathbb{R}^m}u(|\xi|^2)v(|\xi|^2)\,\mathrm{d}\xi+O(h).
 \label{app:cross-diagonal}
\end{equation}
In particular, if $u$ is nonzero, then
\begin{equation}
 h^mW_{u^2,h}(z,z)=c_u+O(h),\qquad
 c_u=\frac{V}{(2\pi)^m}\int_{\mathbb{R}^m}u(|\xi|^2)^2\,\mathrm{d}\xi>0.
 \label{app:spectral-diagonal}
\end{equation}
The remainder constants depend only on the fixed manifold and filters.
\end{lemma}

\begin{proof}
Using the eigenbasis fixed above, write
\[
 E_z(\Lambda)=\sum_{\lambda_k\leq\Lambda}|\phi_k(z)|^2,
 \qquad \kappa_m=(2\pi)^{-m}\omega_m,
\]
where $\omega_m$ is the Euclidean unit-ball volume. The local Weyl law gives
\begin{equation}
 E_z(\Lambda)=V\kappa_m\Lambda^{m/2}
              +O\bigl(1+\Lambda^{(m-1)/2}\bigr),
 \qquad \Lambda\geq0,
 \label{app:local-weyl}
\end{equation}
with a uniform remainder after increasing its constant on a bounded interval. 
Let $q=u^2$, or $q=uv$. Since $q$ is smooth and rapidly decreasing, spectral calculus and integration by parts give
\begin{align*}
 W_{q,h}(z,z)
 =\int_{[0,\infty)}q(h^2\Lambda)\,\mathrm{d}E_z(\Lambda)
 =-\int_0^\infty E_z(t/h^2)q'(t)\,\mathrm{d}t.
\end{align*}
Multiplication of \eqref{app:local-weyl} by $h^m$ yields an integrated remainder bounded by
\[
 C h^m\int_0^\infty|q'(t)|\,\mathrm{d}t
 +C h\int_0^\infty t^{(m-1)/2}|q'(t)|\,\mathrm{d}t
 =O(h),\qquad 0<h\leq1.
\]
Both integrals are finite, including at $m=1$. Hence
\begin{align*}
 h^mW_{q,h}(z,z)
 &=-V\kappa_m\int_0^\infty t^{m/2}q'(t)\,\mathrm{d}t+O(h)\\
 &=V\kappa_m\frac m2\int_0^\infty t^{m/2-1}q(t)\,\mathrm{d}t+O(h)\\
 &=\frac{V}{(2\pi)^m}\int_{\mathbb{R}^m}q(|\xi|^2)\,\mathrm{d}\xi+O(h).
\end{align*}
The boundary terms vanish by smoothness at zero and rapid decay at infinity. For $q=u^2$, the limiting constant is strictly positive when $u$ is nonzero. This proves \eqref{app:spectral-diagonal} and \eqref{app:cross-diagonal}.

\end{proof}

For any two real filters satisfying \eqref{eq:spec-filter},
Parseval's identity applied to the spectral expansions gives
\begin{equation}
 \int_{\mathcal{M}}W_{u,h}(x,z)W_{v,h}(x,z')\,\mathrm{d}\mu(x)
       =W_{uv,h}(z,z').
 \label{app:composition}
\end{equation}
Indeed, the coefficients of the two kernels as functions of $x$
are $u(h^2\lambda_k)\phi_k(z)$ and
$v(h^2\lambda_k)\phi_k(z')$, respectively; their inner product
is the spectral series on the right.
All kernels and inner products in this identity use the same
reference measure $\mu$.

\paragraph{Separation and packing.}

\begin{lemma}
\label{app:ball-volume-lemma}
There are constants $R_0,c_V,C_V>0$, depending only on
$(\mathcal M,g)$, such that
\begin{equation}
 c_V R^m\leq\mu(B(x,R))\leq C_VR^m,
 \qquad x\in\mathcal{M},\quad 0<R\leq R_0.
 \label{app:ball-volumes}
\end{equation}
The same bounds hold for closed balls. After increasing $C_V$,
the upper bound holds for every $R>0$.
\end{lemma}

\begin{proof}
Choose $R_0$ smaller than the injectivity radius. In normal
coordinates centered at $x$, the radius-$R$ geodesic ball
corresponds to the Euclidean radius-$R$ ball. Compactness gives
uniform positive lower and upper bounds on the volume density
for $0<R\leq R_0$, which proves \eqref{app:ball-volumes}.
The boundary sphere has zero volume in these coordinates,
so the estimates also hold for closed balls. For $R>R_0$,
the upper bound follows from $\mu(\mathcal M)=1$ after
increasing $C_V$ to at least $R_0^{-m}$.
\end{proof}

\begin{lemma}
\label{app:packing}
Let $a\geq1$ and $0<r=ah\leq R_0$, where $R_0$ is as in
Lemma~\ref{app:ball-volume-lemma}. Let
$\mathcal Z=\{z_1,\ldots,z_I\}$ be a maximally $r$-separated
set.
Then, the closed balls $\overline B(z_i,r)$ cover $\mathcal M$, and
\begin{equation}
 c r^{-m}\leq I\leq C r^{-m},\qquad
 \#\{i:d(x,z_i)\leq R\}\leq C_P(1+R/r)^m,
 \quad x\in\mathcal{M},\ R>0.
 \label{app:packing-count}
\end{equation}
The constants depend only on $(\mathcal M,g)$, independently
of $a,h$, and the chosen maximal set.
\end{lemma}

\begin{proof}
The open balls $B(z_i,r/2)$ are pairwise disjoint. The closed
radius-$r$ balls cover $\mathcal M$, since otherwise a point
at distance greater than $r$ from every center could be adjoined.
Summing the lower volume estimates from
Lemma~\ref{app:ball-volume-lemma} over the disjoint balls gives
$I\leq C r^{-m}$. Summing the upper estimates over the covering
balls gives $I\geq c r^{-m}$.

If $d(x,z_i)\leq R$, then
$B(z_i,r/2)\subset B(x,R+r/2)$. Hence
\[
 \#\{i:d(x,z_i)\leq R\}\,c_V(r/2)^m
       \leq C_V(R+r/2)^m,
\]
which proves the counting bound. A maximal set exists by
successively adding admissible points: the positive lower
volume of the disjoint half-radius balls bounds the number
of possible additions.
\end{proof}

\paragraph{Gram matrix estimates.}

Set
\[
 b:=\dim_{\mathbb R}\operatorname{span}\{u_1,\ldots,u_J\}\ge1.
\]
After relabeling, choose $u_1,\ldots,u_b$ as a linearly
independent spanning subfamily. At each center, the atoms
from the remaining filters are linear combinations of these
$b$ atoms, so retaining only this subfamily leaves $H_h$
unchanged. We write matrices in $b\times b$ blocks indexed
by the centers. Define
\begin{equation}
 \|T\|_{a,h}:=\max_i\sum_j\|T_{ij}\|_{\mathrm{op}}
                     (1+d(z_i,z_j)/(ah))^{m+3}.
 \label{app:weighted-norm}
\end{equation}

\begin{lemma}
\label{app:defect}
Suppose the centers satisfy Lemma~\ref{app:packing}, with $a\geq1$. If a block matrix $E$ satisfies
\[
 \max_i\|E_{ii}\|_{\mathrm{op}}\leq\delta,\qquad
 \|E_{ij}\|_{\mathrm{op}}\leq C_0(1+d(z_i,z_j)/h)^{-(2m+4)}
 \quad(i\ne j),
\]
then
\begin{equation}
 \|E\|_{a,h}\leq\delta+C C_0a^{-(2m+4)},
 \label{app:defect-bound}
\end{equation}
where $C$ depends only on the packing constant and $m$.
\end{lemma}

\begin{proof}
Put $r=ah$. For a fixed $i$, partition the other centers into
\[
 \mathcal{J}_k(i)=\{j\ne i:2^kr\leq d(z_i,z_j)<2^{k+1}r\},\qquad k\geq0.
\]
Strict separation ensures that these sets cover all $j\ne i$. Packing gives $|\mathcal{J}_k(i)|\leq C2^{km}$. On this annulus,
\[
 (1+d(z_i,z_j)/h)^{-(2m+4)}\leq a^{-(2m+4)}2^{-k(2m+4)},
\]
whereas the weight in \eqref{app:weighted-norm} is at most $C2^{k(m+3)}$. Therefore
\[
 \sum_{j\ne i}\|E_{ij}\|_{\mathrm{op}}
                   (1+d(z_i,z_j)/r)^{m+3}
 \leq CC_0a^{-(2m+4)}\sum_{k\geq0}2^{-k}.
\]
Add the diagonal term and maximize over $i$.
\end{proof}


\begin{lemma}
\label{app:inverse-root}
Let $G=\operatorname{Id}+E$ be a real symmetric block matrix indexed by the centers, and suppose that $\|E\|_{a,h}\leq1/2$. Then
\[
 \frac12\operatorname{Id}\preceq G\preceq\frac32\operatorname{Id},
\]
and its positive inverse square root satisfies
\begin{equation}
 \|G^{-1/2}\|_{a,h}\leq\sqrt2.
 \label{app:inverse-root-norm}
\end{equation}
The corresponding weighted column sums obey the same bound.
\end{lemma}

\begin{proof}
Put $w_{ij}=(1+d(z_i,z_j)/(ah))^{m+3}$. The triangle inequality gives $w_{ij}\leq w_{ik}w_{kj}$, and hence
\[
 \|TU\|_{a,h}\leq\|T\|_{a,h}\|U\|_{a,h},
 \qquad \|\operatorname{Id}\|_{a,h}=1.
\]
Since $E$ is symmetric, both the row sums and the column sums of $\|E_{ij}\|_{\mathrm{op}}$ are at most $1/2$. The Schur estimate gives $\|E\|_{\mathrm{op}}\leq1/2$, which proves the spectral bounds on $G$.

For $|t|<1$,
\[
 (1+t)^{-1/2}=\sum_{k=0}^\infty c_kt^k,
 \qquad c_k=(-1)^k4^{-k}\binom{2k}{k}.
\]
Submultiplicativity shows that this series converges in the weighted norm after substituting $E$. By the spectral theorem its sum is $F=G^{-1/2}$, and
\[
 \|F\|_{a,h}\leq\sum_{k=0}^\infty|c_k|2^{-k}=\sqrt2.
\]
The weighted column sums satisfy the same estimate because $F$ and the weights are symmetric.
\end{proof}

\begin{lemma}
\label{app:spectral-gram-stability}
For the independent subfamily $u_1,\ldots,u_b$ fixed above,
let $G_h^{(b)}$ be the Gram matrix of its atoms, namely
\[
 \psi_{i,\alpha}(x)=h^{m/2}W_{u_\alpha,h}(x,z_i),
 \qquad
 (G_h^{(b)})_{(i,\alpha),(j,\beta)}
 =\langle\psi_{i,\alpha},\psi_{j,\beta}\rangle_{L^2(\mu)}.
\]
Then, the matrix
\[
 \mathsf C_{\alpha\beta}
 =\frac{V}{(2\pi)^m}\int_{\mathbb R^m}
     u_\alpha(|\xi|^2)u_\beta(|\xi|^2)\,\mathrm d\xi,
 \qquad 1\le\alpha,\beta\le b,
\]
is positive definite. Let $c_*>0$ and $C_*<\infty$ be its
smallest and largest eigenvalues, and let $D$ be the
block-diagonal matrix with $I$ copies of $\mathsf C^{-1/2}$.

There exists $a_0\ge1$, depending only on $(\mathcal M,g)$
and the fixed filters, such that for every fixed $a\ge a_0$
there is $h_0>0$, depending additionally on $a$, with the
following property. For every $0<h\le h_0$ and every maximally
$ah$-separated set $\mathcal Z$, the normalized Gram matrix
satisfies
\begin{equation}
 \bigl\|D^{\top}G_h^{(b)}D-\operatorname{Id}\bigr\|_{a,h}
 \le\frac12,
 \label{app:spectral-gram-defect}
\end{equation}
where the weighted block norm is defined in
\eqref{app:weighted-norm}. In particular,
\begin{equation}\label{app:filter-gram-bounds}
 \frac{c_*}{2}\operatorname{Id}
 \preceq G_h^{(b)}
 \preceq\frac{3C_*}{2}\operatorname{Id}.
\end{equation}
All estimates are uniform over the indicated scales and
center sets.
\end{lemma}

\begin{proof}
Rapid decay of the filters makes every entry of $\mathsf C$
finite. For $c\in\mathbb R^b$,
\[
 c^{\top}\mathsf Cc
 =\frac{V}{(2\pi)^m}\int_{\mathbb R^m}
      \left|\sum_{\alpha=1}^b c_\alpha u_\alpha(|\xi|^2)\right|^2
      \,\mathrm d\xi.
\]
If this quantity vanishes, continuity and radial integration
give $\sum_\alpha c_\alpha u_\alpha(t)=0$ for every $t>0$.
Continuity gives the equality also at zero, and linear
independence of the filters implies $c=0$. Thus
$\mathsf C$ is positive definite, so $D$ is well defined.

The composition identity \eqref{app:composition} gives the
exact Gram entries
\begin{equation}
 (G_h^{(b)})_{(i,\alpha),(j,\beta)}
 =h^mW_{u_\alpha u_\beta,h}(z_i,z_j).
 \label{app:spectral-gram-entries}
\end{equation}
For atoms at a common center, the diagonal spectral asymptotics
in Lemma~\ref{app:diagonal-asymptotics} imply
\[
 \sup_i\|(G_h^{(b)})_{ii}-\mathsf C\|_{\mathrm{op}}
 \le Ch,\qquad 0<h\le1.
\]
The products $u_\alpha u_\beta$ also satisfy
\eqref{eq:spec-filter}. Applying Lemma~\ref{app:multiplier}
to these finitely many products with exponent $2m+4$ gives,
for $i\ne j$,
\[
 \|(G_h^{(b)})_{ij}\|_{\mathrm{op}}
 \le C\left(1+\frac{d(z_i,z_j)}h\right)^{-(2m+4)}.
\]
Both constants depend only on the fixed manifold and filters.

Put
\[
 \widetilde G_h=D^{\top}G_h^{(b)}D
              =\operatorname{Id}+E_h.
\]
Multiplication of each center block by the fixed matrix
$\mathsf C^{-1/2}$ therefore yields
\[
 \sup_i\|(E_h)_{ii}\|_{\mathrm{op}}\le C_1h,
 \qquad
 \|(E_h)_{ij}\|_{\mathrm{op}}
 \le C_2\left(1+\frac{d(z_i,z_j)}h\right)^{-(2m+4)}
 \quad(i\ne j),
\]
where $C_1,C_2$ are independent of $a,h$, and the centers.
When $a\ge1$ and $ah\le R_0$, the packing estimate in
Lemma~\ref{app:packing} is uniform over the maximal center
sets. Lemma~\ref{app:defect} consequently gives
\[
 \|E_h\|_{a,h}\le C_1h+C_3a^{-(2m+4)}.
\]
The spatial weight in this norm controls the sum of the
interactions with all other centers, uniformly as $I$ grows.
The constant $C_3$ depends only on the manifold and filters.

Choose $a_0\ge1$ so that $C_3a_0^{-(2m+4)}\le1/4$.
For each fixed $a\ge a_0$, choose $0<h_0\le1$ so that
$C_1h_0\le1/4$ and $ah_0\le R_0$, where $R_0$ is as in
Lemma~\ref{app:ball-volume-lemma}. This proves
\eqref{app:spectral-gram-defect} for every indicated scale and
center set. Since $E_h$ is real symmetric,
Lemma~\ref{app:inverse-root} gives
\[
 \frac12\operatorname{Id}\preceq\widetilde G_h
 \preceq\frac32\operatorname{Id}.
\]
Finally,
$G_h^{(b)}=D^{-\top}\widetilde G_hD^{-1}$, and
$D^{-\top}D^{-1}$ has every diagonal block equal to
$\mathsf C$. Congruence in the last inequality, followed by
$c_*\operatorname{Id}_b\preceq\mathsf C\preceq
C_*\operatorname{Id}_b$, proves
\eqref{app:filter-gram-bounds}.
\end{proof}

\paragraph{Localization property.}

We conclude with the desired localization proprety of our kernel.

\begin{lemma}
\label{app:constructed-kernel-localization}
For the filter family and projection kernel fixed at the
beginning of the appendix, the following holds.
There exists $a_0=a_0(\mathcal M,g,\mathcal U)\ge1$ such that,
for every fixed $a\ge a_0$, there is
$h_0=h_0(\mathcal M,g,\mathcal U,a)>0$ with the following
property.

For $0<h\le h_0$ and every maximally $ah$-separated center
set, the kernel $K_h$ of the projection onto $H_h$ satisfies
\begin{equation}
 |K_h(x,y)|\le C h^{-m}
       \left(1+\frac{d(x,y)}h\right)^{-(m+3)},
 \qquad x,y\in\mathcal M.
 \label{app:kernel-localization}
\end{equation}
The constant $C$ depends only on $(\mathcal M,g)$, the fixed
filter family $\mathcal{U}$, and $a$.
\end{lemma}

\begin{proof}
Use the independent spanning subfamily $u_1,\ldots,u_b$
selected above.
For each center, the atoms of any remaining filter are the
same fixed linear combination of the selected atoms, by
linearity of spectral calculus. Consequently,
\[
 H_h=\operatorname{span}\{\psi_{i,\alpha}:
 1\le i\le I,\ 1\le\alpha\le b\},\qquad
 \psi_{i,\alpha}=h^{m/2}W_{u_\alpha,h}(\cdot,z_i).
\]
In particular, the reduction does not change the projection
whose kernel is $K_h$.

Apply Lemma~\ref{app:spectral-gram-stability} to this independent
subfamily and use its thresholds $a_0,h_0$. Let $A_h$ be the
synthesis map of the reduced atoms, let $G_h^{(b)}=A_h^*A_h$,
and let $D$ be the block-diagonal normalization from that
lemma. Set
\[
 \widetilde A_h=A_hD,\qquad
 \widetilde G_h=D^{\top}G_h^{(b)}D,\qquad
 F_h=\widetilde G_h^{-1/2}.
\]
The lemma verifies, for these particular spectral atoms, that
$\|\widetilde G_h-\operatorname{Id}\|_{a,h}\le1/2$.
Lemma~\ref{app:inverse-root} therefore gives
\[
 \|F_h\|_{a,h}\le\sqrt2.
\]
Because $F_h$ is symmetric, the corresponding weighted
column sums are also at most $\sqrt2$.

Write $\widetilde\psi_i(x)$ for the vector of the $b$ atoms
at center $z_i$ after the change of basis $D$. The multiplier
estimate in Lemma~\ref{app:multiplier}, with exponent $m+3$,
gives
\begin{equation}
 \|\widetilde\psi_i(x)\|_{\mathbb R^b}
 \le C h^{-m/2}
       \left(1+\frac{d(x,z_i)}h\right)^{-(m+3)}.
 \label{app:input-atoms}
\end{equation}
The constant is uniform because the change of basis within
each center is the same fixed matrix. Let $\eta_{i,\alpha}$
be the columns of $\widetilde A_hF_h$. We group their values as
\[
 \eta_i(x)=(\eta_{i,1}(x),\ldots,\eta_{i,b}(x))^\top,
 \qquad
 \|\eta_i(x)\|_{\mathbb R^b}
 =\left(\sum_{\alpha=1}^b|\eta_{i,\alpha}(x)|^2\right)^{1/2}.
\]
This is a Euclidean norm at the fixed point $x$. The vectors satisfy
\[
 \eta_i(x)=\sum_{j=1}^I(F_h)_{ji}^{\top}
                     \widetilde\psi_j(x).
\]
The triangle inequality and $a\ge1$ imply
\[
 1+\frac{d(x,z_i)}h
 \le a\left(1+\frac{d(x,z_j)}h\right)
       \left(1+\frac{d(z_j,z_i)}{ah}\right).
\]
Combining this inequality with \eqref{app:input-atoms} gives
\begin{align*}
 \|\eta_i(x)\|_{\mathbb R^b}
 &\le C a^{m+3}h^{-m/2}
       \left(1+\frac{d(x,z_i)}h\right)^{-(m+3)}
       \sum_j\|(F_h)_{ji}\|_{\mathrm{op}}
       \left(1+\frac{d(z_j,z_i)}{ah}\right)^{m+3}.
\end{align*}
The weighted column bound now yields
\begin{equation}
 \|\eta_i(x)\|_{\mathbb R^b}
 \le C h^{-m/2}
       \left(1+\frac{d(x,z_i)}h\right)^{-(m+3)}.
 \label{app:orthonormal-localization}
\end{equation}
This proves localization for the orthogonalized atoms of the
actual construction.

We next identify their projection. Direct multiplication gives
\[
 (\widetilde A_hF_h)^*(\widetilde A_hF_h)
 =F_h\widetilde G_hF_h=\operatorname{Id}.
\]
Both $D$ and $F_h$ are invertible, so the range of
$\widetilde A_hF_h=A_hDF_h$ is exactly $H_h$. Thus its columns
form an orthonormal basis of $H_h$, and the projection
$\Pi_h$ defined in the main text is
\[
 \Pi_h=(\widetilde A_hF_h)(\widetilde A_hF_h)^*
       =A_h(G_h^{(b)})^{-1}A_h^*.
\]
In particular, its kernel is
\begin{equation}\label{app:constructed-kernel-identity}
 K_h(x,y)=\sum_{i=1}^I\sum_{\alpha=1}^b
                  \eta_{i,\alpha}(x)\eta_{i,\alpha}(y)
          =\sum_{i=1}^I\eta_i(x)^{\top}\eta_i(y).
\end{equation}
All these functions are smooth finite linear combinations
of the original atoms. Equality of the integral operators
therefore identifies their smooth kernels everywhere on
$\mathcal M\times\mathcal M$.

It remains to sum \eqref{app:orthonormal-localization}.
Put $v_i(x)=(1+d(x,z_i)/h)^{-(m+3)}$.
The packing bound in Lemma~\ref{app:packing} gives
\begin{equation}\label{app:atom-sum}
 \begin{split}
 \sup_{x\in\mathcal M}\sum_i v_i(x)
 &\le C_P(1+a^{-1})^m
   +C_P\sum_{k\ge0}2^{-k(m+3)}(1+2^{k+1}/a)^m\\
 &\le C.
 \end{split}
\end{equation}
Indeed, the first term counts centers within distance $h$,
and the $k$th term bounds the contribution of centers at
distances between $2^kh$ and $2^{k+1}h$.
For every center $z_i$, at least one of $d(x,z_i)$ and
$d(y,z_i)$ is at least $d(x,y)/2$. It follows that
\begin{align*}
 \sum_i v_i(x)v_i(y)
 &\le\left(1+\frac{d(x,y)}{2h}\right)^{-(m+3)}
       \left(\sum_i v_i(x)+\sum_i v_i(y)\right)\\
 &\le C\left(1+\frac{d(x,y)}h\right)^{-(m+3)}.
\end{align*}
Finally, \eqref{app:constructed-kernel-identity} and
\eqref{app:orthonormal-localization} imply
\[
 |K_h(x,y)|
 \le C h^{-m}\sum_i v_i(x)v_i(y)
 \le C h^{-m}\left(1+\frac{d(x,y)}h\right)^{-(m+3)}.
\]
This is \eqref{app:kernel-localization} for the kernel in the
main text. Every constant is uniform over the stated scales
and center sets for the fixed $a$ and filter family.
\end{proof}

\subsection{Proof of Theorem \ref{thm:var-reduce-manifold}}
\label{app:variance-filter}

Take
$a\ge a_0$ and $0<h\le h_0$ in the regime of
Lemma~\ref{app:constructed-kernel-localization}. Throughout this
subsection, $K_h$ is the projection kernel constructed from
this family and the centers fixed at the beginning of the
appendix. 

Recall that for a test function $f$
we have
\begin{equation}
 \operatorname{Var}\big [\mathcal S_h(f)\big ]
 =\frac12\iint_{\mathcal{M}^2}|f(x)-f(y)|^2|K_h(x,y)|^2\,
                    \mathrm{d}\mu(x)\mathrm{d}\mu(y).
 \label{app:variance-identity}
\end{equation}

\paragraph{H\"older observables.}
Let $f$ belong to $C^{0,s}(\mathcal{M})$ with $0<s\le 1$.
Fix $x$ and split the $y$ integral into $d(x,y)\leq h$ and the annuli $2^jh<d(x,y)\leq2^{j+1}h$, $j\geq0$. The kernel bound and the volume estimate in Lemma~\ref{app:ball-volume-lemma} give
\[
 \int_{d(x,y)\leq h}d(x,y)^{2s}|K_h(x,y)|^2\,\mathrm{d}\mu(y)
 \leq C h^{2s-2m}\mu(\overline B(x,h))\leq C h^{2s-m}.
\]
On the $j$th annulus the corresponding integral is at most
\[
 C h^{-2m}2^{-2j(m+3)}(2^{j+1}h)^{2s}
                        \mu(\overline B(x,2^{j+1}h))
 \leq C_s h^{2s-m}2^{j(2s-m-6)}.
\]
The series is summable for every $0<s\leq1$. Substituting $|f(x)-f(y)|\leq[f]_{C^{0,s}}d(x,y)^s$ into \eqref{app:variance-identity} and using $\mu(\mathcal M)=1$ yields
\[
 \operatorname{Var}\big[\mathcal S_h(f)\big]
 \leq C_s h^{2s-m}[f]_{C^{0,s}}^2.
\]
Note that $n_h \asymp h^{-m}$. This completes the proof for H\"older observables.

\paragraph{Sobolev observables}
Let $f\in H^s(\mathcal{M})$ with $0<s<1$.
For $t>0$, set $v=t/h$. Since $0<s<1$,
\[
 h^{-2m}(1+t/h)^{-2m-6}
 =h^{2s-m}t^{-m-2s}\frac{v^{m+2s}}{(1+v)^{2m+6}}
 \leq h^{2s-m}t^{-m-2s}.
\]
It follows from \eqref{app:kernel-localization} that
\[
 |K_h(x,y)|^2\leq C h^{2s-m}d(x,y)^{-m-2s},
 \qquad x\ne y.
\]
Substitution into \eqref{app:variance-identity} bounds the variance by $C h^{2s-m}[f]_{H^s}^2$. Using $n_h\asymp h^{-m}$ yields the desired variance reduction for $f\in H^s(\mathcal{M})$ with $0<s<1$.

For $s=1$, namely, $f\in H^1(\mathcal{M})$, the required control of increments follows from
local Poincar\'e inequalities. We first establish the pair
estimate used to integrate the localized kernel.

\begin{lemma}
\label{app:energy}
There is a constant $C$, depending only on $(\mathcal{M},g)$, such that for every $f\in H^1(\mathcal{M})$ and $R>0$,
\begin{equation}
 D_f(R):=\iint_{d(x,y)\leq R}|f(x)-f(y)|^2\,
                   \mathrm{d}\mu(x)\mathrm{d}\mu(y)
 \leq C R^{m+2}\int_{\mathcal{M}}|\nabla f|_g^2\,\mathrm{d}\mu.
 \label{app:energy-pair-bound}
\end{equation}
\end{lemma}

\begin{proof}
Choose $R_*>0$ so small that geodesic balls of radius at most $4R_*$ have uniform volume comparisons and local Poincar\'e inequalities
\[
 \int_{B(z,t)}|f-f_{B(z,t)}|^2\,\mathrm{d}\mu
 \leq C t^2\int_{B(z,t)}|\nabla f|_g^2\,\mathrm{d}\mu,
 \qquad 0<t\leq4R_*.
\]
Here $f_B=\mu(B)^{-1}\int_Bf\,\mathrm{d}\mu$. These inequalities follow by pulling back to Euclidean balls in normal coordinates: the metric and volume density are uniformly comparable with their Euclidean counterparts, and the mean minimizes the squared error over constants.

For $0<R\leq R_*$, take a maximal strictly $R$-separated set $\{x_\ell\}$, and set $B_\ell=B(x_\ell,3R)$. For every pair with $d(x,y)\leq R$, the radius-$R$ covering places $x$ within $R$ of some $x_\ell$, so both $x$ and $y$ belong to $B_\ell$. Thus
\[
 \mathbf1_{\{d(x,y)\leq R\}}
 \leq\sum_\ell\mathbf1_{B_\ell}(x)\mathbf1_{B_\ell}(y)
 \quad\text{for all }x,y.
\]
The family $(B_\ell)$ has bounded overlap. Indeed, when $x\in B_\ell$, the disjoint balls $B(x_\ell,R/2)$ lie in $B(x,7R/2)$, and volume comparison bounds their number uniformly in $R$.

For smooth $f$, the identity
\[
 \iint_{B\times B}|f(x)-f(y)|^2\,\mathrm{d}\mu(x)\mathrm{d}\mu(y)
 =2\mu(B)\int_B|f-f_B|^2\,\mathrm{d}\mu
\]
therefore gives
\begin{align*}
 D_f(R)
 &\leq2\sum_\ell\mu(B_\ell)
                   \int_{B_\ell}|f-f_{B_\ell}|^2\,\mathrm{d}\mu\\
 &\leq C R^{m+2}\sum_\ell\int_{B_\ell}|\nabla f|_g^2\,\mathrm{d}\mu
 \leq C'R^{m+2}\int_{\mathcal{M}}|\nabla f|_g^2\,\mathrm{d}\mu.
\end{align*}
For $R\geq R_*$, connectedness gives $\lambda_2>0$ and the global Poincar\'e inequality $\|f-\int f\,\mathrm d\mu\|_2^2\le\lambda_2^{-1}\int|\nabla f|_g^2\,\mathrm d\mu$. Hence
\begin{align*}
 D_f(R)
 &\leq\iint_{\mathcal{M}^2}|f(x)-f(y)|^2\,\mathrm{d}\mu(x)\mathrm{d}\mu(y)\\
 &=2\left\|f-\int_{\mathcal{M}}f\,\mathrm{d}\mu\right\|_2^2
 \leq C\int_{\mathcal{M}}|\nabla f|_g^2\,\mathrm{d}\mu\\
 &\leq C R_*^{-m-2}R^{m+2}\int_{\mathcal{M}}|\nabla f|_g^2\,\mathrm{d}\mu.
\end{align*}
Finally, smooth functions are dense in $H^1(\mathcal{M})$, and their pair increments converge in $L^2(\mathcal{M}^2)$ because
\[
 \|v(x)-v(y)\|_{L^2(\mathcal{M}^2)}\leq2\|v\|_{L^2(\mu)}.
\]
Passing to the limit proves the assertion for all $f\in H^1(\mathcal{M})$.
\end{proof}

Let $D_f(R)$ be the pair integral in Lemma~\ref{app:energy}. Decompose \eqref{app:variance-identity} into $d(x,y)\leq h$ and the annuli $2^jh<d(x,y)\leq2^{j+1}h$. The kernel bound and Lemma~\ref{app:energy} give
\begin{align*}
 \operatorname{Var}\big[\mathcal S_h(f)\big]
 &\leq C h^{-2m}\left(D_f(h)+
             \sum_{j\geq0}2^{-2j(m+3)}D_f(2^{j+1}h)\right)\\
 &\leq C h^{2-m}\int_{\mathcal{M}}|\nabla f|_g^2\,\mathrm{d}\mu
                  \left(1+\sum_{j\geq0}2^{-j(m+4)}\right).
\end{align*}
Using $n_h\asymp h^{-m}$ yields the result.

\subsection{Proof of Theorem \ref{thm:quadrature-manifold}}
\label{app:quadrature}
\label{aq:heat-pair}
\label{aq:approximation}
\label{aq:quadrature-proof}
\label{ao:optimality}

\paragraph{Uniform intensity bounds}

Write $p_t(x,y)$ for the kernel of $e^{-tL}$ with respect to
$\mu$. 
In this subsection we specialize the family fixed at the
beginning of the appendix to $\mathcal U=\{u_1,u_2\}$, where
\[
 u_1(t)=e^{-t}-e^{-2t},\qquad
 u_2(t)=e^{-2t}-e^{-4t}.
\]
Both filters satisfy \eqref{eq:spec-filter}, and their
atoms are
\[
 \psi_{i,1}(x)=h^{m/2}(p_{h^2}-p_{2h^2})(x,z_i),\qquad
 \psi_{i,2}(x)=h^{m/2}(p_{2h^2}-p_{4h^2})(x,z_i).
\]
Order the $2I$ atoms by center and put
\[
 \Psi_h(x)=
 (\psi_{1,1}(x),\psi_{1,2}(x),\ldots,
   \psi_{I,1}(x),\psi_{I,2}(x))^{\top}.
\]
Let $G_h$ be their Gram matrix in $L^2(\mu)$.

\begin{lemma}
\label{aq:heat-gram-lemma}
For every $h>0$ and every finite set of centers, the Gram matrix
of the two heat-filter atoms has the center blocks
\begin{equation}\label{aq:heat-gram}
 (G_h)_{i\ell}=h^m
 \begin{pmatrix}
 p_{2h^2}-2p_{3h^2}+p_{4h^2}
 &p_{3h^2}-p_{4h^2}-p_{5h^2}+p_{6h^2}\\
 p_{3h^2}-p_{4h^2}-p_{5h^2}+p_{6h^2}
 &p_{4h^2}-2p_{6h^2}+p_{8h^2}
 \end{pmatrix}(z_i,z_\ell).
\end{equation}
\end{lemma}

\begin{proof}
The heat semigroup identity, with all kernels relative to the
same measure $\mu$, is
\[
 \int_{\mathcal M}p_{s h^2}(x,z_i)p_{t h^2}(x,z_\ell)
 \,\mathrm d\mu(x)=p_{(s+t)h^2}(z_i,z_\ell),\qquad s,t>0.
\]
Expanding the products of the two differences gives
\eqref{aq:heat-gram}. 
\end{proof}

Define the $2\times2$ matrix
\[
 \mathsf C_{\alpha\beta}
 :=\frac{V}{(2\pi)^m}\int_{\mathbb R^m}
 u_\alpha(|\xi|^2)u_\beta(|\xi|^2)\,\mathrm d\xi,
 \qquad \alpha,\beta\in\{1,2\}.
\]
Let $c_*$ and $C_*$ denote its smallest and largest eigenvalues.

\begin{lemma}
\label{aq:rank-projection}
The matrix $\mathsf C$ is positive definite. There exists
$a_0=a_0(\mathcal M,g)\ge1$ such that, for every fixed
$a\ge a_0$, there is $h_0=h_0(\mathcal M,g,a)>0$ with the
following property. If $0<h\le h_0$ and
$\mathcal Z=\{z_1,\ldots,z_I\}$ is any maximally
$ah$-separated set, then
\begin{equation}\label{aq:gram-bounds}
 \frac{c_*}{2}\operatorname{Id}_{2I}
 \preceq G_h\preceq
 \frac{3C_*}{2}\operatorname{Id}_{2I}.
\end{equation}
Consequently, the $2I$ atoms are linearly independent,
$n_h=2I\asymp h^{-m}$, and their orthogonal projection has kernel
\begin{equation}\label{aq:kernel-gram}
 K_h(x,y)=\Psi_h(x)^{\top}G_h^{-1}\Psi_h(y),
 \qquad x,y\in\mathcal M.
\end{equation}
The implicit constants depend only on $(\mathcal M,g)$ and $a$,
and are uniform over the admissible scales and center sets.
\end{lemma}

\begin{proof}
The two filters are linearly independent. Indeed, $u_1(t)>0$
for $t>0$, and
\[
 \frac{u_2(t)}{u_1(t)}=e^{-t}(1+e^{-t})
\]
is not constant. The matrix $\mathsf C$ is therefore positive definite: if a vector $c\in\mathbb R^2$ has
$c^{\top}\mathsf Cc=0$, continuity implies
$c_1u_1(t)+c_2u_2(t)=0$ for all $t>0$, and independence gives
$c=0$. The spectral Gram bound
\eqref{app:filter-gram-bounds} in Lemma~\ref{app:spectral-gram-stability} gives \eqref{aq:gram-bounds}
for sufficiently large fixed $a$ and sufficiently small $h$.
In particular, the atoms are independent and $n_h=2I$.
Lemma~\ref{app:packing} gives $I\asymp h^{-m}$ for fixed $a$.

For completeness, let $T_h:\mathbb R^{2I}\to L^2(\mu)$ be
the synthesis map of the atoms. Then $T_h^*T_h=G_h$, and
$T_hG_h^{-1}T_h^*$ is self-adjoint, is the identity on the
range of $T_h$, and vanishes on its orthogonal complement.
It is thus $\Pi_h$, which proves \eqref{aq:kernel-gram}.
\end{proof}

\begin{lemma}
\label{app:heat-profile-lemma}
Let $p_t$ be the heat kernel with respect to $\mu$, and fix $a,c>0$.
Then
\begin{equation}
 \sup_{\substack{z\in\mathcal{M},\ v\in T_z\mathcal M\\ |v|_g\leq a}}
 \left|h^mp_{ch^2}(\exp_z(hv),z)-Vq_c(v)\right|
 \longrightarrow0\quad\text{as }h\downarrow0,
 \label{app:heat-profile}
\end{equation}
where
\[
 q_c(v)=(4\pi c)^{-m/2}e^{-|v|_g^2/(4c)}.
\]
\end{lemma}

\begin{proof}
The local heat-kernel expansion, inside a uniformly chosen normal neighborhood, gives for every fixed $A<\infty$
\[
 p_t(x,z)=V(4\pi t)^{-m/2}e^{-d(x,z)^2/(4t)}
           \bigl(b_0(x,z)+O_A(t)\bigr),
 \qquad d(x,z)\leq A\sqrt t,
\]
where $b_0$ is smooth near the diagonal and $b_0(z,z)=1$. The remainder is inside the displayed prefactor and is uniform in the stated region. For $x=\exp_z(hv)$, $|v|\leq a$, and $t=ch^2$, this gives
\[
 h^mp_{ch^2}(\exp_z(hv),z)
 =Vq_c(v)\bigl(b_0(\exp_z(hv),z)+O_{a,c}(h^2)\bigr).
\]
Compactness implies $b_0(\exp_z(hv),z)\to1$ uniformly for $z\in\mathcal{M}$ and $|v|_g\leq a$. This proves \eqref{app:heat-profile}.
\end{proof}

\begin{lemma}[Uniform intensity bounds]
\label{aq:diagonal}
For every fixed $a\ge a_0$ from Lemma~\ref{aq:rank-projection},
its threshold $h_0$ can be decreased, depending only on
$(\mathcal M,g)$ and $a$, so that the following holds.
For $0<h\le h_0$ and every maximally $ah$-separated set,
there are constants $c_\rho,C_\rho>0$, depending only on
$(\mathcal M,g)$ and $a$, such that
\begin{equation}\label{aq:intensity-bounds}
 c_\rho n_h\le K_h(x,x)\le C_\rho n_h,
 \qquad x\in\mathcal M.
\end{equation}
The constants are uniform in $h$, the center set, and $x$.
\end{lemma}

\begin{proof}
We first bound the diagonal from above. The localization
estimate in Lemma~\ref{app:multiplier} gives
\[
 \|\Psi_h(x)\|_{\mathbb R^{2I}}^2
 \le C h^{-m}\sum_{i=1}^I
          (1+d(x,z_i)/h)^{-2m-6}.
\]
By the packing estimate in Lemma~\ref{app:packing}, the
contribution of $d(x,z_i)\le h$ is bounded, and the
contribution of $2^k h<d(x,z_i)\le2^{k+1}h$ is at most
\[
 C2^{-k(2m+6)}(1+2^{k+1}/a)^m,\qquad k\ge0.
\]
The resulting series converges uniformly for $a\ge1$.
Consequently $\|\Psi_h(x)\|_{\mathbb R^{2I}}^2\le Ch^{-m}$, and the lower
Gram bound in \eqref{aq:gram-bounds} yields
\[
 K_h(x,x)\le \frac{2}{c_*}\|\Psi_h(x)\|_{\mathbb R^{2I}}^2
 \le Ch^{-m}\le C_\rho n_h.
\]

For the lower bound, put
\[
 q_c(v):=(4\pi c)^{-m/2}e^{-|v|^2/(4c)},\qquad
 w_1(v):=q_1(v)-q_2(v),\qquad
 w_2(v):=q_2(v)-q_4(v).
\]
The ratios
\[
 \frac{q_1(v)}{q_2(v)}=2^{m/2}e^{-|v|^2/8},\qquad
 \frac{q_2(v)}{q_4(v)}=2^{m/2}e^{-|v|^2/16}
\]
show that
\[
 w_1(v)=0\ \Longleftrightarrow\ |v|^2=4m\log2,
 \qquad
 w_2(v)=0\ \Longleftrightarrow\ |v|^2=8m\log2.
\]
These zero sets are disjoint because $m\ge1$. For each fixed
$a$, compactness therefore gives
\begin{equation}\label{aq:profile-lower}
 \gamma_a:=\min_{|v|\le a}
       \bigl(|w_1(v)|^2+|w_2(v)|^2\bigr)>0.
\end{equation}
The local heat-kernel asymptotics in Lemma~\ref{app:heat-profile-lemma}
give
\[
 \begin{pmatrix}
 h^m(p_{h^2}-p_{2h^2})(\exp_z(hv),z)\\
 h^m(p_{2h^2}-p_{4h^2})(\exp_z(hv),z)
 \end{pmatrix}
 \longrightarrow
 V\begin{pmatrix}w_1(v)\\w_2(v)\end{pmatrix}
\]
uniformly in $z\in\mathcal M$ and $v\in T_z\mathcal M$ with
$|v|_g\le a$. The functions on the right are radial, so the
formula is independent of the choice of orthonormal coordinates
on $T_z\mathcal M$.

Decrease $h_0$ so that $ah_0$ is below the injectivity radius
and the uniform error in this vector limit is at most
$V\sqrt{\gamma_a}/2$. Maximality supplies, for every
$x\in\mathcal M$, a center $z_i$ with $d(x,z_i)\le ah$.
Writing $x=\exp_{z_i}(hv)$ with $|v|_g\le a$, we obtain
\[
 |\psi_{i,1}(x)|^2+|\psi_{i,2}(x)|^2
 \ge \frac{V^2\gamma_a}{4}h^{-m}.
\]
The upper Gram bound in \eqref{aq:gram-bounds} now implies
\[
 K_h(x,x)
 \ge\frac{2}{3C_*}\|\Psi_h(x)\|_{\mathbb R^{2I}}^2
 \ge\frac{V^2\gamma_a}{6C_*}h^{-m}
 \ge c_\rho n_h.
\]
This proves \eqref{aq:intensity-bounds}. The choice of $h_0$
can depend on the fixed $a$, and so can $c_\rho$ through
$\gamma_a$; no uniform lower bound as $a\to\infty$ is claimed.
\end{proof}

\paragraph{Approximation on the Voronoi cells.}

\begin{lemma}
\label{app:voronoi}
Suppose $ah\le R_0$, with $R_0$ as in
Lemma~\ref{app:ball-volume-lemma}. Let $V_1,\ldots,V_I$
be the geodesic Voronoi cells of $\mathcal Z$, with ties
assigned to the smallest index, and put $v_i=\mu(V_i)$.
These cells are measurable and satisfy
\begin{equation}
 \overline B(z_i,ah/2)\subset V_i\subset\overline B(z_i,ah),\qquad
 c (ah)^m\leq v_i\leq C (ah)^m,\qquad
 \operatorname{diam}(V_i)\leq2ah.
 \label{aq:cell-geometry}
\end{equation}
The constants depend only on $(\mathcal M,g)$.
\end{lemma}

\begin{proof}
Each cell is a finite intersection of sets defined by strict
or non-strict inequalities between continuous distance
functions, so it is measurable. If $d(x,z_i)\leq ah/2$ and
$j\ne i$, strict separation gives
\[
 d(x,z_j)\geq d(z_i,z_j)-d(x,z_i)>ah-d(x,z_i)\geq d(x,z_i).
\]
Thus $z_i$ is the unique closest center, proving the first
inclusion. The covering property of
Lemma~\ref{app:packing} places every point within $ah$ of
some center. Membership in $V_i$ therefore implies
$d(x,z_i)\leq ah$, proving the second inclusion.
The cell-mass bounds follow from
Lemma~\ref{app:ball-volume-lemma}, and the diameter estimate
follows from the triangle inequality. No regularity of the
cell boundaries is required.
\end{proof}

\begin{lemma}
\label{aq:holder-approximation-lemma}
Let $0<s\le1$ and $f\in C^{0,s}(\mathcal M)$. For the Voronoi
partition in \eqref{aq:cell-geometry}, set
\[
 A_hf(x):=\sum_{i=1}^I f(z_i)\mathbf1_{V_i}(x).
\]
Then
\begin{equation}\label{aq:holder-approximation}
 \|f-A_hf\|_{L^2(\mu)}^2
 \le (ah)^{2s}[f]_{C^{0,s}}^2.
\end{equation}
\end{lemma}

\begin{proof}
If $x\in V_i$, then $d(x,z_i)\le ah$. Hence
$|f(x)-f(z_i)|\le (ah)^s[f]_{C^{0,s}}$, and integration with
$\mu(\mathcal M)=1$ proves \eqref{aq:holder-approximation}.
\end{proof}

\begin{lemma}
\label{aq:sobolev-approximation-lemma}
Let $0<s\le1$ and $f\in H^s(\mathcal M)$. For the Voronoi
partition in \eqref{aq:cell-geometry}, let $U_i$ have law
$v_i^{-1}\mu|_{V_i}$, independently across cells. Using any
measurable representative of $f$, set
\[
 A_hf(x):=\sum_{i=1}^I f(U_i)\mathbf1_{V_i}(x).
\]
Then
\begin{equation}\label{aq:sobolev-approximation}
 \mathbb E\|f-A_hf\|_{L^2(\mu)}^2
 \le C_s (ah)^{2s}[f]_{H^s}^2.
\end{equation}
The constant depends only on $(\mathcal M,g)$ and $s$, and
is uniform over the scales, center sets, and observables under
consideration.
\end{lemma}

\begin{proof}
For the random approximation, every sampled value is finite
almost surely and
\[
 \mathbb E\|A_hf\|_{L^2(\mu)}^2
 =\sum_i v_i\mathbb E|f(U_i)|^2
 =\int_{\mathcal M}|f|^2\,\mathrm d\mu<\infty.
\]
Tonelli's theorem gives the exact identity
\begin{equation}\label{aq:anchor-identity}
 \mathbb E\|f-A_hf\|_{L^2(\mu)}^2
 =\sum_{i=1}^I\frac1{v_i}
   \iint_{V_i\times V_i}|f(x)-f(y)|^2
   \,\mathrm d\mu(x)\,\mathrm d\mu(y).
\end{equation}
If $0<s<1$, the diagonal has product measure zero, and
\eqref{aq:cell-geometry} gives
\begin{align*}
 \mathbb E\|f-A_hf\|_{L^2(\mu)}^2
 &\le C(ah)^{-m}(2ah)^{m+2s}
 \sum_i\iint_{(V_i\times V_i)\cap\{x\ne y\}}
 \frac{|f(x)-f(y)|^2}{d(x,y)^{m+2s}}
 \,\mathrm d\mu(x)\,\mathrm d\mu(y)\\
 &\le C_s (ah)^{2s}[f]_{H^s}^2.
\end{align*}
Here the sets $V_i\times V_i$ are pairwise disjoint, so their
integrals are bounded by the integral over $\mathcal M^2$.

For $s=1$, use \eqref{aq:anchor-identity}, the lower cell-mass
bound, and the diameter estimate to obtain
\begin{align*}
 \mathbb E\|f-A_hf\|_{L^2(\mu)}^2
 &\le C(ah)^{-m}
 \iint_{d(x,y)\le2ah}|f(x)-f(y)|^2
 \,\mathrm d\mu(x)\,\mathrm d\mu(y)\\
 &\le C(ah)^2[f]_{H^1}^2,
\end{align*}
where the last inequality is Lemma~\ref{app:energy}.
\end{proof}

\begin{proof}[Proof of Theorem~\ref{thm:quadrature-manifold}]
Choose $a_0$ and, for each fixed $a\ge a_0$, choose $h_0$
so that Lemmas~\ref{aq:rank-projection} and~\ref{aq:diagonal}
apply. These lemmas give the two construction estimates used
in the proof:
\[
 n_h=2I\asymp h^{-m},\qquad
 \rho_h(x):=K_h(x,x)\ge c_\rho n_h
 \quad\text{for every }x\in\mathcal M.
\]
We will show that the mean square error is bounded by
$(c_\rho n_h)^{-1}$ times the mean squared $L^2(\mu)$
approximation error. The cell approximation estimates will
then supply the factor $h^{2s}$.

Let $A_hf$ equal $f(z_i)$ on $V_i$ for the H\"older estimator,
and $f(U_i)$ on $V_i$ for the Sobolev estimator. These are the
approximations in Lemmas~\ref{aq:holder-approximation-lemma}
and~\ref{aq:sobolev-approximation-lemma}. In the latter case,
the points $U_i$ are independent of $\mathcal S_h$.
The estimator has the form
\begin{equation}\label{aq:residual-form}
 \widehat{\mathcal I}_h(f)
 =\int_{\mathcal M}A_hf\,\mathrm d\mu
  +\sum_{x\in\mathcal S_h}
    \frac{f(x)-A_hf(x)}{\rho_h(x)}.
\end{equation}
Thus its first term is the exact integral of the chosen
approximation, while the second term estimates the integral
of the residual. The factor $1/\rho_h$ compensates for the
DPP's sampling intensity.

In the Sobolev case, first condition on $U_1,\ldots,U_I$.
This fixes $A_hf$ and leaves the law of $\mathcal S_h$
unchanged. The residual belongs to $L^2(\mu)$ almost surely
by Lemma~\ref{aq:sobolev-approximation-lemma}; this also holds
for the deterministic H\"older approximation. Moreover,
\[
 \int_{\mathcal M}
 \left|\frac{f-A_hf}{\rho_h}\right|^2\rho_h\,\mathrm d\mu
 \le\frac1{c_\rho n_h}\|f-A_hf\|_{L^2(\mu)}^2<\infty.
\]
Therefore
\begin{align*}
 \mathbb E\bigl[\widehat{\mathcal I}_h(f)
       \mid U_1,\ldots,U_I\bigr]
 =\int_{\mathcal M}A_hf\,\mathrm d\mu
   +\int_{\mathcal M}(f-A_hf)\,\mathrm d\mu =\mathcal I(f).
\end{align*}
For the H\"older estimator the same calculation holds without
conditioning. In particular, the correction removes the
integration error of the approximation exactly in expectation,
for almost every realization of the auxiliary samples.

To bound the remaining fluctuation, observe that
\begin{align*}
 \operatorname{Var}\bigl[
   \widehat{\mathcal I}_h(f)\mid U_1,\ldots,U_I\bigr]
 &\le\int_{\mathcal M}
       \frac{|f-A_hf|^2}{\rho_h}\,\mathrm d\mu\\
 &\le\frac1{c_\rho n_h}\|f-A_hf\|_{L^2(\mu)}^2.
\end{align*}
The conditional mean is the fixed number $\mathcal I(f)$.
Consequently the variance of the conditional mean is zero,
and averaging the conditional variance gives
\begin{equation}\label{aq:residual-variance}
 \mathbb E\left|
   \widehat{\mathcal I}_h(f)-\mathcal I(f)\right|^2
 \le\frac1{c_\rho n_h}
       \mathbb E\|f-A_hf\|_{L^2(\mu)}^2.
\end{equation}
For the deterministic approximation, the expectation on the
right is omitted. This explains why the random cell values
contribute through their approximation error, without an
additional variance term for their weighted sum.

The cells have diameter at most $2ah$. By
\eqref{aq:holder-approximation} and
\eqref{aq:sobolev-approximation}, respectively, the right-hand
side of \eqref{aq:residual-variance} is at most
\[
 C_s n_h^{-1}h^{2s}
 \begin{cases}
 [f]_{C^{0,s}}^2,&\text{for the estimator in
 \eqref{eq:est-holder}},\\
 [f]_{H^s}^2,&\text{for the estimator in
 \eqref{eq:est-sobolev}}.
 \end{cases}
\]
Here the fixed factor $a^{2s}$ has been absorbed into the
constant. The relation $n_h\asymp h^{-m}$ converts
$n_h^{-1}h^{2s}$ into $n_h^{-1-2s/m}$ and proves both mean
square error bounds. The DPP variance bound and the intensity
lower bound together yield the factor $n_h^{-1}$. The factor
$n_h^{-2s/m}$ comes from approximation on cells of diameter
of order $h$.
\end{proof}

\section{Proofs from Section \ref{sec:graphs}}\label{app:main}
\subsection{Proofs of Theorems \ref{thm:heat-epsilon} and \ref{thm:heat-knn}}
\subsubsection{Geometry of the point cloud}
\label{app:geometry}

We use the following consequences of the compactness and
smoothness of $\M$. There exist positive structural constants
$c,C,\ell_0,\kappa_0$ such that
\begin{align}
 c\min\{R^m,1\}
 &\leq \mu(B_{\M}(x,R))
 \leq C\min\{R^m,1\},
 \qquad R>0,
 \label{eq:manifold-volume}\\
 |x-y|
 &\leq d_{\M}(x,y)
 \leq \kappa_0|x-y|,
 \qquad |x-y|\leq\ell_0.
 \label{eq:local-chord}
\end{align}
For sufficiently small $R$, \eqref{eq:manifold-volume}
follows from the uniform comparability of the Riemannian
metric and the Euclidean metric in local coordinates.
The bounds for larger $R$ follow from compactness and
connectedness.

For \eqref{eq:local-chord}, the first inequality follows
from the definition of $d_{\M}$. Since $\M$ is compact
and smoothly embedded, any two points with sufficiently
small Euclidean distance lie in a common coordinate
neighbourhood, where they can be joined by a curve of
length at most $\kappa_0|x-y|$

\begin{lemma}\label{lem:cloud}
Fix $0<\alpha\leq\beta<\infty$ and a sufficiently small $\eta_0>0$.
There are constants $r_0,c,C>0$, depending on these parameters and the
structural data, such that the following holds. Let $0<r\leq r_0$ and
$K=Nr^m\geq1$. There is an event $\mathcal E_{N,r}$ with
\begin{equation}\label{eq:cloud-probability}
 \P(\mathcal E_{N,r})\geq1-Cr^{-m}e^{-cK}
\end{equation}
on which there exist a probability measure $\widetilde\mu$ on $\M$ and
a measurable partition $V_1,\ldots,V_N$ satisfying
\begin{equation}\label{eq:cloud-partition}
 \widetilde\mu(V_i)=\frac1N,\quad
 c\leq\frac{\mathrm d\widetilde\mu}{\mathrm d\mu}\leq C,
 \quad \sup_{x\in V_i}d_{\M}(x,x_i)\leq\eta_0r.
\end{equation}
Every graph satisfying
\begin{equation}\label{eq:edge-sandwich}
 \one_{\{|x_i-x_j|\leq\alpha r\}}
       \leq A_{ij}\leq
 \one_{\{|x_i-x_j|\leq\beta r\}}
\end{equation}
then satisfies \eqref{eq:graph-geometry}--\eqref{eq:graph-volume-lower}
with $K=Nr^m$ and $\delta_N=C_\delta r$. The constants are uniform
over all graphs satisfying \eqref{eq:edge-sandwich}.
\end{lemma}
\begin{proof}
Put $b=\eta_0r/4$. Choose a deterministic maximal strictly
$b$-separated set $\{y_1,\ldots,y_J\}$ in $(\M,d_{\M})$, and let
$U_1,\ldots,U_J$ be the corresponding Voronoi cells, with ties resolved by the order of the centers.
Each $U_\ell$ contains the open ball of radius $b/2$ about $y_\ell$
and is contained in the closed ball of radius $b$. By
\eqref{eq:manifold-volume} and \eqref{eq:model},
\begin{equation}\label{eq:geometric-cells}
 \mu(U_\ell)\asymp b^m,\quad
 \nu(U_\ell)\asymp b^m,\quad J\leq Cb^{-m}.
\end{equation}

Let $M_\ell=\#\{i:x_i\in U_\ell\}$. Since the sets
$U_\ell$ are deterministic, \eqref{eq:model} implies that
$M_\ell$ is binomial with mean $N\nu(U_\ell)$.
A multiplicative Chernoff bound gives
\[
 \P\left(M_\ell\notin
   [N\nu(U_\ell)/2,2N\nu(U_\ell)]\right)
 \leq 2e^{-cN\nu(U_\ell)}.
\]
Taking a union bound and using \eqref{eq:geometric-cells},
we obtain
\[
 \P\left(\text{some }M_\ell\notin
   [N\nu(U_\ell)/2,2N\nu(U_\ell)]\right)
 \leq Cb^{-m}e^{-cNb^m}
 \leq Cr^{-m}e^{-cK}.
\]

Let $\mathcal E_{N,r}$ denote the complementary event.
On this event every cell is nonempty. Define
\begin{equation}\label{eq:adjusted-density}
 \frac{\mathrm d\widetilde\mu}{\mathrm d\mu}(x)
 :=\frac{M_\ell}{N\mu(U_\ell)}
 \quad\text{for }x\in U_\ell.
\end{equation}
Since $\sum_{\ell=1}^J M_\ell=N$, this defines a
probability measure. Moreover, by the definition of
$\mathcal E_{N,r}$ and \eqref{eq:model},
\[
\frac{\rho_{\min}}{2}
\leq\frac{\mathrm d\widetilde\mu}{\mathrm d\mu}
\leq 2\rho_{\max}
\quad \mu\text{-a.e.}
\]

Since $\mu$ is nonatomic, we may divide each $U_\ell$
into $M_\ell$ measurable sets of equal $\mu$-measure
and assign one to each sample point in $U_\ell$.
Denote the set assigned to $x_i$ by $V_i$. Then
$\mu(V_i)=\mu(U_\ell)/M_\ell$, so
\eqref{eq:adjusted-density} gives
$\widetilde\mu(V_i)=1/N$. Moreover, if $x\in V_i$, then $x$ and $x_i$ both
belong to $U_\ell$, and hence
$d_{\M}(x,x_i)\leq 2b\leq\eta_0r$.
This proves \eqref{eq:cloud-partition}.

Define $F(x)=x_i$ for $x\in V_i$. For every $x\in\M$ and $R>\eta_0r$,
\begin{equation}\label{eq:empirical-balls}
 c\mu(B_{\M}(x,R-\eta_0r))
 \leq\frac1N\#\{i:d_{\M}(x,x_i)\leq R\}
 \leq C\mu(B_{\M}(x,R+\eta_0r)).
\end{equation}

Since $\widetilde\mu(V_i)=1/N$, the middle quantity in
\eqref{eq:empirical-balls} equals
$\widetilde\mu(F^{-1}(B_{\M}(x,R)))$.
Moreover, $d_{\M}(y,F(y))\leq\eta_0r$ implies
\[
B_{\M}(x,R-\eta_0r)
\subseteq F^{-1}(B_{\M}(x,R))
\subseteq B_{\M}(x,R+\eta_0r).
\]
The first inclusion holds for $R>\eta_0r$, while the
second holds for every $R>0$. The bounds on
$\mathrm d\widetilde\mu/\mathrm d\mu$ now give
\eqref{eq:empirical-balls}.

We next prove the metric comparison. Decrease $r_0$ so that
$\beta r_0\leq\ell_0$. Applying \eqref{eq:local-chord} to each edge
of a graph path gives
\begin{equation}\label{eq:metric-one-direction}
 d_{\M}(x_i,x_j)\leq\kappa_0 d_N(i,j)
\end{equation}
whenever a path exists.

For the reverse direction, fix $i,j$ and put
$D_0=d_{\M}(x_i,x_j)$. If $D_0\leq\alpha r$, then
$|x_i-x_j|\leq\alpha r$, so there is a direct edge.
Otherwise, let $\gamma:[0,D_0]\to\M$ be a minimizing
geodesic from $x_i$ to $x_j$.
Set $L=\lceil4D_0/(\alpha r)\rceil$ and consider the points
\[
 y_k=\gamma(kD_0/L),\quad 0\leq k\leq L.
\]
Since the sets $U_\ell$ cover $\M$, each contains at least
one sample point, and each has diameter at most $2b$,
we can choose a sample point within distance $2b$ of
each $y_k$. At the endpoints, we retain $x_i$ and $x_j$.

The geodesic distance between consecutive chosen sample
points is at most
\[
 \frac{D_0}{L}+4b
 \leq\frac{\alpha r}{4}+4b<\alpha r,
\]
provided $\eta_0$ is sufficiently small. Hence consecutive
sample points are joined by an edge, and they form a graph
path after repeated vertices are omitted. The total length
of this path is at most
\[
 D_0+4bL\leq CD_0,
\]
where we used $D_0>\alpha r$. Thus the graph is connected
and $d_N(i,j)\leq C d_{\M}(x_i,x_j)$. Together with
\eqref{eq:metric-one-direction}, this proves the metric
part of \eqref{eq:graph-geometry} and bounds the graph
diameter.

The metric comparison and \eqref{eq:empirical-balls} imply
\[
 \mu_N(B_N(i,R))\leq C(R+C\eta_0r)^m
\]
for every $R>0$. They also give
$\mu_N(B_N(i,R))\geq cR^m$ for $C_\delta r\leq R\leq R_0$,
after increasing $C_\delta$ and choosing $R_0$ fixed. Finally, by
\eqref{eq:edge-sandwich}, $d_i$ is bounded below by the number of samples
in $B_{\M}(x_i,\alpha r)$ and above by the number in
$B_{\M}(x_i,\kappa_0\beta r)$. Equations
\eqref{eq:empirical-balls} and \eqref{eq:manifold-volume} give
$cK\leq d_i\leq CK$. All choices of $\eta_0$ and $r_0$ depend only
on the fixed parameters, as required.
\end{proof}

\begin{lemma}\label{lem:knn-radii}
Recall that $R_i$ denotes the ambient Euclidean distance
from $x_i$ to its $k$th nearest neighbour. There are positive
structural constants $\alpha,\beta,r_0,c$ with the following
property. If $r=(k/N)^{1/m}\leq r_0$, then with probability
at least $1-2Ne^{-ck}$,
\begin{equation}\label{eq:radii-bound}
 \alpha r<R_i\leq\beta r\quad(1\leq i\leq N),
\end{equation}
and \eqref{eq:edge-sandwich} holds for $A^k$.
\end{lemma}

\begin{proof}
By \eqref{eq:model}, \eqref{eq:manifold-volume}, and
\eqref{eq:local-chord}, there are $c_1,C_1,R_1>0$ such that
\[
 c_1R^m\leq\nu\{y:|y-x|\leq R\}\leq C_1R^m,
 \quad x\in\M,\ \  0<R\leq R_1.
\]
Conditionally on $x_i=x$, we have
\[
\#\{j\ne i:|x_j-x|\leq R\}
\sim\operatorname{Bin}
\bigl(N-1,\nu\{y:|y-x|\leq R\}\bigr).
\]
Choose $\alpha>0$ sufficiently small that
$C_1\alpha^m\leq1/4$, and $\beta>0$ sufficiently
large that $c_1\beta^m\geq8$. Decrease $r_0$
so that $\beta r_0\leq R_1$. Since $r^m=k/N$
and $(N-1)/N\geq1/2$, we have
\[
(N-1)\nu\{y:|y-x|\leq\alpha r\}\leq k/4,
\quad
(N-1)\nu\{y:|y-x|\leq\beta r\}\geq4k.
\]
Chernoff bounds therefore give, uniformly in $x$,
\[
 \P\big(\#\{j\ne i:|x_j-x_i|\leq\alpha r\}
 \geq k\mid x_i=x\big)
 \leq e^{-ck},
\]
and
\[
 \P\big(\#\{j\ne i:|x_j-x_i|\leq\beta r\}
 <k\mid x_i=x\big)
 \leq e^{-ck}.
\]
Integrating over $x_i$ and taking a union bound over
$i$ shows that $\alpha r<R_i\leq\beta r$
simultaneously for all $i$, with probability at least
$1-2Ne^{-ck}$. This proves \eqref{eq:radii-bound}.

On this event, $|x_i-x_j|\leq\alpha r$ with $j\ne i$
implies $j\in\mathcal N_k(i)$, while
$j\in\mathcal N_k(i)$ implies $|x_i-x_j|\leq\beta r$.
Consequently, \eqref{eq:edge-sandwich} holds for
$A^k$, even when distance ties occur.
\end{proof}

\subsubsection{Heat kernels of local graph walks}\label{app:heat}

Now, we establish the heat-kernel estimates required for the analysis of the determinantal projections.
Our approach is to represent the random walk on the graph as a reversible Markov kernel on the underlying manifold, using the partition constructed in Lemma\ref{lem:cloud}.
We first establish Gaussian estimates for local Markov
kernels on the underlying manifold and transfer these
estimates to the discrete graph random walk using the
partition constructed in Appendix~B. We then pass
from the discrete-time random walk to the continuous-time heat semigroup through its Poisson representation.

For a measured metric space $(Y,d_Y,\lambda)$, let
$V_Y(x,R)=\lambda(B_Y(x,R))$, and define
\begin{equation}\label{eq:difference-gradient}
 |\nabla_b u|^2(x):=\frac1{V_Y(x,b)}
               \int_{B_Y(x,b)}|u(y)-u(x)|^2\dd\lambda(y).
\end{equation}
For a Markov kernel $p_1$ that is symmetric with respect to $\lambda$, we say that $p_1$ is weakly $(b,b')$-compatible if
\begin{equation}\label{eq:compatibility}
 \frac{c}{V_Y(x,b)}\one_{\{d_Y(x,y)\leq b\}}
 \leq p_1(x,y)\leq
 \frac{C}{V_Y(x,b')}\one_{\{d_Y(x,y)\leq b'\}},
 \quad 0<b\leq b'<\infty.
\end{equation}

We begin by recalling a consequence of the Gaussian
heat-kernel estimates established by
\cite{MuruganSaloffCoste2023}, which will serve as the
main analytic input in this appendix.

\begin{proposition}[\cite{MuruganSaloffCoste2023},
Theorem 9.10 and Lemmas 4.11 and 4.16]
\label{prop:msc}
Let $(Y,d_Y,\lambda)$ be a complete geodesic metric
measure space of infinite diameter satisfying volume
doubling. Suppose that, for some fixed $b>0$, there
exist constants $C,\Lambda>0$ such that
\begin{equation}\label{eq:nonlocal-poincare}
    \int_D |u-u_D|^2\dd\lambda
    \leq CR^2\int_{\Lambda D}|\nabla_bu|^2\dd\lambda,
    \quad D=B_Y(x,R),\quad R\geq b.
\end{equation}
Here $u_D$ denotes the average of $u$ over $D$,
and $\Lambda D = B_{Y}(x,\Lambda R)$.

If $p_1$ is a symmetric Markov kernel satisfying
\eqref{eq:compatibility}, then its iterated densities
satisfy
\begin{align}
    p_j(x,y)
    &\leq \frac{C_1}{V_Y(x,\sqrt j)}
    \exp\left(-c_1\frac{d_Y(x,y)^2}{j}\right),
    \label{eq:msc-upper}\\
    p_j(x,y)
    &\geq \frac{c_2}{V_Y(x,\sqrt j)}
    \exp\left(-C_2\frac{d_Y(x,y)^2}{j}\right),
    \quad d_Y(x,y)\leq c_3j,
    \label{eq:msc-lower}
\end{align}
for all integers $j\geq j_0$, where $j_0$ and the
constants in these estimates depend only on the
volume-doubling and Poincaré constants, on
$b,b'$ and on the constants in \eqref{eq:compatibility}.
\end{proposition}

We explain how the preceding formulation follows from
\cite{MuruganSaloffCoste2023}. In the terminology of
that paper, a symmetric Markov kernel $p_1$ is
$(b,b')$-compatible if it satisfies the two-sided
locality bounds \eqref{eq:compatibility} and, in
addition, there exists $\alpha>0$ such that
\[
    p_2(x,y)\geq \alpha p_1(x,y),
\]
where $p_2$ denotes the two-step transition density.
A kernel satisfying only the symmetry and locality
conditions is called weakly $(b,b')$-compatible.

Theorem 9.10 of \cite{MuruganSaloffCoste2023}
establishes Gaussian estimates under the stronger
compatibility assumption. To apply this result to
weakly compatible kernels, we use Lemma 4.11,
which shows that a sufficiently large convolution
power $p_k$ is compatible, provided the space is
quasi-$b_0$-geodesic for some $b_0<b$.
Since every geodesic space satisfies this condition,
Theorem 9.10 applies to $p_k$ and yields Gaussian
bounds for its iterates. Finally, Lemma 4.16
transfers these bounds back to the original kernel
$p_1$, giving the stated conclusion.

\begin{lemma}\label{lem:manifold-kernel}
Let $\sigma$ be a probability measure on $\M$ satisfying
\[
    c_0\leq\frac{\mathrm d\sigma}{\mathrm d\mu}
    \leq C_0
\]
for some fixed constants $0<c_0\leq C_0<\infty$.
Suppose that a $\sigma$-symmetric Markov kernel $p_1$
satisfies
\begin{equation}\label{eq:manifold-kernel-sandwich}
    c_1r^{-m}\one_{\{d_{\M}(x,y)\leq br\}}
    \leq p_1(x,y)\leq
    C_1r^{-m}\one_{\{d_{\M}(x,y)\leq b'r\}}
\end{equation}
for fixed $c_1,C_1>0$ and $0<b\leq b'$.
Then there exist $r_0>0$, an integer $j_0\geq1$,
and constants $c,C,c'>0$ such that, for every
$0<r\leq r_0$, the iterated kernels satisfy
\begin{align}
    p_j(x,y)
    &\leq\frac{C}{\min\{(r\sqrt j)^m,1\}}
    \exp\left(-c\frac{d_{\M}(x,y)^2}{r^2j}\right),
    \quad j\geq1,
    \label{eq:manifold-discrete-upper}\\
    p_j(x,y)
    &\geq\frac{c}{\min\{(r\sqrt j)^m,1\}}
    \exp\left(-C\frac{d_{\M}(x,y)^2}{r^2j}\right),
    \quad j\geq j_0,\quad d_{\M}(x,y)\leq c'rj.
    \label{eq:manifold-discrete-lower}
\end{align}
The estimates hold $\sigma\otimes\sigma$-almost everywhere,
with constants depending only on $(\M,g)$,
$c_0,C_0,c_1,C_1,b$, and $b'$, independently
of $r$, $\sigma$, and $p_1$.
\end{lemma}

\begin{proof}
We verify the hypotheses of Proposition~\ref{prop:msc}
on the product space $(Y_r,d_r,\lambda)$, where 
\begin{equation}\label{eq:product-space}
    Y_r:=(\M,g_r)\times\R,\quad d_r((x,z),(y,w))
    :=\sqrt{r^{-2}d_{\M}(x,y)^2+|z-w|^2}, \quad
    \lambda:=\sigma\otimes\mathrm{Leb}, 
\end{equation}
where $g_r=r^{-2}g$. The product with $\mathbb{R}$ ensures that $Y_r$ has
an infinite diameter, as required by Proposition~\ref{prop:msc}.
Unless otherwise specified, gradients
on $Y_r$ are taken with respect to the product metric, so that
\[
    |\nabla v|^2
    =|\nabla_{g_r}v|^2+|\partial_zv|^2
    =r^2|\nabla_gv|^2+|\partial_zv|^2.
\]

\paragraph{Volume doubling.}
The space $Y_r$ is complete and geodesic. Moreover, a ball
of radius $R$ in $Y_r$ contains the product of the factor
balls of radius $R/\sqrt{2}$ and is contained in the product
of the factor balls of radius $R$. Using
\eqref{eq:manifold-volume} and the bounds on
$\mathrm d\sigma/\mathrm d\mu$, we obtain
\begin{equation}\label{eq:product-volume}
    \lambda(B_{d_r}((x,z),R))
    \asymp R\min\{(rR)^m,1\}.
\end{equation}
The comparison constants depend only on $(\M,g)$ and
$c_0,C_0$, independently of $r$ and $\sigma$.
In particular, $Y_r$ satisfies volume doubling
with a uniform constant.

\paragraph{Poincar\'e inequalities on the compact factor:}
For a measurable set $A\subseteq\M$ of positive measure,
write
\[
    w_{A,\sigma}:=\frac1{\sigma(A)}\int_Aw\,\dd\sigma,
    \quad
    w_{A,\mu}:=\frac1{\mu(A)}\int_Aw\,\dd\mu.
\]
Choose $\rho_*>0$ smaller than the injectivity radius
of $\M$. By decreasing $\rho_*$ if necessary, the metric
coefficients and volume densities in normal coordinates
on $B_{\M}(x,\rho_*)$ are uniformly comparable to their
Euclidean counterparts, independently of $x$.
The Euclidean Poincar\'e inequality therefore gives
\[
    \int_{B_{\M}(x,\rho)}
    |w-w_{B_{\M}(x,\rho),\mu}|^2\,\dd\mu
    \leq C\rho^2
    \int_{B_{\M}(x,\rho)}|\nabla_gw|^2\,\dd\mu,
    \quad 0<\rho\leq\rho_*.
\]
Since the average minimizes the $L^2$ distance to
constants, the bounds on $\mathrm d\sigma/\mathrm d\mu$
imply
\begin{align*}
    &\int_{B_{\M}(x,\rho)}
    |w-w_{B_{\M}(x,\rho),\sigma}|^2\,\dd\sigma
    \leq
    \int_{B_{\M}(x,\rho)}
    |w-w_{B_{\M}(x,\rho),\mu}|^2\,\dd\sigma\\
    &\quad\leq C_0
    \int_{B_{\M}(x,\rho)}
    |w-w_{B_{\M}(x,\rho),\mu}|^2\,\dd\mu
    \leq C\rho^2
    \int_{B_{\M}(x,\rho)}|\nabla_gw|^2\,\dd\sigma.
\end{align*}

For larger radii, we use the global Poincar\'e inequality
on the compact, connected manifold $\M$. By the same
comparison of measures,
\[
    \int_{\M}|w-w_{\M,\sigma}|^2\,\dd\sigma
    \leq C\int_{\M}|\nabla_gw|^2\,\dd\sigma.
\]
Fix $\Lambda_{\M}>1$ such that
$\Lambda_{\M}\rho_*>\operatorname{diam}(\M)$.
For $\rho\geq\rho_*$, the ball
$B_{\M}(x,\Lambda_{\M}\rho)$ contains all of $\M$.
Consequently,
\begin{align*}
    \int_{B_{\M}(x,\rho)}
    |w-w_{B_{\M}(x,\rho),\sigma}|^2\,\dd\sigma
    \leq
    \int_{\M}|w-w_{\M,\sigma}|^2\,\dd\sigma
    \leq C\int_{\M}|\nabla_gw|^2\,\dd\sigma\\
    \leq C\rho^2
    \int_{B_{\M}(x,\Lambda_{\M}\rho)}
    |\nabla_gw|^2\,\dd\sigma,
\end{align*}
where the constant in the last inequality includes
the fixed factor $\rho_*^{-2}$. Combining the two
ranges of $\rho$, we obtain
\begin{equation}\label{eq:mk-factor-poincare}
    \int_{B_{\M}(x,\rho)}
    |w-w_{B_{\M}(x,\rho),\sigma}|^2\,\dd\sigma
    \leq C\rho^2
    \int_{B_{\M}(x,\Lambda_{\M}\rho)}
    |\nabla_gw|^2\,\dd\sigma,
    \quad \rho>0.
\end{equation}

Let $B_r(x,R)$ denote the ball in $(\M,g_r)$.
Since $B_r(x,R)=B_{\M}(x,rR)$ and
$|\nabla_{g_r}w|^2=r^2|\nabla_gw|^2$,
\eqref{eq:mk-factor-poincare} becomes
\begin{equation}\label{eq:mk-rescaled-poincare}
    \int_{B_r(x,R)}
    |w-w_{B_r(x,R),\sigma}|^2\,\dd\sigma
    \leq CR^2
    \int_{B_r(x,\Lambda_{\M}R)}
    |\nabla_{g_r}w|^2\,\dd\sigma,
    \quad R>0.
\end{equation}
The constants are independent of $r$ and $\sigma$.
In particular, the estimate holds also for radii
exceeding $\operatorname{diam}(\M)/r$: in this range,
the compact factor is all of $\M$, and the rescaled
global Poincar\'e constant is of order
$r^{-2}\lesssim R^2$.

\paragraph{Poincar\'e inequalities on product balls.}
We next extend \eqref{eq:mk-rescaled-poincare} to balls
in the product space $Y_r$.
Fix $\xi_0=(x_0,z_0)\in Y_r$ and $R>0$, and set
\[
    D=B_{d_r}(\xi_0,R),\quad
    B=B_r(x_0,R),\quad
    B^+=B_r(x_0,\Lambda_{\M}R),
\]
and $I=(z_0-R,z_0+R)$. Write $Q=B\times I$ and
$Q^+=B^+\times I$. With
$\Lambda=(\Lambda_{\M}^2+1)^{1/2}$, we have
\[
    D\subseteq Q\subseteq Q^+\subseteq\Lambda D.
\]

For a smooth function $v$ on $Y_r$, define
\[
    a(t):=\frac1{\sigma(B)}
    \int_Bv(y,t)\,\dd\sigma(y),
    \quad
    v_Q:=\frac1{\lambda(Q)}\int_Qv\,\dd\lambda
    =\frac1{|I|}\int_Ia(t)\,\dd t.
\]
Since
$\int_B(v(y,t)-a(t))\,\dd\sigma(y)=0$, expanding
the square gives
\begin{align*}
    \int_Q|v-v_Q|^2\,\dd\lambda
    &=\int_I\int_B
    |v(y,t)-a(t)|^2\,\dd\sigma(y)\,\dd t
    +
    \sigma(B)\int_I|a(t)-v_Q|^2\,\dd t.
\end{align*}
By \eqref{eq:mk-rescaled-poincare}, the first term
is bounded by
\[
    CR^2\int_I\int_{B^+}
    |\nabla_{g_r}v(y,t)|^2\,\dd\sigma(y)\,\dd t.
\]
For the second term, the one-dimensional Poincar\'e
inequality and Cauchy--Schwarz yield
\begin{align*}
    \sigma(B)\int_I|a(t)-v_Q|^2\,\dd t
    \leq CR^2\sigma(B)\int_I|a'(t)|^2\,\dd t
    \leq CR^2\int_I\int_B
    |\partial_tv(y,t)|^2\,\dd\sigma(y)\,\dd t,
\end{align*}
where
\[
    a'(t)=\frac1{\sigma(B)}
    \int_B\partial_tv(y,t)\,\dd\sigma(y).
\]
Combining these bounds, we obtain
\[
    \int_Q|v-v_Q|^2\,\dd\lambda
    \leq CR^2\int_{Q^+}|\nabla v|^2\,\dd\lambda.
\]
Writing $v_D$ for the $\lambda$-average over $D$,
the minimizing property of the average and the
inclusions above give
\begin{equation}\label{eq:product-poincare}
    \int_D|v-v_D|^2\,\dd\lambda
    \leq\int_D|v-v_Q|^2\,\dd\lambda
    \leq\int_Q|v-v_Q|^2\,\dd\lambda
    \leq CR^2\int_{\Lambda D}
    |\nabla v|^2\,\dd\lambda.    
\end{equation}
This holds for every $R>0$, with constants
independent of $r$ and $\sigma$.

\paragraph{Construction of the smoothing kernel:}
To obtain the nonlocal Poincar\'e inequality in
Proposition~\ref{prop:msc}, we use a local smoothing
operator on $Y_r$.
Fix $b_1:=\frac12\min\{b,1\}$ and $b_0:=\frac{b_1}{2}$.

The injectivity radius of $Y_r$ is
$r^{-1}\operatorname{inj}(\M,g)$. We choose $r_0>0$
sufficiently small that this radius exceeds $2b_0$
whenever $0<r\leq r_0$.

Let $\chi\in C^\infty([0,\infty))$ be a smooth cutoff satisfying
$
    0\leq\chi\leq1,\ 
    \chi=1\ \text{on }[0,1/4],\text{ and }
    \chi=0\ \text{on }[1,\infty).
$
For $\xi,\zeta\in Y_r$, define
\[
    \Phi(\xi,\zeta)
    :=\chi\left(\frac{d_r(\xi,\zeta)^2}{b_0^2}\right),
    \  \ 
A(\xi):=\int_{Y_r}\Phi(\xi,\zeta)\,\dd\lambda(\zeta), \text{ and } J(\xi,\zeta):=\frac{\Phi(\xi,\zeta)}{A(\xi)}.
\]
The function $\Phi$ is smooth, since the squared
distance is smooth within the chosen injectivity
radius and the cutoff vanishes before reaching
the cut locus.

Put
\[
    V_\xi:=\lambda(B_{d_r}(\xi,b_0)).
\]
By the properties of $\chi$ and volume doubling,
\[
    \lambda(B_{d_r}(\xi,b_0/2))
    \leq A(\xi)\leq V_\xi,
    \quad
    cV_\xi\leq A(\xi)\leq V_\xi.
\]
Moreover,
\[
    |\nabla_\xi\Phi(\xi,\zeta)|
    \leq\frac{C}{b_0}
    \one_{\{d_r(\xi,\zeta)<b_0\}}.
\]
Differentiation under the integral is justified
locally by the compact support of $\Phi$ and the
bounded density of $\sigma$. Thus
\[
    \nabla A(\xi)
    =\int_{Y_r}\nabla_\xi\Phi(\xi,\zeta)
    \,\dd\lambda(\zeta),
    \quad
    |\nabla A(\xi)|\leq\frac{C}{b_0}V_\xi.
\]
Applying the quotient rule,
\[
    \nabla_\xi J(\xi,\zeta)
    =\frac{\nabla_\xi\Phi(\xi,\zeta)}{A(\xi)}
    -\frac{\Phi(\xi,\zeta)\nabla A(\xi)}{A(\xi)^2}.
\]
Consequently,
\begin{equation}\label{eq:smoothing-kernel}
\begin{aligned}
    \int_{Y_r}J(\xi,\zeta)\,\dd\lambda(\zeta)=1, \quad
    0\leq J(\xi,\zeta)
    \leq\frac{C}{V_\xi}
    \one_{\{d_r(\xi,\zeta)<b_0\}}, \quad
    |\nabla_\xi J(\xi,\zeta)|
    \leq\frac{C}{b_0V_\xi}
    \one_{\{d_r(\xi,\zeta)<b_0\}}.
\end{aligned}
\end{equation}
Notice that the derivatives fall only on $\Phi$
and its integral $A$. In particular, no regularity
of $\mathrm d\sigma/\mathrm d\mu$ beyond boundedness
is required.

\paragraph{The nonlocal Poincar\'e inequality.}
For $u\in L^2_{\mathrm{loc}}(Y_r,\lambda)$, define
\[
    Su(\xi):=\int_{Y_r}
    J(\xi,\zeta)u(\zeta)\,\dd\lambda(\zeta).
\]
The smoothness and compact support properties of
$J$ imply that $Su$ is smooth.
By \eqref{eq:smoothing-kernel} and Jensen's inequality,
\begin{align*}
    |u(\xi)-Su(\xi)|^2
    \leq\int_{Y_r}J(\xi,\zeta)
    |u(\zeta)-u(\xi)|^2\,\dd\lambda(\zeta)
    \leq\frac{C}{V_\xi}
    \int_{B_{d_r}(\xi,b_0)}
    |u(\zeta)-u(\xi)|^2\,\dd\lambda(\zeta)\\
    =C|\nabla_{b_0}u|^2(\xi).
\end{align*}
On the other hand, differentiating the first
identity in \eqref{eq:smoothing-kernel} gives
\[
    \int_{Y_r}\nabla_\xi J(\xi,\zeta)
    \,\dd\lambda(\zeta)=0.
\]
Therefore,
\[
    \nabla Su(\xi)
    =\int_{Y_r}\nabla_\xi J(\xi,\zeta)
    \big(u(\zeta)-u(\xi)\big)\,\dd\lambda(\zeta).
\]
By \eqref{eq:smoothing-kernel} and Cauchy--Schwarz,
\begin{align*}
    &|\nabla Su(\xi)|^2
    \leq\frac{C}{b_0^2V_\xi^2}
    \left(
    \int_{B_{d_r}(\xi,b_0)}
    |u(\zeta)-u(\xi)|\,\dd\lambda(\zeta)
    \right)^2\\
    &\leq\frac{C}{b_0^2V_\xi}
    \int_{B_{d_r}(\xi,b_0)}
    |u(\zeta)-u(\xi)|^2\,\dd\lambda(\zeta)
    =\frac{C}{b_0^2}
    |\nabla_{b_0}u|^2(\xi).
\end{align*}
Since $b_0$ is fixed, we have proved
\begin{equation}\label{eq:smoothing-estimates}
    |u-Su|^2+|\nabla Su|^2
    \leq C|\nabla_{b_0}u|^2,
\end{equation}
with constants independent of $r$ and $\sigma$.

Let $D=B_{d_r}(\xi_0,R)$, where $R\geq b_1$.
Applying \eqref{eq:product-poincare} to $Su$,
and then \eqref{eq:smoothing-estimates}, gives
\begin{align*}
    \int_D|u-u_D|^2\,\dd\lambda
    \leq\int_D|u-(Su)_D|^2\,\dd\lambda
    \leq2\int_D|u-Su|^2\,\dd\lambda
    +2\int_D|Su-(Su)_D|^2\,\dd\lambda\\
    \leq C\int_D|\nabla_{b_0}u|^2\,\dd\lambda
    +CR^2\int_{\Lambda D}|\nabla Su|^2\,\dd\lambda
    \leq C\left(1+\frac{R^2}{b_0^2}\right)
    \int_{\Lambda D}|\nabla_{b_0}u|^2\,\dd\lambda\\
    \leq CR^2
    \int_{\Lambda D}|\nabla_{b_0}u|^2\,\dd\lambda.
\end{align*}
In the last inequality, we used
$R\geq b_1=2b_0$.

Since $b_1=2b_0$, volume doubling also gives
\begin{align*}
    |\nabla_{b_0}u|^2(\xi)
    \leq
    \frac{\lambda(B_{d_r}(\xi,b_1))}
         {\lambda(B_{d_r}(\xi,b_0))}
    |\nabla_{b_1}u|^2(\xi)
    \leq C|\nabla_{b_1}u|^2(\xi).
\end{align*}
Combining the preceding estimates, we obtain
\[
    \int_D|u-u_D|^2\,\dd\lambda
    \leq CR^2\int_{\Lambda D}
    |\nabla_{b_1}u|^2\,\dd\lambda,
    \quad R\geq b_1.
\]
This is precisely \eqref{eq:nonlocal-poincare}
at scale $b_1$, with constants uniform in $r$
and $\sigma$.

\paragraph{Application to the product kernel.}
We now construct a Markov kernel on $Y_r$ to which
Proposition~\ref{prop:msc} can be applied.
Let
\[
    q_1(z,w):=\frac12\one_{\{|z-w|\leq1\}},
\]
and define
\begin{equation}\label{eq:product-kernel}
    \widehat p_1((x,z),(y,w))
    :=p_1(x,y)q_1(z,w).
\end{equation}
This kernel is symmetric and Markov with respect to
$\lambda$. By \eqref{eq:manifold-kernel-sandwich},
it is bounded below by a positive multiple of
$r^{-m}$ whenever
$d_r((x,z),(y,w))\leq b_1$. Moreover, it vanishes
unless
\[
    d_r((x,z),(y,w))\leq b_2,
    \quad b_2:=\sqrt{(b')^2+1},
\]
and is bounded above by a constant multiple of
$r^{-m}$. For sufficiently small $r$,
\eqref{eq:product-volume} gives
\[
    \lambda(B_{d_r}(\xi,b_i))\asymp r^m,
    \quad i=1,2.
\]
Thus $\widehat p_1$ satisfies
\eqref{eq:compatibility} with inner radius $b_1$
and outer radius $b_2$. Together with volume
doubling and the nonlocal Poincar\'e inequality
established above, this verifies the hypotheses
of Proposition~\ref{prop:msc} uniformly in $r$
and $\sigma$.

The iterated product kernels factor as
\[
    \widehat p_j((x,z),(y,w))
    =p_j(x,y)q_j(z,w).
\]
The one-dimensional Poincar\'e inequality and the
same smoothing argument verify the hypotheses of
Proposition~\ref{prop:msc} for $q_1$ on $\R$.
Consequently,
\[
    q_j(0,0)\asymp j^{-1/2},
    \quad j\geq j_0,
\]
after increasing $j_0$ so that $j_0\geq2$.
The Gaussian estimates, initially valid almost
everywhere, extend to $(0,0)$ because $q_j$
is continuous for $j\geq2$.

Applying Proposition~\ref{prop:msc} to
$\widehat p_j$, we obtain Gaussian estimates
on $Y_r$. By \eqref{eq:product-volume},
\[
    \lambda(B_{d_r}((x,0),\sqrt j))
    \asymp \sqrt j\min\{(r\sqrt j)^m,1\}.
\]
Using the factorization
$\widehat p_j((x,0),(y,0))=p_j(x,y)q_j(0,0)$
and $q_j(0,0)\asymp j^{-1/2}$, the Gaussian bounds
for $\widehat p_j$ yield
\eqref{eq:manifold-discrete-upper} and
\eqref{eq:manifold-discrete-lower} for $j\geq j_0$.

We justify evaluating the estimates for $\widehat p_j$ at
$z=w=0$. Fix $j\geq j_0$. By Fubini's theorem, for
$\sigma\otimes\sigma$-almost every $(x,y)$, the estimates
for $\widehat p_j((x,z),(y,w))$ hold for almost every
$(z,w)\in\R^2$. For each such $(x,y)$, choose a sequence
$(z_k,w_k)\to(0,0)$ along which these estimates hold.
Since
$
\widehat p_j((x,z_k),(y,w_k))
=p_j(x,y)q_j(z_k,w_k),
$
and $q_j$ is continuous at $(0,0)$ for $j\geq2$,
we may let $k\to\infty$. The volume comparison
\eqref{eq:product-volume} is uniform in $z_k,w_k$,
so the upper bound passes to the limit.

For the lower bound, fix $0<c'<c_3$, where $c_3$
is defined in \eqref{eq:msc-lower}. If
$d_{\M}(x,y)\leq c'rj$, then
\[
d_r((x,0),(y,0))
=\frac{d_{\M}(x,y)}{r}
\leq c'j<c_3j.
\]
By continuity of $d_r$, the condition in
\eqref{eq:msc-lower} also holds for $(z,w)$
sufficiently close to $(0,0)$. We may therefore
pass to the limit as above, obtaining
\eqref{eq:manifold-discrete-lower}.

Taking a countable intersection over $j\geq j_0$,
we obtain a common set of full $\sigma\otimes\sigma$
measure on which \eqref{eq:manifold-discrete-upper}
and \eqref{eq:manifold-discrete-lower} hold for
every $j\geq j_0$.

It remains to establish the upper bound for
the finitely many integers $1\leq j<j_0$.
By the Markov property and
\eqref{eq:manifold-kernel-sandwich},
\[
    p_j(x,y)\leq C_1r^{-m}.
\]
Moreover, $p_j(x,y)=0$ unless
$d_{\M}(x,y)\leq jb'r$. On this support,
\[
    \frac{d_{\M}(x,y)^2}{r^2j}
    \leq j(b')^2\leq j_0(b')^2.
\]
Since
\[
    \frac1{\min\{(r\sqrt j)^m,1\}}
    \geq r^{-m}j^{-m/2},
\]
increasing $C$ in
\eqref{eq:manifold-discrete-upper} establishes
the same upper bound for these remaining
values of $j$. This completes the proof.
\end{proof}

We now transfer the Gaussian estimates of
Lemma~\ref{lem:manifold-kernel} to local random walks
on the graph. We use the event $\mathcal E_{N,r}$ and
the partition $V_1,\ldots,V_N$ constructed in
Lemma~\ref{lem:cloud}. Recall that Lemma~\ref{lem:cloud} gives us a probability measure $\tilde{\mu}$ on $\M$ satisfying $\tilde{\mu}(V_i) = 1/N$ for all $1\leq i \leq N$.

\begin{lemma}\label{lem:lift}
Let $P\in\R^{N\times N}$ be a stochastic matrix
reversible with respect to a probability vector
$\omega=(\omega_1,\ldots,\omega_N)$. Suppose that,
for fixed $0<\alpha\leq\beta$ and positive constants
$c,C$, we have
\begin{equation}\label{eq:local-walk}
    \frac cN\leq\omega_i\leq\frac CN,\quad
    \frac{c}{Nr^m}\one_{\{|x_i-x_j|\leq\alpha r\}}
    \leq P_{ij}\leq
    \frac{C}{Nr^m}\one_{\{|x_i-x_j|\leq\beta r\}},
\end{equation}
and that these inequalities hold for all $1\leq i,j\leq N$,
including $i=j$.

If $\eta_0$ in Lemma~\ref{lem:cloud} is chosen
sufficiently small, then there exist an integer
$j_0\geq1$ and constants $c_0,c,C>0$ such that,
on $\mathcal E_{N,r}$, the iterates of $P$ satisfy
\begin{align}
    (P^j)_{uv}
    &\leq\frac{C}{N\min\{(r\sqrt j)^m,1\}}
    \exp\left(-c\frac{d_{\M}(x_u,x_v)^2}{r^2j}\right),
    \quad j\geq1,
    \label{eq:walk-upper}\\
    (P^j)_{uv}
    &\geq\frac{c}{N\min\{(r\sqrt j)^m,1\}}
    \exp\left(-C\frac{d_{\M}(x_u,x_v)^2}{r^2j}\right),
    \quad j\geq j_0,\quad d_{\M}(x_u,x_v)\leq c_0rj.
    \label{eq:walk-lower}
\end{align}
The estimates hold for all vertices $u,v$, with
constants depending only on the fixed parameters
in \eqref{eq:local-walk} and the structural data.
\end{lemma}

\begin{proof}
We first represent the discrete graph walk as a Markov
kernel on $\M$. Let $\widetilde\mu$ and
$V_1,\ldots,V_N$ be as in Lemma~\ref{lem:cloud},
and define
\begin{equation}\label{eq:lifted-kernel}
    \mathrm d\sigma(x):=N\omega_i\mathrm d\widetilde\mu(x)
    \text{ for } x\in V_i,\text{  and  }
    p_1(x,y):=\frac{P_{ij}}{\omega_j}
    \text{ for } x\in V_i,\ y\in V_j.
\end{equation}
Since $\widetilde\mu(V_i)=1/N$, we have
$\sigma(V_i)=\omega_i$, and hence $\sigma$ is a
probability measure. Moreover,
\eqref{eq:cloud-partition} and
\eqref{eq:local-walk} imply that
\[
    c\leq\frac{\mathrm d\sigma}{\mathrm d\mu}\leq C.
\]
By the reversibility of $P$,
${P_{ij}}/{\omega_j}
    ={P_{ji}}/{\omega_i},$
so $p_1$ is symmetric with respect to $\sigma$.
Furthermore, for $x\in V_i$,
\[
    \int_{\M}p_1(x,y)\,\dd\sigma(y)
    =\sum_{j=1}^N\frac{P_{ij}}{\omega_j}\sigma(V_j)
    =\sum_{j=1}^NP_{ij}=1.
\]
Thus $p_1$ is a $\sigma$-symmetric Markov kernel.

We next verify the locality bounds
\eqref{eq:manifold-kernel-sandwich}. For
$x\in V_i$ and $y\in V_j$, the partition
constructed in Lemma~\ref{lem:cloud} gives
\[
    d_{\M}(x,x_i)\leq\eta_0r,\quad
    d_{\M}(y,x_j)\leq\eta_0r.
\]
If $d_{\M}(x,y)\leq\alpha r/2$ and
$\eta_0\leq\alpha/4$, then
\[
    |x_i-x_j|
    \leq d_{\M}(x_i,x)+d_{\M}(x,y)
         +d_{\M}(y,x_j)
    \leq\alpha r.
\]
Consequently, the lower bound in
\eqref{eq:local-walk} implies
\[
    p_1(x,y)\geq cr^{-m}.
\]
Conversely, if $p_1(x,y)>0$, then
$P_{ij}>0$ and hence $|x_i-x_j|\leq\beta r$.
By \eqref{eq:local-chord}, for sufficiently
small $r$,
\[
    d_{\M}(x,y)
    \leq d_{\M}(x,x_i)+d_{\M}(x_i,x_j)
        +d_{\M}(x_j,y)
    \leq(\kappa_0\beta+2\eta_0)r.
\]
Together with the upper bound in
\eqref{eq:local-walk}, this proves
\eqref{eq:manifold-kernel-sandwich}, with
$b=\alpha/2$ and
$b'=\kappa_0\beta+2\eta_0$.

The construction also gives an exact
correspondence between the iterates of the two
kernels. Indeed, since $\sigma(V_i)=\omega_i$,
the Chapman--Kolmogorov identity yields, by
induction,
\begin{equation}\label{eq:lifted-powers}
    p_j(x,y)=\frac{(P^j)_{uv}}{\omega_v},
    \quad x\in V_u,\quad y\in V_v.
\end{equation}
We may therefore apply
Lemma~\ref{lem:manifold-kernel} to $p_j$.
For $x\in V_u$ and $y\in V_v$, the triangle
inequality gives
\[
    \big|d_{\M}(x,y)-d_{\M}(x_u,x_v)\big|
    \leq2\eta_0r.
\]

The preceding inequality gives
\[
\frac{d_{\M}(x,y)^2}{r^2j}
\geq
\frac{d_{\M}(x_u,x_v)^2}{2r^2j}
-\frac{4\eta_0^2}{j}.
\]
Since $j\geq1$, applying
\eqref{eq:manifold-discrete-upper} and
\eqref{eq:lifted-powers}, with
$\omega_v\asymp N^{-1}$, proves
\eqref{eq:walk-upper}.

For the lower bound, let $c'>0$ be the constant
in \eqref{eq:manifold-discrete-lower}.
Choose $c_0\leq c'/2$ and increase $j_0$, if
necessary, so that $2\eta_0\leq c'j/2$
whenever $j\geq j_0$. Then
\[
    d_{\M}(x_u,x_v)\leq c_0rj
    \quad\Longrightarrow\quad
    d_{\M}(x,y)\leq c'rj
\]
for all $x\in V_u$ and $y\in V_v$.
Applying \eqref{eq:manifold-discrete-lower}
and using \eqref{eq:lifted-powers} proves
\eqref{eq:walk-lower}.

Finally, the estimates of
Lemma~\ref{lem:manifold-kernel} hold
$\sigma\otimes\sigma$-almost everywhere.
Since each $V_u\times V_v$ has positive
$\sigma\otimes\sigma$ measure and $p_j$ is
constant on these products by
\eqref{eq:lifted-powers}, the preceding
bounds hold for every pair of vertices
$u,v$. This completes the proof.
\end{proof}

We now pass from the discrete-time random walk to its
continuous-time semigroup using the Poisson representation.
The Gaussian estimates obtained in Lemma~\ref{lem:lift}
then yield corresponding bounds for the continuous-time
transition probabilities.

\begin{lemma}\label{lem:poisson}
Under the hypotheses of Lemma~\ref{lem:lift}, define
\begin{equation}\label{eq:walk-semigroup}
    H_t:=e^{-(t/r^2)(I-P)},\quad
    u:=t/r^2,\quad K:=Nr^m.
\end{equation}
For every fixed $A_0>0$, there exists $C_{A_0}>0$
such that, whenever $0<t\leq1$ and $u\geq1$,
\begin{equation}\label{eq:poisson-upper}
    (H_t)_{ij}
    \leq e^{-u}\one_{\{i=j\}}
    +\frac{C_{A_0}}{Nt^{m/2}}
    \left(1+\frac{d_{\M}(x_i,x_j)}{\sqrt t}\right)^{-A_0}.
\end{equation}
Moreover, for every fixed $a'>0$, there exist constants
$u_{a'},c_{a'}>0$ such that, for $0<t\leq1$,
\begin{equation}\label{eq:poisson-lower}
    u\geq u_{a'},\quad
    d_{\M}(x_i,x_j)\leq a'\sqrt t
    \quad\Longrightarrow\quad
    (H_t)_{ij}\geq\frac{c_{a'}}{Nt^{m/2}}.
\end{equation}
Finally, if $K\geq1$ and $u\geq2\log(eK)$, the
diagonal term $e^{-u}\one_{\{i=j\}}$ in
\eqref{eq:poisson-upper} can be absorbed into
the second term by increasing $C_{A_0}$.
\end{lemma}

\begin{proof}
By the exponential series, the continuous-time
semigroup admits the Poisson representation
\begin{equation}\label{eq:poisson-series}
    H_t=e^{-u}\sum_{\ell=0}^{\infty}
    \frac{u^\ell}{\ell!}P^\ell.
\end{equation}
We first establish \eqref{eq:poisson-upper}, treating
the term $\ell=0$ separately. Set
$R=d_{\M}(x_i,x_j)/r$. For every integer
$\ell\geq1$, we have
\[
    1+\frac{R}{\sqrt u}
    \leq\left(1+\frac{R}{\sqrt\ell}\right)
    \left(1+\sqrt{\frac{\ell}{u}}\right).
\]
Since $\exp(-cv^2)(1+v)^{A_0}$ is bounded
uniformly for $v\geq0$, it follows that
\begin{equation}\label{eq:gaussian-polynomial}
    e^{-cR^2/\ell}
    \leq C_{A_0}
    \left(1+\frac{R}{\sqrt u}\right)^{-A_0}
    \left(1+\sqrt{\frac{\ell}{u}}\right)^{A_0}.
\end{equation}

Let $J_u$ be a Poisson random variable with mean $u$.
We claim that, uniformly for $u\geq1$,
\begin{align}
    \E\left[
    \one_{\{J_u\geq1\}}
    \left(1+\sqrt{J_u/u}\right)^{A_0}
    \right]
    &\leq C_{A_0},
    \label{eq:poisson-positive}\\
    \E\left[
    \one_{\{J_u\geq1\}}
    (J_u/u)^{-m/2}
    \left(1+\sqrt{J_u/u}\right)^{A_0}
    \right]
    &\leq C_{A_0,m}.
    \label{eq:poisson-negative}
\end{align}
To prove \eqref{eq:poisson-negative}, split the
expectation according to whether $J_u<u/2$ or
$J_u\geq u/2$. On $1\leq J_u<u/2$, the integrand
is bounded by $Cu^{m/2}$, while the Poisson
lower-tail bound gives
\[
    \P(J_u<u/2)\leq e^{-cu}.
\]
The contribution from this event is therefore
uniformly bounded for $u\geq1$.
On $J_u\geq u/2$, the integrand is bounded by
$(1+J_u/u)^q$ for some fixed integer $q$.
Using the factorial moments
$\E[(J_u)_k]=u^k$, we obtain
\[
    \sup_{u\geq1}
    \E\left[(1+J_u/u)^q\right]<\infty.
\]
This proves \eqref{eq:poisson-negative}. Moreover, since
$
\left(1+\sqrt{J_u/u}\right)^{A_0}
\leq C_{A_0}(1+J_u/u)^q
$
for any integer $q\geq A_0/2$, the preceding
moment estimate also proves \eqref{eq:poisson-positive}.

Next, observe that
\[
    \frac{1}{\min\{(r\sqrt\ell)^m,1\}}
    \leq 1+r^{-m}\ell^{-m/2}.
\]
Applying \eqref{eq:walk-upper} and
\eqref{eq:gaussian-polynomial} to the terms
$\ell\geq1$ in \eqref{eq:poisson-series}, and
then using \eqref{eq:poisson-positive} and
\eqref{eq:poisson-negative}, we obtain
\begin{align*}
    &e^{-u}\sum_{\ell=1}^{\infty}
    \frac{u^\ell}{\ell!}(P^\ell)_{ij}\\
    &\quad\leq
    \frac{C_{A_0}}{N}
    \left(1+\frac{R}{\sqrt u}\right)^{-A_0}
    \left(1+r^{-m}u^{-m/2}\right)\\
    &\quad=
    \frac{C_{A_0}}{N}(1+t^{-m/2})
    \left(1+\frac{d_{\M}(x_i,x_j)}{\sqrt t}
    \right)^{-A_0}.
\end{align*}
Since $t\leq1$, the factor $1+t^{-m/2}$ is
bounded by $2t^{-m/2}$. Adding the contribution
$e^{-u}\one_{\{i=j\}}$ from $\ell=0$ proves
\eqref{eq:poisson-upper}.

We now establish the near-diagonal lower bound
\eqref{eq:poisson-lower}. Restrict the sum in
\eqref{eq:poisson-series} to integers satisfying
$u/2\leq\ell\leq2u$. Since
$\E J_u=\var(J_u)=u$, Chebyshev's inequality gives
\[
    \P\left(|J_u-u|>\frac u2\right)
    \leq\frac4u.
\]
Consequently, for $u\geq8$,
\[
    \P(u/2\leq J_u\leq2u)\geq\frac12.
\]
Choose $u_{a'}$ sufficiently large that
$u/2\geq j_0$ and
$a'\sqrt u\leq c_0u/2$ whenever $u\geq u_{a'}$.
If $d_{\M}(x_i,x_j)\leq a'\sqrt t$, then,
for every $u/2\leq\ell\leq2u$,
\[
    d_{\M}(x_i,x_j)
    \leq a'r\sqrt u
    \leq c_0r\ell.
\]
Thus the lower bound \eqref{eq:walk-lower}
applies throughout this interval. Moreover,
since $\ell\leq2u$ and $t=r^2u$,
\[
    \min\{(r\sqrt\ell)^m,1\}
    \leq (2t)^{m/2},
    \qquad
    \frac{d_{\M}(x_i,x_j)^2}{r^2\ell}
    \leq 2(a')^2.
\]
It follows that
\[
    (P^\ell)_{ij}
    \geq \frac{c}{Nt^{m/2}}e^{-C(a')^2},
    \qquad u/2\leq\ell\leq2u.
\]
Substituting this estimate into
\eqref{eq:poisson-series}, we obtain
\begin{align*}
    (H_t)_{ij}
    &\geq
    \frac{c}{Nt^{m/2}}e^{-C(a')^2}
    \P(u/2\leq J_u\leq2u)\\
    &\geq\frac{c_{a'}}{Nt^{m/2}},
\end{align*}
which proves \eqref{eq:poisson-lower}.

Finally, we show that the diagonal term in
\eqref{eq:poisson-upper} can be absorbed into
the second term when $u\geq2\log(eK)$.
Since $Nt^{m/2}=Ku^{m/2}$, we have
\[
    Nt^{m/2}e^{-u}
    =Ku^{m/2}e^{-u}
    \leq C_mKe^{-u/2}
    \leq C_m.
\]
Therefore,
\[
    e^{-u}\leq\frac{C_m}{Nt^{m/2}}.
\]
Since the distance factor in
\eqref{eq:poisson-upper} equals one on the
diagonal, the first term can be absorbed
into the second by increasing $C_{A_0}$.
\end{proof}

\begin{proposition}\label{prop:normalized-heat}
Fix $A_0>0$ and $a\geq1$. On the event
$\mathcal E_{N,r}$ of Lemma~\ref{lem:cloud},
suppose that \eqref{eq:edge-sandwich} holds.
Then there exist constants $C_*,h_0>0$ such that
the heat semigroup $T_t$ defined in
\eqref{eq:laplacian} satisfies
\eqref{eq:heat-upper}--\eqref{eq:heat-lower}
simultaneously for all $h$ satisfying
\begin{equation}\label{eq:general-heat-window}
    C_*r^2\log(eK)\leq h^2\leq h_0^2.
\end{equation}
The constants have the dependence specified in
Theorem~\ref{thm:heat-epsilon}, with additional
dependence on the fixed parameters $\alpha,\beta$.
\end{proposition}

\begin{proof}
Define the random-walk transition matrix and its
stationary distribution by
\begin{equation}\label{eq:random-walk}
    P:=D^{-1}A,\qquad
    \omega_i:=\frac{d_i}{\sum_jd_j}.
\end{equation}
Since $A$ is symmetric, $P$ is stochastic and
reversible with respect to $\omega$.
By Lemma~\ref{lem:cloud}, the degrees satisfy
$d_i\asymp K$, and hence $\omega_i\asymp N^{-1}$.
Together with \eqref{eq:edge-sandwich}, this
verifies \eqref{eq:local-walk}, so the estimates
of Lemma~\ref{lem:poisson} apply to $P$.

The symmetric normalized graph Laplacian is
similar to the random-walk Laplacian. More precisely,
\begin{equation}\label{eq:semigroup-similarity}
    T_t=D^{1/2}H_tD^{-1/2},\qquad
    (T_t)_{ij}=\sqrt{\frac{d_i}{d_j}}\,(H_t)_{ij},
\end{equation}
where $H_t$ is defined in \eqref{eq:walk-semigroup}.
Since the degrees are uniformly comparable, the
bounds of Lemma~\ref{lem:poisson} transfer to
$T_t$, with modified constants.

We first establish \eqref{eq:heat-upper}.
By \eqref{eq:graph-geometry}, the graph distance
$d_N$ is comparable to the geodesic distance
$d_{\M}$ on the event $\mathcal E_{N,r}$.
Applying \eqref{eq:poisson-upper} at $t=h^2$
and $t=2h^2$, and using the last assertion of
Lemma~\ref{lem:poisson}, gives
\eqref{eq:heat-upper}, provided
$h_0\leq1/\sqrt{2}$ and $C_*\geq2$.
The resulting upper-bound constant depends on
$A_0$ and the fixed geometric parameters, but
not on the separation parameter $a$.

For the diagonal lower bound
\eqref{eq:heat-diagonal}, we apply
\eqref{eq:poisson-lower} with $i=j$ and
$t=2h^2$. Increasing $C_*$ if necessary
ensures that $2h^2/r^2$ exceeds the threshold
required in that estimate.

Next, suppose $d_N(i,j)\leq ah$.
By the metric comparison in
\eqref{eq:graph-geometry},
\[
    d_{\M}(x_i,x_j)\leq\kappa_0ah.
\]
We may therefore apply
\eqref{eq:poisson-lower} with
$a'=\kappa_0a$, provided
$h^2/r^2\geq u_{\kappa_0a}$.
Since $\log(eK)\geq1$, this condition
holds throughout \eqref{eq:general-heat-window}
after increasing $C_*$.
Together with \eqref{eq:semigroup-similarity},
this proves \eqref{eq:heat-lower}.
The increase in $C_*$ does not affect the
upper-bound constant in \eqref{eq:heat-upper}.

Finally, we choose the constants so that the
geometric restrictions required in the projection
construction are also satisfied. Increasing
$C_*$ further ensures
\[
    h\geq C_\delta r,\qquad
    ah\geq2C_\delta r,
\]
since $\log(eK)\geq1$.
We then decrease $h_0$, if necessary, so that
\[
    h_0\leq\frac1{\sqrt2},
    \qquad ah_0\leq R_0.
\]
All these choices depend only on $A_0$, $a$,
$\alpha$, $\beta$, and the structural data,
independently of $N$, $r$, $h$, and the
realized point cloud.
\end{proof}

\subsubsection{Proof of Theorem \ref{thm:heat-epsilon}}
\begin{proof}[Proof of Theorem~\ref{thm:heat-epsilon}]
Take $\alpha=\beta=1$ in Lemma~\ref{lem:cloud} and set
$\Gcal_N^{\eps}:=\mathcal E_{N,\eps}$. The matrix
\eqref{eq:epsilon-adjacency} satisfies \eqref{eq:edge-sandwich}.
Since $K=N\eps^m\geq1$, we have $\eps^{-m}=N/K\leq N$.
Equation~\eqref{eq:cloud-probability} therefore gives
\eqref{eq:epsilon-probability}. The same lemma proves
\eqref{eq:graph-geometry}--\eqref{eq:graph-volume-lower}.

Fix $A_0>0$ and $a\geq1$. Proposition~\ref{prop:normalized-heat}
proves \eqref{eq:heat-upper}--\eqref{eq:heat-lower} on
\eqref{eq:heat-window}, with the stated choices of constants. The event
$\mathcal E_{N,\eps}$ only concerns the counts in fixed cells at scale
$\eps$. All later estimates are deterministic consequences of this
event, and hold for every $h$ in the interval. No union bound over $h$
is required.
\end{proof}

\subsubsection{Proof of Theorem \ref{thm:heat-knn}}
\begin{proof}[Proof of Theorem~\ref{thm:heat-knn}]
Choose $\alpha,\beta$ from Lemma~\ref{lem:knn-radii}. Intersect its
event with $\mathcal E_{N,r}$ from Lemma~\ref{lem:cloud}, with
$r=(k/N)^{1/m}$ and $\eta_0$ chosen for these fixed $\alpha,\beta$.
Call the intersection $\Gcal_N^k$. Since $Nr^m=k$ and
$r^{-m}=N/k\leq N$,
\[
 \P((\Gcal_N^k)^c)\leq2Ne^{-ck}+Cr^{-m}e^{-ck}
 \leq CNe^{-c'k}.
\]
On $\Gcal_N^k$, \eqref{eq:edge-sandwich} holds. Applying
Lemma~\ref{lem:cloud} and Proposition~\ref{prop:normalized-heat} proves
all the required conclusions. As in the preceding proof, they hold
simultaneously for every admissible $h$.
\end{proof}
\subsection{Proofs of Theorem \ref{thm:projection} and Corollary \ref{cor:parameters}}
\subsubsection{Projections from localized columns}\label{app:projection}
We now turn to the proof of Theorem~\ref{thm:projection}.
We first establish the required estimates under the
following abstract assumptions, and then show that these
assumptions follow from Theorems~\ref{thm:heat-epsilon}
and \ref{thm:heat-knn} for the two graph constructions.

Let $V$ be a finite set of $N$ labels equipped with a
metric $d_V$ and the uniform probability measure $\mu_V$.
Let $L$ be a symmetric matrix and set $T_t=e^{-tL}$.

\begin{assumption}\label{ass:heat}
There are positive constants $c,C,R_0,\delta$ and an
exponent $m\geq1$ such that
\begin{align}
 \mu_V(B_V(i,R))&\leq C(R+\delta)^m
 \quad(R>0),
 \label{eq:abstract-volume-upper}\\
 \mu_V(B_V(i,R))&\geq cR^m
 \quad(\delta\leq R\leq R_0).
 \label{eq:abstract-volume-lower}
\end{align}
Fix $B>m+2$, $A_0>B+m$, and $h\geq\delta$.
For $\tau\in\{h^2,2h^2\}$, assume
\begin{equation}\label{eq:abstract-upper}
 0\leq(T_\tau)_{ij}\leq\frac{C_{A_0}}{Nh^m}
       (1+d_V(i,j)/h)^{-A_0},\quad
 (T_{2h^2})_{ii}\geq\frac{c}{Nh^m}.
\end{equation}
For the fixed value of $a\geq1$ used in
\eqref{eq:centers}, assume $2\delta\leq ah\leq R_0$
and
\begin{equation}\label{eq:abstract-lower}
 d_V(i,j)\leq ah\quad\Longrightarrow\quad
 (T_{h^2})_{ij}\geq\frac{c_a}{Nh^m}.
\end{equation}
The constants in
\eqref{eq:abstract-volume-upper}--\eqref{eq:abstract-upper}
are independent of $a$. The constant $c_a$ may
depend on $a$.
\end{assumption}

\begin{lemma}\label{lem:packing}
Under \eqref{eq:abstract-volume-upper}--\eqref{eq:abstract-volume-lower},
let $Z=\{z_1,\ldots,z_n\}$ be maximal strictly $ah$-separated, where
$2\delta\leq ah\leq R_0$. Then
\begin{equation}\label{eq:packing}
 n\asymp(ah)^{-m},\quad
 \#\{\ell:d_V(i,z_\ell)\leq R\}
       \leq C(1+R/(ah))^m\quad(R>0).
\end{equation}
The comparison constants in \eqref{eq:packing} are independent of $a,h$.
\end{lemma}
\begin{proof}
Put $b=ah$. The balls $B_V(z_\ell,b/2)$ are disjoint, and the balls
$B_V(z_\ell,b)$ cover $V$. Thus
\[
 nc(b/2)^m\leq1\leq nC(b+\delta)^m\leq nC(3b/2)^m.
\]
This proves the first assertion. The balls $B_V(z_\ell,b/2)$ whose
centers lie in $B_V(i,R)$ are disjoint and contained in
$B_V(i,R+b/2)$. Consequently,
\[
 \#\{\ell:d_V(i,z_\ell)\leq R\}\,c(b/2)^m
 \leq C(R+b/2+\delta)^m\leq C(R+b)^m.
\]
Dividing by $c(b/2)^m$ proves the second assertion.
\end{proof}

\begin{lemma}\label{lem:gram}
Suppose \eqref{eq:packing} holds. Let
$U=(u_1,\ldots,u_n)\in\R^{N\times n}$ have columns satisfying
\begin{equation}\label{eq:column-decay}
 |u_\ell(i)|\leq\frac{C}{\sqrt{Nh^m}}
                (1+d_V(i,z_\ell)/h)^{-B}.
\end{equation}
Write $H=U^*U=I+F$ and define
\begin{equation}\label{eq:weighted-norm}
 \|M\|_{B,ah}:=\max_\ell\sum_j|M_{\ell j}|
          (1+d_V(z_\ell,z_j)/(ah))^B.
\end{equation}
If $\|F\|_{B,ah}\leq1/2$, then $H$ is positive definite,
$\frac12 I\preceq H\preceq\frac32 I$, and
$\Pi=UH^{-1}U^*$ satisfies
\begin{equation}\label{eq:gram-projection-bound}
 |\Pi_{ij}|\leq\frac{C'}{Nh^m}
                  (1+d_V(i,j)/h)^{-B}.
\end{equation}
Here $C'$ may depend on $a,B,m$ and the constants in
\eqref{eq:packing} and \eqref{eq:column-decay}, but not on $N,n,h$.
\end{lemma}
\begin{proof}
Set $w_{\ell j}=(1+d_V(z_\ell,z_j)/(ah))^B$. The triangle inequality
gives $w_{\ell j}\leq w_{\ell k}w_{kj}$. Hence
\[
 \sum_j|(MN)_{\ell j}|w_{\ell j}
 \leq\sum_k|M_{\ell k}|w_{\ell k}\sum_j|N_{kj}|w_{kj}
 \leq\|N\|_{B,ah}\sum_k|M_{\ell k}|w_{\ell k}.
\]
Taking the maximum over $\ell$ proves
\begin{equation}\label{eq:algebra}
 \|MN\|_{B,ah}\leq\|M\|_{B,ah}\|N\|_{B,ah},\quad
 \|I\|_{B,ah}=1.
\end{equation}
Since $F$ is symmetric, we have
\[
\|F\|_{\op}
\leq \max_\ell\sum_j|F_{\ell j}|
\leq \|F\|_{B,ah}
\leq \frac12.
\]
Here the first inequality follows from the symmetry of $F$,
and the second from the definition in
\eqref{eq:weighted-norm}. Since $H=I+F$, we conclude that
$\frac12 I\preceq H\preceq\frac32 I$.

Let $c_j=(-1)^j4^{-j}\binom{2j}{j}$. By \eqref{eq:algebra}, the series
\[
 H^{-1/2}=\sum_{j=0}^{\infty}c_jF^j
\]
converges in \eqref{eq:weighted-norm}, and
\begin{equation}\label{eq:inverse-half}
 \|H^{-1/2}\|_{B,ah}
 \leq\sum_{j=0}^{\infty}|c_j|2^{-j}=\sqrt2.
\end{equation}
The series also converges in operator norm to $H^{-1/2}$.
Put $Q=UH^{-1/2}$. Then $Q^*Q=I$ and $\Pi=QQ^*$.

For any $i,j,\ell$, the triangle inequality implies
\[
 (1+d_V(i,z_j)/h)^{-B}
 \leq a^B(1+d_V(i,z_\ell)/h)^{-B}w_{j\ell}.
\]
Since $H^{-1/2}$ and $w_{j\ell}$ are symmetric,
\eqref{eq:inverse-half} also gives
\[
\sum_j |(H^{-1/2})_{j\ell}|w_{j\ell}\leq\sqrt{2},
\quad 1\leq\ell\leq n.
\] By \eqref{eq:column-decay} and
\eqref{eq:inverse-half},
\begin{equation}\label{eq:orthogonal-column}
 |Q_{i\ell}|\leq\frac{C_a}{\sqrt{Nh^m}}
                    (1+d_V(i,z_\ell)/h)^{-B}.
\end{equation}
Moreover, \eqref{eq:packing} and $B>m$ give
\begin{equation}\label{eq:sum-centers}
 \sup_i\sum_\ell(1+d_V(i,z_\ell)/h)^{-B}
 \leq C\left(1+\sum_{j\geq0}2^{-jB}(1+2^{j+1}/a)^m\right)<\infty.
\end{equation}
For each $\ell$, at least one of $d_V(i,z_\ell)$ and
$d_V(j,z_\ell)$ is at least $d_V(i,j)/2$. 
Splitting the sum over these two possibilities and applying
\eqref{eq:sum-centers}, we obtain
\[
 \sum_\ell(1+d_V(i,z_\ell)/h)^{-B}
            (1+d_V(j,z_\ell)/h)^{-B}
 \leq C(1+d_V(i,j)/h)^{-B}.
\]
Since $\Pi_{ij}=\sum_\ell Q_{i\ell}Q_{j\ell}$, this bound
and \eqref{eq:orthogonal-column} prove
\eqref{eq:gram-projection-bound}.
\end{proof}

\begin{proposition}\label{prop:abstract-projection}
There is $a_0\geq1$, depending only on the constants in
\eqref{eq:abstract-volume-upper}--\eqref{eq:abstract-upper}, such that
the following holds. Fix $a\geq a_0$ and suppose
Assumption~\ref{ass:heat} holds with this value of $a$. Then
\eqref{eq:projection} satisfies
\eqref{eq:rank-bound}--\eqref{eq:intensity-bound}, with $d_N$ replaced
by $d_V$. If the vertices have positions $x_i\in\R^d$ satisfying
$|x_i-x_j|\leq d_V(i,j)$, then \eqref{eq:raw-variance} also holds.
\end{proposition}
\begin{proof}
Lemma~\ref{lem:packing} gives $n\asymp(ah)^{-m}$. For $1\leq\ell\leq n$, define
\begin{equation}\label{eq:normalized-columns}
 b_\ell:=(T_{2h^2})_{z_\ell z_\ell},\quad
 u_\ell:=b_\ell^{-1/2}T_{h^2}e_{z_\ell},\quad
 U:=(u_1,\ldots,u_n),\quad H:=U^*U.
\end{equation}
By symmetry of $L$,
\begin{equation}\label{eq:normalized-gram}
 H_{\ell j}=\frac{(T_{2h^2})_{z_\ell z_j}}{\sqrt{b_\ell b_j}},
 \quad H_{\ell\ell}=1,\quad
 b_\ell\asymp(Nh^m)^{-1}.
\end{equation}
Let $F=H-I$. For $\ell\ne j$, \eqref{eq:abstract-upper} gives
\[
 |F_{\ell j}|\leq C(1+d_V(z_\ell,z_j)/h)^{-A_0}.
\]
Fix $\ell$. There are at most $C2^{(v+1)m}$ centers with
$2^v ah<d_V(z_\ell,z_j)\leq2^{v+1}ah$. Therefore,
\begin{align}
 \sum_j|F_{\ell j}|(1+d_V(z_\ell,z_j)/(ah))^B
 &\leq C\sum_{v\geq0}2^{(v+1)m}(1+a2^v)^{-A_0}
                          (1+2^{v+1})^B\notag\\
 &\leq Ca^{-A_0}\sum_{v\geq0}2^{v(m+B-A_0)}.
 \label{eq:gram-small}
\end{align}
The series converges because $A_0>B+m$. We fix $a$ sufficiently large
that the right-hand side is at most $1/2$.

Equations \eqref{eq:abstract-upper} and \eqref{eq:normalized-gram} imply
\eqref{eq:column-decay}. Lemma~\ref{lem:gram} proves
\eqref{eq:projection-decay}. Normalizing the columns does not change
their span, so $UH^{-1}U^*$ equals \eqref{eq:projection}. We also have
$n\asymp h^{-m}$, since $a$ is now fixed. In particular,
\[
 \pi_i\leq\frac{C}{Nh^m}\lesssim\frac nN.
\]
For the reverse bound, choose $z_\ell$ with $d_V(i,z_\ell)\leq ah$.
By \eqref{eq:abstract-lower} and \eqref{eq:normalized-gram},
\[
 |u_\ell(i)|^2
 =\frac{|(T_{h^2})_{iz_\ell}|^2}{b_\ell}
 \geq\frac{c_a^2/(N^2h^{2m})}{C/(Nh^m)}
 \geq\frac{c'}{Nh^m}.
\]
Since $H^{-1}\succeq\frac23I$, the $i$th row $u(i)$ of $U$ satisfies
\[
 \pi_i=u(i)H^{-1}u(i)^*\geq\frac23\|u(i)\|^2
       \geq\frac{2c'}{3Nh^m}\gtrsim\frac nN.
\]
This proves \eqref{eq:intensity-bound}.

To prove \eqref{eq:raw-variance}, write $H_f=[f]_{s,\X}$. By
Lemma~\ref{lem:dpp},
\begin{equation}\label{eq:variance-sum}
 \var[\Sset_{h,N}(f)\mid\X]
 \leq\frac{CH_f^2}{N^2h^{2m}}
   \sum_{i,j}d_V(i,j)^{2s}(1+d_V(i,j)/h)^{-2B}.
\end{equation}
For fixed $i$, first sum over $d_V(i,j)\leq h$, and then over
$2^v h<d_V(i,j)\leq2^{v+1}h$. By
\eqref{eq:abstract-volume-upper} and $h\geq\delta$,
\begin{align}
 \frac1N\sum_jd_V(i,j)^{2s}(1+d_V(i,j)/h)^{-2B}
 &\leq Ch^{m+2s}
       \left(1+\sum_{v\geq0}2^{v(m+2s-2B)}\right)\notag\\
 &\leq C_s h^{m+2s}.
 \label{eq:moment-bound}
\end{align}
Substitution into \eqref{eq:variance-sum} gives
$\var[\Sset_{h,N}(f)\mid\X]\leq C_sH_f^2h^{2s-m}$.
Since $n\asymp h^{-m}$, division by $n^2$ proves the claim.
\end{proof}

\subsubsection{Proof of Theorem \ref{thm:projection}}
\begin{proof}
Fix $B>m+2$ and $A_0>B+m$. The constants in
\eqref{eq:graph-volume-upper}--\eqref{eq:graph-volume-lower},
\eqref{eq:heat-upper}, and \eqref{eq:heat-diagonal}
are independent of $a$. We may therefore choose $a\geq1$
sufficiently large that the right-hand side of
\eqref{eq:gram-small} is at most $1/2$. For this fixed
$a$, we then choose $C_*$ and $h_0$ as in
Proposition~\ref{prop:normalized-heat}.

On $\Gcal_N^{\eps}$ or $\Gcal_N^k$, respectively, the
graph is connected. For every $h$ satisfying
\eqref{eq:heat-window} or \eqref{eq:knn-heat-window},
the estimates
\eqref{eq:graph-volume-upper}--\eqref{eq:heat-lower}
verify Assumption~\ref{ass:heat}, with
$V=\{1,\ldots,N\}$, $d_V=d_N$, $L=\Lcal_N$, and
$\delta=\delta_N$. Moreover, \eqref{eq:chord-graph} gives
$|x_i-x_j|\leq d_N(i,j)$, as required in
Proposition~\ref{prop:abstract-projection}.
Applying that proposition gives
\eqref{eq:rank-bound}--\eqref{eq:raw-variance}.
\end{proof}

\subsubsection{Proof of Corollary \ref{cor:parameters}}
\begin{proof}

For \eqref{eq:explicit-epsilon}, we have $N\eps_N^m=K_0\log N$.
Choose $K_0$ larger than the constant in \eqref{eq:epsilon-regime}.
The failure probability in \eqref{eq:epsilon-probability} is
\[
 CN e^{-cK_0\log N}=CN^{1-cK_0}.
\]
Increasing $K_0$ so that $cK_0>q+2$, and then taking $N$ sufficiently
large, makes this at most $N^{-q}$. Moreover,
\[
 h_N^2=C_*\left(\frac{K_0\log N}{N}\right)^{2/m}
                  \log(eK_0\log N)\longrightarrow0.
\]
Thus $\eps_N\leq r_0$ and $h_N\leq h_0$ for all sufficiently large
$N$. The lower endpoint of \eqref{eq:heat-window} holds by definition.
Theorem~\ref{thm:projection} applies and gives
\[
 n\asymp h_N^{-m}
 =\frac{N}{K_0\log N\,[C_*\log(eK_0\log N)]^{m/2}}
 \asymp_q\frac{N}{\log N(\log\log N)^{m/2}}.
\]

For \eqref{eq:explicit-knn}, we have
$k_N=\lceil K_0\log N\rceil$ and
$r_N=(k_N/N)^{1/m}\to0$. Increasing $K_0$ if necessary
ensures that \eqref{eq:knn-regime} holds and
$CNe^{-ck_N}\leq N^{-q}$ for all sufficiently large $N$.
Moreover,
\[
h_N^2=C_*\left(\frac{k_N}{N}\right)^{2/m}
\log(ek_N)\longrightarrow0.
\]
Thus $h_N\leq h_0$ eventually, while the lower endpoint
of \eqref{eq:knn-heat-window} holds by definition.
Applying Theorem~\ref{thm:projection}, we obtain
\[
n\asymp h_N^{-m}
=\frac{N}{k_N[C_*\log(ek_N)]^{m/2}}
\asymp_q\frac{N}{\log N(\log\log N)^{m/2}}.
\]
Finally, the conclusions of Theorem~\ref{thm:projection}
hold simultaneously for every $h_N\leq h\leq h_0$
on the same event, proving the last assertion.
\end{proof}
\subsection{Proofs of Theorem \ref{thm:estimator} and Proposition \ref{prop:population}}
We first record the estimate for a general approximation to $f$.

\begin{lemma}\label{lem:residual}
Conditional on $\X$, let $\Pi$ be a rank-$n$ orthogonal
projection with $\Pi_{ii}=\pi_i\geq cn/N$, and let $\Sset$
be the associated projection DPP. Suppose that $Af$ is
determined by $\X$ and evaluations of $f$ at vertices
chosen independently of $\Sset$, and that $I_N(Af)$
can be computed from these evaluations. Then
\begin{equation}\label{eq:general-estimator}
 \widehat I(f):=I_N(Af)+\frac1N\sum_{i\in\Sset}
                  \frac{f(x_i)-Af(x_i)}{\pi_i}
\end{equation}
is conditionally unbiased for $I_N(f)$ and satisfies
\begin{equation}\label{eq:residual-bound}
 \var[\widehat I(f)\mid\X]
 \leq\frac1{N^2}\sum_i
 \frac{|f(x_i)-Af(x_i)|^2}{\pi_i}
 \leq\frac1{cn}\|f-Af\|_{L^2(\mu_N)}^2.
\end{equation}
\end{lemma}

\begin{proof}
Put $v_i=f(x_i)-Af(x_i)$ and $u_i=v_i/(N\pi_i)$. All these quantities
are fixed conditionally on the cloud and the values used to define $Af$.
Equation~\eqref{eq:dpp-mean} gives
\[
 \E[\widehat I(f)\mid\X]
 =I_N(Af)+\sum_i\pi_i u_i
 =I_N(Af)+I_N(f-Af)=I_N(f).
\]
Equation~\eqref{eq:dpp-bound} gives
\[
 \var[\widehat I(f)\mid\X]
 \leq\sum_i\pi_i u_i^2
 =\frac1{N^2}\sum_i\frac{v_i^2}{\pi_i}
 \leq\frac1{cnN}\sum_i v_i^2.
\]
This is \eqref{eq:residual-bound}.
\end{proof}

\begin{proof}[Proof of Theorem~\ref{thm:estimator}]
Condition on a cloud in the event of Theorem~\ref{thm:projection}.
The centers and the partition in \eqref{eq:approximation} are fixed
before drawing $\Sset_{h,N}$. For $i\in C_\ell$, equations
\eqref{eq:centers} and \eqref{eq:chord-graph} give
\[
 |f(x_i)-A_hf(x_i)|
 =|f(x_i)-f(x_{z_\ell})|
 \leq[f]_{s,\X}|x_i-x_{z_\ell}|^s
 \leq[f]_{s,\X}(ah)^s.
\]
Consequently,
\begin{equation}\label{eq:approximation-error}
 \|f-A_hf\|_{L^2(\mu_N)}^2
 \leq a^{2s}h^{2s}[f]_{s,\X}^2.
\end{equation}
The value of $I_N(A_hf)$ is given by \eqref{eq:approximation}, and
\eqref{eq:intensity-bound} verifies the hypothesis of
Lemma~\ref{lem:residual}. Applying \eqref{eq:residual-bound} and then
\eqref{eq:rank-bound} proves
\[
 \var[\widehat I_{h,N}(f)\mid\X]
 \leq C_s[f]_{s,\X}^2n^{-1}h^{2s}
 \leq C_s'[f]_{s,\X}^2n^{-1-2s/m}.
\]
Since the estimator is conditionally unbiased, this proves
\eqref{eq:mse}. There are $n$ centers and exactly $n$ sampled
vertices, so at most $2n$ function evaluations are needed. Finally,
Corollary~\ref{cor:parameters} gives the stated probability.
\end{proof}

\begin{proof}[Proof of Proposition~\ref{prop:population}]
Write
\[
 \widehat I_N-\int_{\M}f\dd\nu
 =\widehat I_N-I_N(f)+I_N(f)-\int_{\M}f\dd\nu.
\]
The first term has conditional mean zero given $\X$, whereas the second
is determined by $\X$. Their expected product is therefore zero.
Squaring and taking expectations gives
\[
 \E\left|\widehat I_N-\int_{\M}f\dd\nu\right|^2
 =\E\big[\var(\widehat I_N\mid\X)\big]
   +\var(I_N(f)).
\]
Independence of the $x_i$ gives
$\var(I_N(f))=\var_\nu(f)/N$, proving \eqref{eq:population}.
\end{proof}
\subsection{Proof of Theorem~\ref{clt:thm:main}}
\label{clt:sec:proof}
We prove the matching variance lower bounds and conditional
central limit theorems for both statistics in
Theorem~\ref{clt:thm:main}. Unless otherwise stated, we work
with a point cloud belonging to the event in
Theorem~\ref{thm:projection}. All expectations and variances
are taken with respect to the DPP, with the cloud held fixed.
The constants are uniform over point clouds in this event.

\begin{lemma}\label{clt:lem:positive}
Fix $A_0$ and $B$ as in Theorem~\ref{thm:projection}.
There exist constants $a_0\geq 1$ and $c>0$,
depending only on $A_0$, $B$, and the structural data,
such that the following holds for every $a\geq a_0$.
On the corresponding high-probability event of
Theorem~\ref{thm:projection}, for every $h$ in the range
specified therein,
\begin{equation}\label{clt:eq:positive}
    (\Pi_h)_{ij}\geq\frac{c}{Nh^m},
    \qquad i,j\in B_N(z_\ell,h),
\end{equation}
for every center $z_\ell \in Z$.
\end{lemma}

\begin{proof}
We consider the matrix of normalized columns $U$ and its
Gram matrix $H$ introduced in \eqref{eq:normalized-columns}.
Recall the weighted matrix norm
\[
 \|M\|_{B,ah}
 :=\max_{1\leq\ell\leq n}
 \sum_{j=1}^n |M_{\ell j}|
 \left(1+\frac{d_N(z_\ell,z_j)}{ah}\right)^B.
\]
Writing $H=I+F$, \eqref{eq:gram-small}, gives us that
\begin{equation}\label{eq:theta}
     \|F\|_{B,ah}\leq C_Fa^{-A_0}=:\theta(a),
\end{equation}

where $C_F$ is independent of $a$.

By \eqref{eq:normalized-columns}, we have $u_\ell(i)=b_\ell^{-1/2}(T_{h^2})_{iz_\ell}$.
The upper bound on the heat kernel in \eqref{eq:heat-upper},
together with the estimate $b_\ell\asymp (Nh^m)^{-1}$ from
\eqref{eq:normalized-gram}, therefore gives
\[
0\leq u_\ell(i)\leq
\frac{C}{\sqrt{Nh^m}}
\left(1+\frac{d_N(i,z_\ell)}{h}\right)^{-A_0}.
\]

Next using the packing estimate \eqref{eq:packing}, we get that,
for any $\beta>m$, we have
\begin{align}
\sum_{\ell=1}^n
\left(1+\frac{d_N(i,z_\ell)}{h}\right)^{-\beta}
=
\beta\int_0^\infty (1+t)^{-\beta-1}
\#\{\ell:d_N(i,z_\ell)\leq th\}\,\mathrm dt
\\
\leq
C\beta\int_0^\infty
\left(1+\frac{t}{a}\right)^m
(1+t)^{-\beta-1}\,\mathrm dt
\leq C_\beta,
\label{clt:eq:center-sum}
\end{align}
where $C_\beta$ is independent of $a$.
Consequently, if
$u(i)$ denotes the $i$th row of $U$, then
\begin{equation}\label{clt:eq:row-bound}
 \|u(i)\|^2\leq\frac{C_U}{Nh^m},
\end{equation}
with $C_U$ independent of $a$.

For each fixed $a$, we choose $C_*$ and $h_0$ in
Theorem~\ref{thm:projection} so that \eqref{eq:heat-lower} holds both for $d_N(i,j)\leq h$ and for
$d_N(i,j)\leq ah$, with the constant in the former
estimate independent of $a$.

We next establish a lower bound for $u_\ell(i)$ when
$d_N(i,z_\ell)\leq h$, with a constant independent of $a$.
By \eqref{eq:heat-lower}, we have
$(T_{h^2})_{iz_\ell}\geq c_1/(Nh^m)$ for
$d_N(i,z_\ell)\leq h$, where $c_1>0$ is independent of $a$.
Since $u_\ell(i)=b_\ell^{-1/2}(T_{h^2})_{iz_\ell}$
and $b_\ell\leq C/(Nh^m)$ by
\eqref{eq:normalized-gram}, we obtain, for
$i,j\in B_N(z_\ell,h)$,
\begin{equation}\label{clt:eq:unorthogonal-positive}
(UU^*)_{ij}\geq u_\ell(i)u_\ell(j)
\geq\frac{c_0}{Nh^m},
\end{equation}
where $c_0>0$ is independent of $a$.

Choose $a\geq4$ sufficiently large that
\[
 \theta(a)\leq\min\{1/2,c_0/(4C_U)\},
\]
where $\theta(a)$ is defined in \eqref{eq:theta}. Since $F$ is symmetric, the definition of the weighted norm
in \eqref{eq:weighted-norm} gives
$\|F\|_{\mathrm{op}}\leq \max_\ell\sum_j |F_{\ell j}|\leq \|F\|_{B,ah}\leq \theta(a).$
Thus, using the fact that $\lambda_{\min}(H) \ge  1-\theta(a) \ge 1/2$, we get $\|H^{-1}-I\|_{\mathrm{op}} \leq \|H^{-1}\|_{\mathrm{op}}\|F\|_{\mathrm{op}}\leq2\theta(a)$, and
\eqref{clt:eq:row-bound} gives
\[
 |(\Pi_h-UU^*)_{ij}|
 =|u(i)(H^{-1}-I)u(j)^*|
 \leq\frac{2C_U\theta(a)}{Nh^m}
 \leq\frac{c_0}{2Nh^m}.
\]
Combining this with \eqref{clt:eq:unorthogonal-positive} proves
\eqref{clt:eq:positive}. 
\end{proof}

We now turn to the variance bounds. To apply
\eqref{clt:eq:positive}, we need sufficiently many vertices
within distance $h$ of the centers in $Z$. By
\eqref{eq:empirical-balls} and
\eqref{eq:manifold-volume}, there exist positive constants
$c,C,R_*$ such that
\begin{equation}\label{clt:eq:ball-mass}
 \mu_N(B_{\M}(x,R))\geq cR^m,
 \quad x\in\M,\ \  Cr\leq R\leq R_*.
\end{equation}
Indeed, for $R\geq 2\eta_0r$, the lower bound in
\eqref{eq:empirical-balls} is at least
$c\mu(B_{\M}(x,R/2))$.

Moreover, the graph parameters in
Corollary~\ref{cor:parameters} satisfy $K\asymp\log N$.
Every admissible sequence in Theorem~\ref{clt:thm:main}
therefore satisfies
\begin{equation}\label{clt:eq:scales}
 h\longrightarrow0,\quad
 \frac{r}{h}\leq
 \frac1{\sqrt{C_*\log(eK)}}\longrightarrow0,
 \quad
 \frac{\delta_N}{h}\longrightarrow0.
\end{equation}
In particular, \eqref{clt:eq:ball-mass} applies to balls
of radius $\rho h$ for any fixed $\rho>0$ and all
sufficiently large $N$.

We recall the following classical identities for projection DPPs. For a proof, see, for example, Chapter 4 of \cite{HKPV09}.

\begin{lemma}\label{lem:dpp}
Let $\Pi\in\R^{N\times N}$ be a rank-$n$ orthogonal projection and
let $\Sset$ have law \eqref{eq:dpp-law} with $\Pi_h$ replaced by $\Pi$.
Then, for every $u\in\R^N$,
\begin{align}
 \E\sum_{i\in\Sset}u_i&=\sum_i\Pi_{ii}u_i,\label{eq:dpp-mean}\\
 \var\left[\sum_{i\in\Sset}u_i\right]
 &=\sum_i\Pi_{ii}u_i^2-\sum_{i,j}\Pi_{ij}^2u_iu_j
 \leq\sum_i\Pi_{ii}u_i^2,\label{eq:dpp-bound}\\
 \var\left[\sum_{i\in\Sset}u_i\right]
 &=\frac12\sum_{i,j}(u_i-u_j)^2\Pi_{ij}^2.
 \label{eq:dpp-identity}
\end{align}
\end{lemma}

\begin{lemma}\label{clt:lem:variance}
Fix $a\geq a_0$, where $a_0$ is as in
Lemma~\ref{clt:lem:positive}, and let $h=h(N)$
satisfy \eqref{clt:eq:scales}. For every fixed nonconstant
$f\in C^1(\M)$, there exist constants $c_f,C_f>0$
such that, for all sufficiently large $N$, the variances
defined in \eqref{clt:eq:variances} satisfy
\begin{equation}\label{clt:eq:two-variances}
 c_fnh^2\leq V_{h,N}(f)\leq C_fnh^2,\qquad
 c_f\frac{h^2}{n}\leq W_{h,N}(f)\leq C_f\frac{h^2}{n}.
\end{equation}
The constants and the threshold for $N$ are uniform
over point clouds in the event of
Theorem~\ref{thm:projection}.
\end{lemma}

\begin{proof}
Since $\M$ is connected and $f$ is
nonconstant, there are $x_0\in\M$, $R>0$, and $g>0$ such that
$R\leq R_*$ and
\[
 |\nabla f(x)|\geq g \quad 
 \forall x\in\overline{B_{\M}(x_0,R)}.
\]
Let $U_0=B_{\M}(x_0,R/4)$, $U_1=B_{\M}(x_0,R/2)$, and put
$J_N=\{\ell:x_{z_\ell}\in U_1\}$. By
\eqref{eq:graph-geometry}, there are fixed constants
$\kappa_-,\kappa_+\geq1$ such that
\begin{equation}\label{clt:eq:metric}
 d_{\M}(x_i,x_j)\leq\kappa_-d_N(i,j),\qquad
 d_N(i,j)\leq\kappa_+d_{\M}(x_i,x_j).
\end{equation}
For large $N$, every vertex in $U_0$ lies within graph distance
$ah$ of a center in $U_1$, since $\kappa_-ah<R/4$.
Equations~\eqref{clt:eq:ball-mass} and
\eqref{eq:graph-volume-upper} therefore give
\[
 c(R/4)^m\leq\mu_N(U_0)
 \leq\sum_{\ell\in J_N}\mu_N(B_N(z_\ell,ah))
 \leq C|J_N|(a+1)^mh^m.
\]
It follows that
\begin{equation}\label{clt:eq:many-centers}
 |J_N|\geq c_fh^{-m}.
\end{equation}

Set $t_0=(4\kappa_+)^{-1}$. For $\ell\in J_N$, define
\[
v_\ell:=\frac{\nabla f(x_{z_\ell})}
                 {|\nabla f(x_{z_\ell})|},
\quad
y_\ell^\pm:=\exp_{x_{z_\ell}}(\pm t_0h v_\ell).
\]
Consider the geodesics
\[
s\longmapsto \exp_{x_{z_\ell}}(\pm s v_\ell),
\quad 0\leq s\leq t_0h,
\]
which join $x_{z_\ell}$ to $y_\ell^\pm$. Since
\[
\left.\frac{d}{ds}
f\bigl(\exp_{x_{z_\ell}}(\pm s v_\ell)\bigr)
\right|_{s=0}
=\pm|\nabla f(x_{z_\ell})|,
\]
the first-order expansion of $f$ gives
\[
f(y_\ell^\pm)-f(x_{z_\ell})
=\pm t_0h|\nabla f(x_{z_\ell})|+o(h),
\]
where the remainder is uniform in $\ell\in J_N$
by the continuity of $\nabla f$ and compactness of
the unit tangent bundle of $\M$.

Since $|\nabla f(x_{z_\ell})|\geq g$ for
$\ell\in J_N$, we conclude that
$\pm\bigl(f(y_\ell^\pm)-f(x_{z_\ell})\bigr)
\geq ({gt_0}/{2})h$
for all sufficiently large $N$, uniformly in $\ell\in J_N$.

Choose
\[
\rho_0=\min\left\{\frac{t_0}{4},
\frac{gt_0}{8\|\nabla f\|_\infty}\right\},
\qquad b_f=\frac{gt_0}{4},
\]
and let $V_\ell^\pm$ be the sets of labels $i$ such
that $x_i\in B_{\M}(y_\ell^\pm,\rho_0h)$.

For every $i\in V_\ell^\pm$, we have
$|f(x_i)-f(y_\ell^\pm)|
\leq \|\nabla f\|_\infty
d_{\M}(x_i,y_\ell^\pm)
\leq \|\nabla f\|_\infty\rho_0h.$
Combining this with the preceding estimate for
$f(y_\ell^\pm)-f(x_{z_\ell})$, we obtain
\[
\pm(f(x_i)-f(x_{z_\ell}))
\geq \frac{gt_0}{2}h
     -\|\nabla f\|_\infty\rho_0h
\geq \frac{3gt_0}{8}h
 \geq b_fh.
\]
Moreover, since $r/h\to0$ and $\rho_0>0$ is fixed,
\eqref{clt:eq:ball-mass} applies with $R=\rho_0h$
for all sufficiently large $N$. We have therefore established
\begin{equation}\label{clt:eq:signed-sets}
|V_\ell^\pm|\geq \rho_fNh^m, \text{ and }
\pm(f(x_i)-f(x_{z_\ell}))\geq b_fh
\quad \text{for }i\in V_\ell^\pm,
\end{equation}
with $\rho_f > 0$. Finally, for $i\in V_\ell^\pm$, the metric comparison
\eqref{clt:eq:metric} gives
\[
d_N(i,z_\ell)
\leq\kappa_+d_{\M}(x_i,x_{z_\ell})
\leq\kappa_+(t_0+\rho_0)h
\leq\frac{5h}{16}<\frac h2.
\]
Since $d_N(z_\ell,z_j)>ah\geq h$ for $j\neq\ell$,
we have 
$d_N(i,z_j)>h-{5h}/{16}
>d_N(i,z_\ell).
$
Thus every vertex in $V_\ell^\pm$ belongs to
$C_\ell$. The sets corresponding to different
centers are therefore disjoint, while
$V_\ell^+\cap V_\ell^-=\varnothing$ by
\eqref{clt:eq:signed-sets}.

For every $\ell\in J_N$, \eqref{clt:eq:signed-sets} gives
$f(x_i)-f(x_j)\geq 2b_fh,
\text{ for } i\in V_\ell^+, \ j\in V_\ell^-.$
Moreover, since $V_\ell^\pm\subseteq B_N(z_\ell,h)$,
we have $(\Pi_h)_{ij}\geq c/(Nh^m)$ by
\eqref{clt:eq:positive}.

Since $V_\ell^\pm\subseteq C_\ell$ and the sets $C_\ell$
are pairwise disjoint, the pairs $(i,j)$ considered
above are distinct for different values of $\ell$.
Applying \eqref{eq:dpp-identity} and using the symmetry
of $\Pi_h$, we obtain
\begin{align*}
V_{h,N}(f)
&\geq \sum_{\ell\in J_N}
\sum_{i\in V_\ell^+}\sum_{j\in V_\ell^-}
(f(x_i)-f(x_j))^2(\Pi_h)_{ij}^2\\
&\geq \sum_{\ell\in J_N}
|V_\ell^+|\,|V_\ell^-|(2b_fh)^2
\frac{c^2}{N^2h^{2m}}\\
&\geq c_f|J_N|h^2
\geq c_fh^{2-m}
\asymp_f nh^2.
\end{align*}
Here the last two inequalities follow from
\eqref{clt:eq:signed-sets} and the previously established
bound $|J_N|\gtrsim h^{-m}$.

For the second statistic, put
\begin{equation}\label{clt:eq:residual-values}
 u_i=\frac{f(x_i)-A_hf(x_i)}{N\pi_i}.
\end{equation}
On $V_\ell^\pm$ we have $A_hf(x_i)=f(x_{z_\ell})$. Also,
$\pi_i\leq C/(Nh^m)$ by \eqref{eq:intensity-bound} and
\eqref{eq:rank-bound}. Hence
\[
 u_i\geq c_fh^{m+1},  \text{ for } i\in V_\ell^+,\quad
 u_j\leq-c_fh^{m+1} \text{ for } j\in V_\ell^-.
\]
The term $I_N(A_hf)$ is fixed conditionally on $\X$. Applying
\eqref{eq:dpp-identity} to \eqref{clt:eq:residual-values} gives
\begin{align*}
 W_{h,N}(f)
 &\geq\sum_{\ell\in J_N}
       |V_\ell^+|\,|V_\ell^-|c_fh^{2m+2}
       \frac{c^2}{N^2h^{2m}}\\
 &\geq c_f|J_N|h^{2m+2}
 \geq c_fh^{m+2}
 \asymp_f\frac{h^2}{n}.
\end{align*}

Finally, $[f]_{1,\M}<\infty$. For pairs with sufficiently small
Euclidean distance this follows from \eqref{eq:local-chord} and
the geodesic Lipschitz bound; for the remaining pairs it follows
from boundedness of $f$. The upper bounds in
\eqref{clt:eq:two-variances} now follow from
\eqref{eq:raw-variance}, \eqref{eq:mse}, and
\eqref{eq:rank-bound}.
\end{proof}

\begin{lemma}\label{clt:lem:localized}
Consider a sequence of point clouds and scales $h=h(N)$
for which \eqref{eq:graph-volume-upper} and the packing
estimate \eqref{eq:packing} hold with $d_V=d_N$, a fixed
$a\geq1$, and constants independent of $N$. Assume also
that $\delta_N\leq h$.

Let $Q\in\R^{N\times n}$ satisfy $Q^*Q=I$ and
\begin{equation}\label{clt:eq:columns}
 |Q_{i\ell}|\leq\frac{C}{\sqrt{Nh^m}}
 \left(1+\frac{d_N(i,z_\ell)}{h}\right)^{-B},
 \quad B>m+2,
\end{equation}
where $B$ and $C$ are independent of $N$, and suppose
$n\to\infty$. Let $\Sset_N$ be the projection DPP associated
with $QQ^*$, and suppose that the real values $f_i$
satisfy
\begin{equation}\label{clt:eq:lipschitz}
 |f_i-f_j|\leq Ld_N(i,j),
\end{equation}
where $L$ is independent of $N$.

Define $Y_N=\sum_{i\in\Sset_N}f_i$ and
$v_N=\var(Y_N)$. If $v_N\geq cnh^2$ for some
$c>0$ independent of $N$, then
\begin{equation}\label{clt:eq:abstract-limit}
 \frac{Y_N-\E Y_N}{\sqrt{v_N}}
 \Longrightarrow\mathcal N(0,1).
\end{equation}
\end{lemma}

\begin{proof}
Put $D_f=\diag(f_1,\ldots,f_N)$,
$\gamma_\ell=f_{z_\ell}$, and
$\Gamma=\diag(\gamma_1,\ldots,\gamma_n)$. For real $t$, define
\begin{equation}\label{clt:eq:matrix}
 \mathcal A(t)=e^{-it\Gamma/2}Q^*e^{itD_f}Qe^{-it\Gamma/2}.
\end{equation}
Recall that the Cauchy--Binet formula gives
\[
\det(Q^*e^{itD_f}Q)
=\sum_{\substack{J\subseteq\{1,\ldots,N\}\\|J|=n}}
e^{it\sum_{i\in J}f_i}\det(Q_{J,:})^2,
\]
where $Q_{J,:}$ denotes the submatrix of $Q$ formed
by the rows indexed by $J$. Since
\[
\P(\Sset_N=J)
=\det((QQ^*)_{J,J})
=\det(Q_{J,:})^2
\]
by \eqref{eq:dpp-law}, we obtain
\begin{equation}\label{clt:eq:determinant}
 \E e^{itY_N}=\det(Q^*e^{itD_f}Q),\qquad
 \det\mathcal A(t)=e^{-it\Tr\Gamma}\E e^{itY_N}.
\end{equation}
By the definition of $\mathcal A(t)$ and $Q^*Q=I$,
we have $\mathcal A(0)=I$. Moreover,
\[
 \mathcal A(t)_{\ell k}
 =\sum_iQ_{i\ell}Q_{ik}e^{it\Delta_{i\ell k}},
 \quad
 \Delta_{i\ell k}=f_i-\frac{\gamma_\ell+\gamma_k}{2}.
\]

Now, fix $0<\eta\leq1$ such that $2+\eta<B-m$. For
$p\in\{1,2,2+\eta\}$, set
\[
 W_p(\ell,k)=\sum_i|Q_{i\ell}Q_{ik}|\,|\Delta_{i\ell k}|^p,
 \qquad r_{i\ell}=d_N(i,z_\ell)/h.
\]
By \eqref{clt:eq:lipschitz},
$|\Delta_{i\ell k}|\leq Lh(r_{i\ell}+r_{ik})/2$. Thus
\eqref{clt:eq:columns} gives
\begin{align}
 \sum_kW_p(\ell,k)
 \leq\frac{C_ph^p}{Nh^m}\sum_i\bigg[&
 (1+r_{i\ell})^{-(B-p)}\sum_k(1+r_{ik})^{-B}\notag\\
 &+(1+r_{i\ell})^{-B}\sum_k(1+r_{ik})^{-(B-p)}\bigg].
 \label{clt:eq:moment-sum}
\end{align}

Since $B-p>m$, applying \eqref{clt:eq:center-sum}
with $\beta=B$ and $\beta=B-p$ gives
\[
\sup_i\sum_k(1+r_{ik})^{-B}\leq C,
\quad
\sup_i\sum_k(1+r_{ik})^{-(B-p)}\leq C.
\]
Moreover, for every $\beta>m$, \eqref{eq:graph-volume-upper}
and $\delta_N\leq h$ give
\begin{align*}
 \sum_i(1+r_{i\ell})^{-\beta}
 &=\beta\int_0^\infty
      \#B_N(z_\ell,ht)(1+t)^{-\beta-1}\,\mathrm dt\\
 &\leq CNh^m\beta\int_0^\infty
      (1+t)^{m-\beta-1}\,\mathrm dt
 \leq C_\beta Nh^m.
\end{align*}
Substituting these estimates into
\eqref{clt:eq:moment-sum}, we obtain
\begin{equation}\label{clt:eq:moment-bound}
 \max_\ell\sum_k W_p(\ell,k)\leq C_ph^p.
\end{equation}
Since $W_p$ is symmetric, we also have
\[
\|W_p\|_{\op}
\leq\max_\ell\sum_k W_p(\ell,k)
\leq C_ph^p.
\]
\\
Define real symmetric matrices $M_1,M_2$ by
\[
 (M_j)_{\ell k}
       =\sum_iQ_{i\ell}Q_{ik}\Delta_{i\ell k}^{j},
 \qquad j=1,2.
\]

Now, by the definition of $W_p(\ell,k)$,
for every $\ell,k$, we have
\[
|(M_j)_{\ell k}|\leq W_j(\ell,k),
\quad j=1,2.
\]
Moreover, since $\mathcal A(0)=I$, the inequalities
$|e^{iu}-1|\leq |u|$ and
$|e^{iu}-1-iu|\leq u^2/2$ give
\[
\begin{aligned}
|(\mathcal A(t)-I)_{\ell k}|
&\leq |t|W_1(\ell,k),\\
|(\mathcal A(t)-I-itM_1)_{\ell k}|
&\leq \frac{t^2}{2}W_2(\ell,k).
\end{aligned}
\]
Similarly, using
$|e^{iu}-1-iu+u^2/2|\leq C_\eta|u|^{2+\eta}$
for real $u$, we obtain
\[
\left|
\left(\mathcal A(t)-I-itM_1
+\frac{t^2}{2}M_2\right)_{\ell k}
\right|
\leq C_\eta|t|^{2+\eta}W_{2+\eta}(\ell,k).
\]

By \eqref{clt:eq:moment-bound}, the absolute row
and column sums in each of these estimates are
bounded by the corresponding constants $C_ph^p$.
Consequently,
\begin{align}
 \|M_j\|_{\mathrm{op}}&\leq Ch^j\quad(j=1,2),\notag\\
 \|\mathcal A(t)-I\|_{\mathrm{op}}&\leq C|t|h,\notag\\
 \|\mathcal A(t)-I-itM_1\|_{\mathrm{op}}&\leq Ct^2h^2,
 \label{clt:eq:first-expansion}\\
 \mathcal A(t)&=I+itM_1-\frac{t^2}{2}M_2+R(t),\notag\\
 \|R(t)\|_{\mathrm{op}}&\leq C|th|^{2+\eta}.
 \label{clt:eq:second-expansion}
\end{align}

Write $F=Q^*D_fQ$ and $G=Q^*D_f^2Q$.
Substituting the definition of $\Delta_{i\ell k}$ into
the expressions for $M_1$ and $M_2$, and using
$Q^*Q=I$, we obtain
\[
M_1=F-\Gamma,\qquad
M_2=G-\Gamma F-F\Gamma+\Gamma^2.
\]

Consequently, the mean and variance identities for a DPP imply
\begin{equation}\label{clt:eq:traces}
 \Tr M_1=\E Y_N-\Tr\Gamma,\qquad
 \Tr(M_2-M_1^2)=\Tr(G-F^2)=v_N.
\end{equation}

For $|t|h$ sufficiently small, put $E(t)=\mathcal A(t)-I$ and
expand the matrix logarithm in its convergent series about $I$:
\[
 \Tr\log\mathcal A(t)
 =\Tr E(t)-\frac12\Tr(E(t)^2)
       +O\bigl(n\|E(t)\|_{\mathrm{op}}^3\bigr).
\]
By \eqref{clt:eq:first-expansion},
$E(t)^2=-t^2M_1^2+O_{\mathrm{op}}(|th|^3)$. Using
\eqref{clt:eq:second-expansion} and
$|\Tr M|\leq n\|M\|_{\mathrm{op}}$ therefore yields
\begin{equation}\label{clt:eq:logarithm}
 \Tr\log\mathcal A(t)
 =it\Tr M_1-\frac{t^2}{2}\Tr(M_2-M_1^2)
       +O(n|th|^{2+\eta}).
\end{equation}
Here the cubic errors are absorbed by $|th|^{2+\eta}$, since
$\eta\leq1$.

Fix $u\in\R$ and set $t=u/\sqrt{v_N}$. The variance lower bound
gives
\[
 |t|h\leq\frac{|u|}{\sqrt{cn}},\qquad
 n|th|^{2+\eta}\leq C|u|^{2+\eta}n^{-\eta/2}\longrightarrow0.
\]
Since $\det\mathcal A(t)=\exp(\Tr\log\mathcal A(t))$, equations
\eqref{clt:eq:determinant}, \eqref{clt:eq:traces}, and
\eqref{clt:eq:logarithm} give
\[
 \E\exp\left(iu\frac{Y_N-\E Y_N}{\sqrt{v_N}}\right)
 =\exp\{-u^2/2+o(1)\}.
\]
The continuity theorem for characteristic functions proves
\eqref{clt:eq:abstract-limit}.
\end{proof}

\begin{lemma}\label{clt:lem:bounded}
Let $\Sset_N$ have a rank-$n$ projection-DPP law,
where $n\to\infty$. Suppose that the real values
$g_N(i)$ satisfy $\sup_i|g_N(i)|\leq C$ and
\[
v_N:=\var\left(\sum_{i\in\Sset_N}g_N(i)\right)
\geq cn,
\]
where $c,C>0$ are independent of $N$. Then
\[
\frac{\displaystyle\sum_{i\in\Sset_N}g_N(i)
-\E\left[\sum_{i\in\Sset_N}g_N(i)\right]}
{\sqrt{v_N}}
\Longrightarrow\mathcal N(0,1).
\]
\end{lemma}

\
\begin{proof}
Let $Q\in\R^{N\times n}$ have orthonormal columns such that
$QQ^*$ is the projection associated with $\Sset_N$, and put
$D=\diag(g_N(1),\ldots,g_N(N))$. Since $Q^*Q=I$ and
$\|D\|_{\op}\leq C$, Taylor's formula gives
\[
 Q^*e^{itD}Q
 =I+itQ^*DQ-\frac{t^2}{2}Q^*D^2Q+R(t),
 \qquad
 \|R(t)\|_{\op}\leq C|t|^3.
\]
Moreover, by \eqref{eq:dpp-mean} and \eqref{eq:dpp-bound},
\[
\Tr(Q^*DQ)=\E\sum_{i\in\Sset_N}g_N(i),
\qquad
\Tr\bigl(Q^*D^2Q-(Q^*DQ)^2\bigr)=v_N.
\]
Expanding the matrix logarithm as in
\eqref{clt:eq:logarithm}, we obtain
\[
 \Tr\log(Q^*e^{itD}Q)
 =it\E\sum_{i\in\Sset_N}g_N(i)
 -\frac{t^2v_N}{2}+O(n|t|^3).
\]

For fixed $u\in\R$, set $t=u/\sqrt{v_N}$. Since
$v_N\geq cn$, we have
\[
n|t|^3\leq C_u n^{-1/2}\longrightarrow0.
\]
Thus \eqref{clt:eq:determinant}, with $D_f=D$ and
$\Gamma=0$, gives
\[
\E\exp\left(
iu\frac{\sum_{i\in\Sset_N}g_N(i)
-\E\sum_{i\in\Sset_N}g_N(i)}{\sqrt{v_N}}
\right)
\longrightarrow e^{-u^2/2}.
\]
The conclusion follows from L\'evy's continuity theorem.
\end{proof}

\begin{proof}[Proof of Theorem~\ref{clt:thm:main}]
Fix $a$ as in Lemma~\ref{clt:lem:positive}.
By Corollary~\ref{cor:parameters}, for $q>1$,
the failure probabilities of $\Gcal_N^\eps$
and $\Gcal_N^k$ are summable in $N$.
The Borel--Cantelli lemma therefore implies that,
for almost every realization of the infinite sequence
$(x_i)_{i\geq1}$, both events hold for all sufficiently
large $N$. Fix such a realization.

Since \eqref{clt:eq:scales} holds,
Lemma~\ref{clt:lem:variance} gives
\eqref{clt:eq:variance-orders}. Moreover,
$h\to0$ and \eqref{eq:rank-bound} imply
$n\to\infty$. The lower bounds in
\eqref{clt:eq:variance-orders} also show that
$V_{h,N}(f)$ and $W_{h,N}(f)$ are positive
for all sufficiently large $N$.

The matrix $Q=UH^{-1/2}$ constructed in Lemma~\ref{lem:gram}
satisfies $Q^*Q=I$, $QQ^*=\Pi_h$, and
\eqref{clt:eq:columns} by \eqref{eq:orthogonal-column}.
Furthermore, \eqref{clt:eq:metric} gives
\[
 |f(x_i)-f(x_j)|
 \leq\|\nabla f\|_\infty d_{\M}(x_i,x_j)
 \leq\kappa_-\|\nabla f\|_\infty d_N(i,j).
\]
Thus Lemma~\ref{clt:lem:localized} and the first lower bound in
\eqref{clt:eq:two-variances} give \eqref{clt:eq:usual}.

For \eqref{clt:eq:residual}, define
\[
 g_N(i)=\frac{n}{h}
       \frac{f(x_i)-A_hf(x_i)}{N\pi_i}.
\]
By \eqref{eq:intensity-bound} and \eqref{eq:centers},
\[
 \sup_i|g_N(i)|
 \leq C\kappa_-a\|\nabla f\|_\infty.
\]
Also, \eqref{clt:eq:two-variances} gives
\[
 \var\left[\sum_{i\in\Sset_{h,N}}g_N(i)\,\middle|\,\X\right]
 =\frac{n^2}{h^2}W_{h,N}(f)\geq c_fn.
\]
Lemma~\ref{clt:lem:bounded} applies. 
Since
\[
 \sum_{i\in\Sset_{h,N}}g_N(i)
 -\E\left[\sum_{i\in\Sset_{h,N}}g_N(i)\,\middle|\,\X\right]
 =\frac{n}{h}\bigl(\widehat I_{h,N}(f)-I_N(f)\bigr),
\]
Lemma~\ref{clt:lem:bounded} gives \eqref{clt:eq:residual}.
This proves both conditional central limit theorems for every
cloud sequence fixed above. Since $\Phi$ is continuous, the
convergence of the conditional distribution functions holds
for every $t\in\R$, as stated.

\end{proof}

\section{Proofs from Section \ref{sec:block-dpp}}\label{app:haar}
\subsection{Proof of Proposition~\ref{blk:prop:law}}

\begin{proof}
Fix $\X$ and let $J\subseteq\{1,\ldots,N\}$ have cardinality $n$.
If $J$ contains two vertices of $C_\ell$, the corresponding rows of
$(P_{\mathrm{sc}})_{J,J}$ are equal by \eqref{blk:eq:projection}.
Its determinant is then zero. Otherwise, since $|J|=n$ and
there are $n$ cells, $J$ contains exactly one vertex $u_\ell$
from each $C_\ell$. In the order $u_1,\ldots,u_n$,
the matrix $(P_{\mathrm{sc}})_{J,J}$ is diagonal with entries
$M_1^{-1},\ldots,M_n^{-1}$. Its determinant is the product in
\eqref{blk:eq:law}, which is the joint probability of $n$ independent
uniform choices $u_\ell\in C_\ell$. In particular, the inclusion
probability of $i\in C_\ell$ is $M_\ell^{-1}$.

For the equal splits in the proposition, let $D_{\pm}$ be diagonal,
with entry $1$ at $i\in C_\ell^+$ and $-1$ at $i\in C_\ell^-$.
Then $\psi_\ell=D_{\pm}\phi_\ell$ and
$P_{\mathrm H}=D_{\pm}P_{\mathrm{sc}}D_{\pm}$. For every $J$,
\[
 \det((P_{\mathrm H})_{J,J})
 =\det((D_{\pm})_{J,J})^2\det((P_{\mathrm{sc}})_{J,J})
 =\det((P_{\mathrm{sc}})_{J,J}).
\]
Thus the two projections give the same law.
\end{proof}

\subsection{Proof of Theorem~\ref{blk:thm:geometry}}

\begin{proof}
Let $d_N^{(0)}$ be the shortest-path distance associated
with $A^{t_N}$, and write $B_N^{(0)}(i,R)$ for the
corresponding closed balls. Let $\Gamma_N$ be the event
$\Gcal_N^{\eps}$ of Theorem~\ref{thm:heat-epsilon}
with $\eps=t_N$. Since $Nt_N^m=K_0\log N$,
\eqref{eq:epsilon-probability} gives
\[
\P(\Gamma_N^c)
\leq CN\exp(-cK_0\log N)
\leq N^{-q},
\]
provided $K_0$ is sufficiently large and $N\geq N_0$.
We also increase $N_0$, if necessary, so that $t_N$
satisfies \eqref{eq:epsilon-regime}.

Put $\delta=C_\delta t_N$. By
\eqref{eq:graph-geometry}--\eqref{eq:graph-volume-lower},
on $\Gamma_N$ the graph $A^{t_N}$ is connected and
\begin{align}
d_N^{(0)}(i,j)
&\leq C_g d_{\M}(x_i,x_j),
\label{blk:eq:base-metric}\\
\mu_N(B_N^{(0)}(i,R))
&\leq C_v(R+\delta)^m
&& (R>0),
\label{blk:eq:base-upper}\\
\mu_N(B_N^{(0)}(i,R))
&\geq c_vR^m
&& (\delta\leq R\leq R_0).
\label{blk:eq:base-lower}
\end{align}
Here $C_g,C_v,c_v$ are structural constants.

\emph{Comparison of $d_N$ and $d_N^{(0)}$.}
Fix $\X\in\Gamma_N$. We first show that every graph
in \eqref{blk:eq:graph-ranges} contains $A^{t_N}$
and has all its edge lengths bounded by $\ell_0$.

For the $\varepsilon$-graph, these properties follow
directly from $t_N\leq\eps\leq\ell_0$.
For $A^k$, we verify them separately.

If $|x_i-x_j|\leq t_N$, then
$d_N^{(0)}(i,j)\leq t_N$. Hence
\eqref{blk:eq:base-upper} gives
\[
\begin{aligned}
\#\{j:|x_i-x_j|\leq t_N\}
&\leq N\mu_N(B_N^{(0)}(i,t_N))\\
&\leq C_vN(t_N+\delta)^m\\
&=C_v(1+C_\delta)^mK_0\log N.
\end{aligned}
\]
Choose $K_1>C_v(1+C_\delta)^mK_0$.
For $k\geq K_1\log N$, every $j\neq i$ satisfying
$|x_i-x_j|\leq t_N$ belongs to $\mathcal N_k(i)$,
including when several vertices have the same distance
from $x_i$. Thus
\[
A^{t_N}_{ij}\leq A^k_{ij},
\qquad 1\leq i,j\leq N.
\]

We next show that $R_i\leq\ell_0$ for every $i$.
Put $b=\min\{\ell_0,R_0\}/2$ and increase $N_0$
so that $\delta\leq b$. By
\eqref{blk:eq:base-lower},
\[
\#\{j\neq i:d_N^{(0)}(i,j)\leq b\}
\geq c_vNb^m-1.
\]
Since $|x_i-x_j|\leq d_N^{(0)}(i,j)$ by
\eqref{eq:chord-graph}, every vertex counted
above lies within Euclidean distance $b$ of $x_i$.

Choose $0<c_*<\min\{1,c_vb^m/2\}$. For all
sufficiently large $N$ and every $k\leq c_*N$,
we have
\[
c_vNb^m-1\geq k.
\]
Consequently, $R_i\leq b$ for every $i$, and
every edge of $A^k$ has length at most
$b\leq\ell_0$.

Now let $A$ be either adjacency matrix in
\eqref{blk:eq:graph-ranges}, with shortest-path
distance $d_N$. We have shown that
$A^{t_N}_{ij}\leq A_{ij}$ for every $i,j$.
Since both graphs assign length $|x_i-x_j|$
to each edge, it follows that
\[
d_N(i,j)\leq d_N^{(0)}(i,j).
\]
Moreover, every edge of $A$ has length at most
$\ell_0$. Applying \eqref{eq:local-chord} along
a path from $i$ to $j$, and then taking the
infimum over all such paths, gives
\[
d_{\M}(x_i,x_j)\leq\kappa_0d_N(i,j).
\]
Together with \eqref{blk:eq:base-metric}, we obtain
\[
d_N(i,j)\leq d_N^{(0)}(i,j)
\leq C_gd_{\M}(x_i,x_j)
\leq C_g\kappa_0d_N(i,j).
\]
In particular,
\[
B_N^{(0)}(i,R)\subseteq B_N(i,R)
\subseteq B_N^{(0)}(i,C_g\kappa_0R).
\]

Applying \eqref{blk:eq:base-upper} to
$B_N^{(0)}(i,C_g\kappa_0R)$ and
\eqref{blk:eq:base-lower} to $B_N^{(0)}(i,R)$
gives
\begin{equation}\label{blk:eq:volume}
\begin{aligned}
\mu_N(B_N(i,R))
&\leq C(R+\delta)^m
&& (R>0),\\
\mu_N(B_N(i,R))
&\geq cR^m
&& (\delta\leq R\leq R_0).
\end{aligned}
\end{equation}
The estimates hold on the same event $\Gamma_N$,
simultaneously for every choice of graph parameters
in \eqref{blk:eq:graph-ranges}.

\emph{Bounds for $n$ and $C_\ell$.}
Fix $2\delta\leq h\leq R_0$ and let
$Z=\{z_1,\ldots,z_n\}$ be defined by
\eqref{blk:eq:centers}. By construction,
the balls $B_N(z_\ell,h/2)$ are pairwise
disjoint, while the balls $B_N(z_\ell,h)$
cover all vertices. Applying
\eqref{blk:eq:volume}, we obtain
\[
nc(h/2)^m
\leq\sum_{\ell=1}^n
\mu_N(B_N(z_\ell,h/2))
\leq1
\]
and
\[
1\leq\sum_{\ell=1}^n
\mu_N(B_N(z_\ell,h))
\leq nC(h+\delta)^m.
\]
Since $\delta\leq h/2$, these inequalities
give $n\asymp h^{-m}$.

For either the first-cover or Voronoi partition,
we have $C_\ell\subseteq B_N(z_\ell,h)$.
Consequently, \eqref{blk:eq:volume} gives
\[
w_\ell=\mu_N(C_\ell)
\leq\mu_N(B_N(z_\ell,h))
\leq C(h+\delta)^m
\leq\frac{C'}{n}.
\]
Furthermore, for every $i,j\in C_\ell$,
\[
\begin{aligned}
d_{\M}(x_i,x_j)
&\leq\kappa_0d_N(i,j)\\
&\leq\kappa_0\bigl(
d_N(i,z_\ell)+d_N(z_\ell,j)\bigr)
\leq2\kappa_0h.
\end{aligned}
\]
This proves \eqref{blk:eq:geometry}.

For the Voronoi partition, we also obtain a
lower bound for $w_\ell$. If
$i\in B_N(z_\ell,h/2)$, then for every
$j\neq\ell$, the strict separation in
\eqref{blk:eq:centers} gives
\[
\begin{aligned}
d_N(i,z_j)
&\geq d_N(z_\ell,z_j)-d_N(i,z_\ell)\\
&>h-d_N(i,z_\ell)
\geq d_N(i,z_\ell).
\end{aligned}
\]
Thus $i\in C_\ell$, and hence
$B_N(z_\ell,h/2)\subseteq C_\ell$.
By \eqref{blk:eq:volume},
\[
w_\ell\geq
\mu_N(B_N(z_\ell,h/2))
\geq c(h/2)^m
\geq\frac{c'}{n}.
\]
Finally, \eqref{blk:eq:law} gives
$\pi_i=M_\ell^{-1}$ for $i\in C_\ell$,
proving the remaining assertion.
\end{proof}

\subsection{Proofs of Theorems \ref{blk:thm:statistic} and
\ref{blk:thm:quadrature}}

\begin{proof}[Proof of Theorem~\ref{blk:thm:statistic}]
Condition on $\X$ and put
\[
 \sigma_\ell^2:=\frac{1}{M_\ell}\sum_{i\in C_\ell}
                  |f(x_i)-\bar f_\ell|^2.
\]
By Proposition~\ref{blk:prop:law}, $f(x_{U_\ell})$ has mean
$\bar f_\ell$ and variance $\sigma_\ell^2$. Independence gives
\[
 \E[\Sset_{\mathrm{bl}}(f)\mid\X]=\sum_{\ell=1}^n\bar f_\ell,
 \qquad
 \var[\Sset_{\mathrm{bl}}(f)\mid\X]=\sum_{\ell=1}^n\sigma_\ell^2.
\]
Expanding the squared differences within $C_\ell$ gives
\begin{equation}\label{blk:eq:pair-identity}
 \frac{1}{2M_\ell^2}\sum_{i,j\in C_\ell}|f(x_i)-f(x_j)|^2
 =\frac{1}{M_\ell}\sum_{i\in C_\ell}f(x_i)^2-\bar f_\ell^2
 =\sigma_\ell^2.
\end{equation}
Summing \eqref{blk:eq:pair-identity} proves \eqref{blk:eq:raw-exact}.
For $i,j\in C_\ell$, the definition of $\Delta_\ell$ gives
\begin{equation}\label{blk:eq:increment}
 |f(x_i)-f(x_j)|\leq[f]_{C^{0,s}}\Delta_\ell^s .
\end{equation}
There are $M_\ell^2$ ordered pairs in $C_\ell^2$, so
\eqref{blk:eq:raw-exact} gives \eqref{blk:eq:raw-diameter}. On
$\Gamma_N$, \eqref{blk:eq:geometry} bounds $\sum_\ell\Delta_\ell^{2s}$
by $n(2\kappa_0 h)^{2s}$. Dividing by $n^2$ and using
$h\asymp n^{-1/m}$ proves \eqref{blk:eq:raw-rate}.
\end{proof}

\begin{proof}[Proof of Theorem~\ref{blk:thm:quadrature}]
By \eqref{blk:eq:estimator},
\[
 \E[\widehat I_{\mathrm{bl}}(f)\mid\X]
 =\sum_{\ell=1}^n w_\ell\bar f_\ell
 =\frac1N\sum_{\ell=1}^n\sum_{i\in C_\ell}f(x_i)=I_N(f).
\]
The conditional mean squared error is therefore
$\sum_{\ell=1}^n w_\ell^2\sigma_\ell^2$, by independence of the
$U_\ell$. In \eqref{blk:eq:pair-identity} the factor $M_\ell^{-2}$
cancels with the $M_\ell^2$ in $w_\ell^2=M_\ell^2/N^2$. This gives
\eqref{blk:eq:mse-exact}, and applying \eqref{blk:eq:increment}
gives \eqref{blk:eq:mse-diameter}. On $\Gamma_N$ we have
\[
 \sum_{\ell=1}^n w_\ell^2\Delta_\ell^{2s}
 \leq (2\kappa_0 h)^{2s}\max_\ell w_\ell
       \sum_{\ell=1}^n w_\ell
 \leq C_s h^{2s}/n
\]
by \eqref{blk:eq:geometry} and $\sum_\ell w_\ell=1$.
Substituting $h\asymp n^{-1/m}$ proves \eqref{blk:eq:mse-rate}.
\end{proof}

\subsection{Proof of Corollary \ref{blk:cor:population}}

\begin{proof}
Theorem~\ref{blk:thm:quadrature} gives conditional unbiasedness on
every cloud. Since $f$ is bounded on $\M$ and $\sum_\ell w_\ell=1$,
$\widehat I_{\mathrm{bl}}(f)$ is square integrable. Thus
\eqref{eq:population} identifies the difference in the corollary with
$\E[\var(\widehat I_{\mathrm{bl}}(f)\mid\X)]$.
On $\Gamma_N$, \eqref{blk:eq:geometry} and \eqref{blk:eq:mse-rate}
bound this conditional variance by $C_s[f]_{C^{0,s}}^2h^{m+2s}$.
On $\Gamma_N^c$ we still have $\Delta_\ell\leq\operatorname{diam}(\M)$
and $\sum_\ell w_\ell^2\leq1$. Hence \eqref{blk:eq:mse-diameter}
bounds the conditional variance there by
$\tfrac12[f]_{C^{0,s}}^2\operatorname{diam}(\M)^{2s}$.
Taking expectations and using $\P(\Gamma_N^c)\leq N^{-q}$ proves
the claim.
\end{proof}

\section{Approximation of the heat columns}\label{app:numerics}
Throughout this appendix, matrix norms and factorizations
are taken with respect to the standard Euclidean inner
products on $\R^N$ and $\R^n$.

\begin{lemma}\label{lem:heat-approximation}
Let $T^{(q)}$ and $q$ be defined by
\eqref{eq:matrix-series}--\eqref{eq:polynomial-degree}. Then
\begin{equation*}
 \|T^{(q)}-T_{h^2}\|_{\op}\leq\eta,\qquad
 \|T^{(q)}E_Z-B_h\|_{\op}\leq\eta.
\end{equation*}
\end{lemma}
\begin{proof}
For \eqref{eq:matrix-series}, $S$ is symmetric and similar to a
stochastic matrix. Its eigenvalues therefore lie in $[-1,1]$ and
$\|S\|_{\op}\leq1$. The exponential series gives
\[
 \|T^{(q)}-T_{h^2}\|_{\op}
 \leq e^{-u}\sum_{j=q+1}^{\infty}\frac{u^j}{j!}
 =\P(J_u\geq q+1).
\]
For $v=q+1>u$, the exponential Markov inequality with
$e^\theta=v/u$ gives
\[
 \P(J_u\geq v)
 \leq e^{-u}(eu/v)^v\leq(eu/v)^v.
\]
The choice \eqref{eq:polynomial-degree} ensures $v\geq2eu$ and
$v\geq2\log(1/\eta)$. Hence the last expression is at most
$2^{-v}\leq\eta$. Since $\|E_Z\|_{\op}=1$, the second inequality
in the lemma follows.
\end{proof}

\begin{proposition}\label{prop:projection-approximation}
Suppose the conclusions of Theorem~\ref{thm:projection} hold, and let
$\widetilde B\in\R^{N\times n}$ satisfy
$\|\widetilde B-B_h\|_{\op}\leq\eta$. Write
$\kappa=\sigma_{\min}(B_h)$. Then
\begin{equation}\label{eq:smallest-singular-value}
 \kappa\geq c\sqrt{n/N}.
\end{equation}
If $\eta\leq\kappa/2$, the matrix $\widetilde B$ has rank $n$, and
the projection
$\widetilde\Pi=\widetilde B(\widetilde B^*\widetilde B)^{-1}
\widetilde B^*$ satisfies
\begin{equation}\label{eq:projector-error}
 \Delta:=\|\widetilde\Pi-\Pi_h\|_{\op}
       \leq4\eta/\kappa.
\end{equation}
If $\Delta\leq c_\pi n/(2N)$, where $c_\pi$ is a lower constant
in \eqref{eq:intensity-bound}, then
$\widetilde\pi_i:=\widetilde\Pi_{ii}\geq c_\pi n/(2N)$.
The estimator \eqref{eq:estimator}, using a DPP with projection
$\widetilde\Pi$ and its own weights $\widetilde\pi_i$, remains unbiased
and satisfies \eqref{eq:mse} with a changed constant.

Furthermore, for $\operatorname{osc}_{\X}(f)=\max_i f(x_i)-\min_i f(x_i)$,
\begin{equation}\label{eq:variance-perturbation}
 \left|\var_{\widetilde\Pi}[\Sset(f)]-
             \var_{\Pi_h}[\Sset(f)]\right|
 \leq\sqrt2\,n\Delta\,\operatorname{osc}_{\X}(f)^2.
\end{equation}
In particular, for fixed $0<s\leq1$, the sufficient accuracy
\begin{equation}\label{eq:sufficient-accuracy}
 \eta\leq c\sqrt{n/N}\min\{n/N,h^{2s}\}
\end{equation}
preserves both \eqref{eq:raw-variance} and \eqref{eq:mse}.
\end{proposition}

\begin{proof}
Recall the definitions of $U$, $H$, and $b_\ell$ in
\eqref{eq:normalized-columns}. Let
$D_b=\diag(b_1,\ldots,b_n)$. Since $B_h=UD_b^{1/2}$,
we have
\[
 B_h^*B_h=D_b^{1/2}HD_b^{1/2}
 \succeq\tfrac12 D_b
 \succeq\frac{c}{Nh^m}I
 \succeq\frac{c'n}{N}I.
\]
This proves \eqref{eq:smallest-singular-value}. Moreover,
for every $v\in\R^n$,
$
\|\widetilde Bv\|
\geq\|B_hv\|-\|(\widetilde B-B_h)v\|
\geq(\kappa-\eta)\|v\|.
$
Thus $\widetilde B$ has rank $n$ whenever $\eta<\kappa$.

We next prove \eqref{eq:projector-error}. For any unit
vector $w$ in the range of $\widetilde\Pi$, write
$w=\widetilde Bv$, where
$\|v\|\leq(\kappa-\eta)^{-1}$. Since
$(I-\Pi_h)B_h=0$, we obtain
$
\|(I-\Pi_h)\widetilde\Pi\|_{\op}
\leq{\eta}/({\kappa-\eta}).
$
Similarly, using $(I-\widetilde\Pi)\widetilde B=0$
and $\sigma_{\min}(B_h)=\kappa$, we have
$
\|(I-\widetilde\Pi)\Pi_h\|_{\op}
\leq{\eta}/{\kappa}.
$
Since
\[
\Pi_h-\widetilde\Pi
=\Pi_h(I-\widetilde\Pi)
 -(I-\Pi_h)\widetilde\Pi,
\]
it follows that
\[
\Delta\leq\frac{\eta}{\kappa-\eta}
+\frac{\eta}{\kappa}
\leq\frac{4\eta}{\kappa},
\]
where we used $\eta\leq\kappa/2$.

For every $i$, we have
\[
|\widetilde\pi_i-\pi_i|
=|e_i^*(\widetilde\Pi-\Pi_h)e_i|
\leq\Delta.
\]
Consequently, \eqref{eq:intensity-bound} gives
$\widetilde\pi_i\geq c_\pi n/(2N)$ whenever
$\Delta\leq c_\pi n/(2N)$. Lemma~\ref{lem:residual}
and \eqref{eq:approximation-error} then establish
conditional unbiasedness and \eqref{eq:mse} for
$\widetilde\Pi$, with a possibly different constant.

By \eqref{eq:dpp-identity} and Cauchy--Schwarz,
\begin{align*}
\left|\var_{\widetilde\Pi}[\Sset(f)]
-\var_{\Pi_h}[\Sset(f)]\right|
\leq
\frac{\operatorname{osc}_{\X}(f)^2}{2}
\sum_{i,j}
|\widetilde\Pi_{ij}-(\Pi_h)_{ij}|
|\widetilde\Pi_{ij}+(\Pi_h)_{ij}|\\
\leq
\frac{\operatorname{osc}_{\X}(f)^2}{2}
\|\widetilde\Pi-\Pi_h\|_{\HS}
\|\widetilde\Pi+\Pi_h\|_{\HS}.
\end{align*}
Since both projections have rank $n$, we have
\[
\|\widetilde\Pi-\Pi_h\|_{\HS}
\leq\sqrt{2n}\,\Delta,\qquad
\|\widetilde\Pi+\Pi_h\|_{\HS}
\leq2\sqrt n.
\]
Substituting these estimates proves
\eqref{eq:variance-perturbation}.

Finally, by \eqref{eq:smallest-singular-value}
and \eqref{eq:projector-error}, choosing the constant
in \eqref{eq:sufficient-accuracy} sufficiently small
ensures that
\[
\Delta\leq\frac{c_\pi n}{2N},
\qquad
\Delta\leq Ch^{2s}.
\]
The first inequality preserves \eqref{eq:mse},
as shown above. Moreover,
$\operatorname{osc}_{\X}(f)
\leq[f]_{s,\X}\diam(\M)^s$, so
\eqref{eq:variance-perturbation} gives
\[
\left|\var_{\widetilde\Pi}[\Sset(f)]
-\var_{\Pi_h}[\Sset(f)]\right|
\leq C_s[f]_{s,\X}^2nh^{2s}.
\]
Together with \eqref{eq:raw-variance}, this proves
the last assertion.
\end{proof}

\end{document}